\documentclass{article}

\newif\ificml
\icmlfalse

\usepackage{etoolbox} 

\ificml

\usepackage[dvipsnames]{xcolor}
\usepackage{color, colortbl}
\usepackage{multirow}
\usepackage{enumitem}
\usepackage{placeins}
\usepackage{subcaption}
\usepackage{microtype}
\usepackage{graphicx}
\usepackage{booktabs} 
\usepackage{multirow}
\usepackage{enumitem}
\usepackage{graphicx}
\usepackage{color}
\usepackage{xcolor}
\usepackage{colortbl}
\definecolor{dullGray}{HTML}{F3F3F3}
\definecolor{smallColorT}{HTML}{fff1e6}
\definecolor{largeColorT}{HTML}{edf2fb}

\usepackage{hyperref}

\newcommand{\theHalgorithm}{\arabic{algorithm}}

\usepackage[accepted]{ICML/icml2025}

\usepackage{amsmath}
\usepackage{amssymb}
\usepackage{mathtools}
\usepackage{amsthm}

\usepackage[capitalize,noabbrev]{cleveref}
\usepackage[table]{xcolor}
\usepackage{subcaption}
\theoremstyle{plain}

\theoremstyle{definition}

\theoremstyle{remark}

\usepackage[textsize=tiny]{todonotes}
\usepackage{xspace}

\definecolor{LightGrey}{HTML}{F3F3F3}
\newcommand\glb{ \rowcolor{LightGrey}}

\newcommand{\ourmethod}{\texttt{FARMS}\xspace}
\newcommand{\TB}{\texttt{TempBalance}\xspace}
\newcommand{\CAL}{\texttt{CAL}\xspace}
\newcommand{\Alphapruning}{\texttt{AlphaPruning}\xspace}
\newcommand{\TBSigmoid}{\texttt{TB\_Sigmoid}\xspace}
\newcommand{\AlphaHill}{\texttt{PL\_Alpha\_Hill}\xspace}
\newcommand{\Alpha}{\texttt{PL\_Alpha}\xspace}
\newcommand{\AlphaLoRA}{\texttt{AlphaLoRA}\xspace}
\newif\ifshowcomments
\showcommentstrue

\icmltitlerunning{Eigenspectrum Analysis of Neural Networks without Aspect Ratio Bias}

\begin{document}

\twocolumn[
\icmltitle{Eigenspectrum Analysis of Neural Networks without Aspect Ratio Bias}


\icmlsetsymbol{explanation}{*}

\begin{icmlauthorlist}
\icmlauthor{Yuanzhe Hu}{ucsd}
\icmlauthor{Kinshuk Goel}{dartmouth,srm}
\icmlauthor{Vlad Killiakov}{ucb}
\icmlauthor{Yaoqing Yang}{dartmouth}

\end{icmlauthorlist}

\icmlaffiliation{dartmouth}{Department of Computer Science, Dartmouth College, NH, USA}
\icmlaffiliation{ucsd}{Department of Computer Science and Engineering, University of California, San Diego, CA, USA}
\icmlaffiliation{srm}{Department of Computer Science \& Engineering, SRM Institute of Science \& Technology, India}
\icmlaffiliation{ucb}{Independent Researcher, University of California, Berkeley, CA, USA}


\icmlcorrespondingauthor{Yuanzhe Hu}{yuanzhe.hu@dartmouth.edu}

\icmlkeywords{Machine Learning, ICML}

\vskip 0.3in
]



\newcommand{\icmlNotice}{}
\printAffiliationsAndNotice{\icmlNotice}  

\begin{abstract}

Diagnosing deep neural networks (DNNs) by analyzing the eigenspectrum of their weights has been an active area of research in recent years.
One of the main approaches involves measuring the \emph{heavytailness} of the empirical spectral densities (ESDs) of weight matrices.
This analysis has been shown to provide insights to help diagnose whether a model is well-trained or undertrained, and has been used to guide training methods involving layer-wise hyperparameter assignment.
In this paper, we address an often-overlooked challenge in estimating the heavytailness of these ESDs: the impact of the aspect ratio of weight matrices. 
We demonstrate that matrices of varying sizes (and aspect ratios) introduce a non-negligible bias in estimating the heavytailness of ESDs, leading to inaccurate model diagnosis and layer-wise hyperparameter assignment.
To overcome this challenge, we propose \ourmethod (\textbf{F}ixed-\textbf{A}spect-\textbf{R}atio \textbf{M}atrix \textbf{S}ubsampling), a method that normalizes the weight matrices by subsampling submatrices with a fixed aspect ratio.
Instead of measuring the heavytailness of the original ESD, we measure the average ESD of these subsampled submatrices.
We show that this method effectively mitigates the aspect ratio bias.
We validate our approach across various optimization techniques and application domains that involve eigenspectrum analysis of weights, including image classification in computer vision (CV) models, scientific machine learning (SciML) model training, and large language model (LLM) pruning. Our results show that despite its simplicity, \ourmethod uniformly improves the accuracy of eigenspectrum analysis while enabling more effective layer-wise hyperparameter assignment.
In one of the LLM pruning experiments, \ourmethod reduces the perplexity of the LLaMA-7B model by 17.3\% when compared with state-of-the-art methods.

\end{abstract}

\section{Introduction}

\begin{figure*}[!th]
    \centering
    \begin{subfigure}{\linewidth}
        \includegraphics[width=\linewidth]{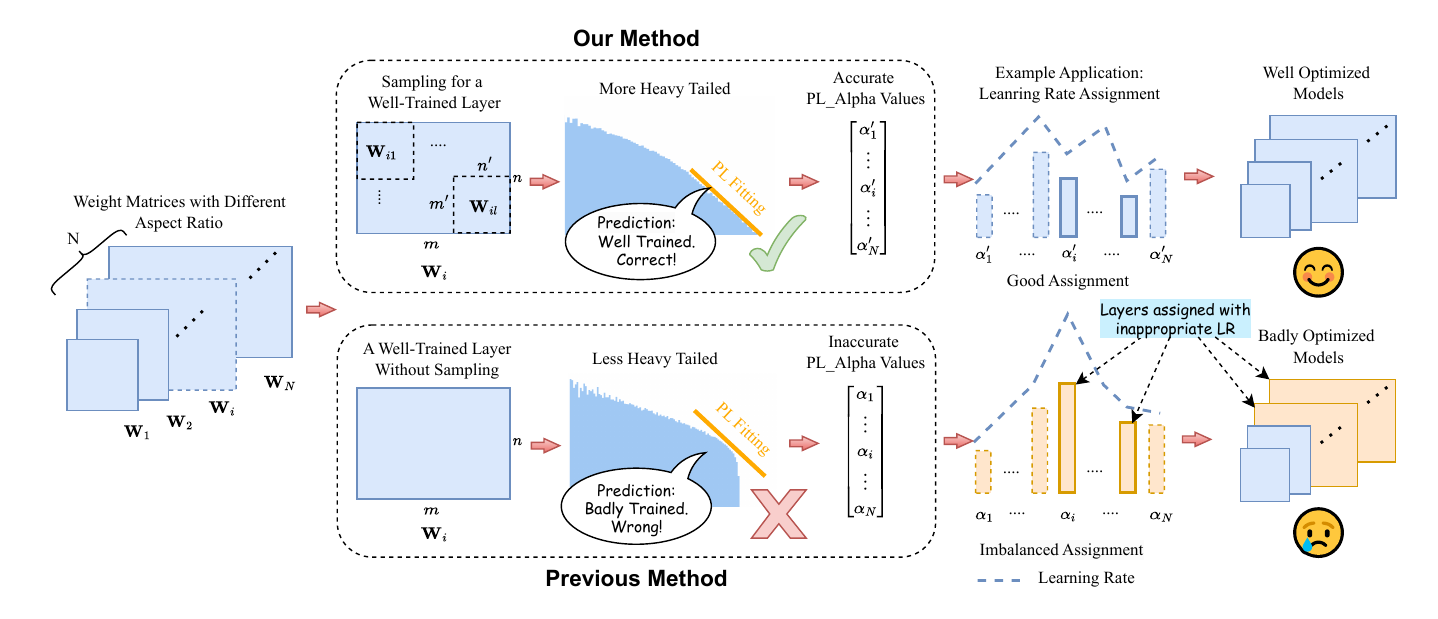}
    \end{subfigure}
    \caption{Comparing \ourmethod with common HT-SR methods in analyzing the weight matrices of a DNN. We use the weight matrix $\mathbf{W}_i$ of a well-trained layer with a large aspect ratio as an example (i.e., $\gamma=m/n$ is much larger than 1). 
    Due to the influence of the aspect ratio $\gamma=m/n \gg 1$, the ESD is more concentrated and less HTed than other layers.
    As a result, previous methods for fitting a power-law distribution (shown as ``PL Fitting'') tend to overestimate the heavytailness metric \Alpha (the HT index) and incorrectly suggest that the layer is poorly trained.
    Due to this aspect ratio bias, some optimization methods (such as \TB) cannot accurately assign per-layer hyperparameters, leading to suboptimal training.
    In contrast, our method, \ourmethod, which conducts ESD analysis with a fixed aspect ratio, accurately measures the training quality of the layer. See a concrete numerical example in Figure \ref{fig:final_layer_analysis_resnet_vgg} in Appendix \ref{appendix:matrix_analysis}.  } 
    \label{fig: method_introduction}
\end{figure*}

Random Matrix Theory (RMT) has long been used as a foundational tool in multiple research domains~\citep{guhr_random-matrix_1998, tulino_random_2004, bai2010spectral, tao2012topics, nadakuditi2012graph} and has now been applied to providing precise characterizations of neural networks (NNs)~\citep{pennington_nonlinear_2017,dobriban2018high, adlam2020neural, mei2022generalization, hastie2022surprises, hu2022universality, ba2022high, couillet2022rmt, derezinski2020exact}.
Among several emerging topics in this field, Heavy-Tailed Self-Regularization (HT-SR)~\citep{martin2021implicit, martin2019htsr,martin2021trends} Theory has been gaining significant attention. Unlike conventional RMT studies, HT-SR Theory focuses on studying weight matrices with strongly correlated elements, a typical characteristic of well-trained, practical DNNs. These strong correlations lead to heavy-tailed (HTed) ESDs~\citep{martin2021implicit}.
Analyzing these HT ESDs provides valuable insights into the quality of trained models, enabling both model diagnostics and improvements in model training.
For example,~\citet{martin2021implicit} show that one can locate undertrained layers by finding weight matrices with less HTed ESDs, which provides useful metrics for model selection and ranking~\citep{martin2021trends,yang2023generalization}.
Furthermore, one can use a larger learning rate on the undertrained layers~\citep{zhou2024temperature}, which provides an empirically successful training method to balance these layers with other, more well-trained ones. 
This technique has been developed into a general approach to improve test accuracy, training efficiency, and model interpretability in various applications, such as image classification~\citep{zhou2024temperature}, LLM model pruning~\citep{lu2024alphapruning}, and SciML model training and fine-tuning~\citep{liu2024model,liu2025lift}.

The core of the HT-SR theory lies in the measurement of the heavytailness of ESDs. In this context, we identified an often-overlooked issue in prior studies: weight matrices of different sizes are analyzed using the same HT measuring procedure, even though \emph{their ESDs can differ in theory}. Specifically, algorithms inspired by the HT-SR theory often begin by measuring the ESD of a weight matrix $\mathbf{W}$, that is, the empirical distribution of eigenvalues of the correlation matrix $\mathbf{W}^\top \mathbf{W}$. The shape of the ESD is then analyzed by fitting it to certain distribution families, such as the power-law (PL) distribution.
For weight matrices initialized using the i.i.d. random Gaussian elements, it is well-known that the ESD converges to the Marchenko–Pastur (MP) distribution as the dimensions of the matrix grow infinitely large in the proportional limit, i.e., when the number of rows $m$ and the number of columns $n$ increase but the ratio $m/n$ stays fixed.
The MP distribution depends explicitly on the aspect ratio $\gamma=m/n$, since this ratio determines the limiting bulk range, $[(1 - \sqrt{\gamma})^2, (1 + \sqrt{\gamma})^2]$, and it influences the overall shape of the ESD.
As a result, weight matrices of different aspect ratios exhibit distinct shapes in their ESDs, especially when they transition from the i.i.d. initialization to being correlated.
Comparing ESDs with different aspect ratios $\gamma$ directly, as is commonly done in prior work, overlooks this dependency on aspect ratio and can lead to inaccurate shape estimates.
Therefore, a careful consideration of the aspect ratio is critical when analyzing or comparing weight matrices, especially in eigenspectrum analysis.

Considering these challenges, we introduce \ourmethod (\textbf{F}ixed-\textbf{A}spect-\textbf{R}atio \textbf{M}atrix \textbf{S}ubsampling), a new method for measuring HT characteristics.
\ourmethod partitions each weight matrix into (overlapping) submatrices of fixed aspect ratios and then does an eigenspectrum analysis on the average ESD of these submatrices. This approach enables accurate computation of heavytailness metrics regardless of variations in matrix aspect ratios, ensuring a robust evaluation of layer quality across diverse matrix sizes that one can have in a DNN. 

To validate the effectiveness of \ourmethod, we conduct experiments across various application domains, including CV, SciML, and LLM pruning. In addition, these experiments are conducted on different parameter settings and model architectures. We compare \ourmethod with several prior HT-SR approaches that use weight eigenspectrum analysis for layer-wise hyperparameter assignment~\citep{zhou2024temperature,lu2024alphapruning,liu2024model}.
We also apply \ourmethod to measure various post-training and pruned models, making \ourmethod useful for model compression.
Our findings demonstrate that models optimized using \ourmethod exhibit lower mean and variation in HT-SR metrics across layers, a sign of good-quality training as reported in prior work~\citep{martin2021trends,liu2024model}. Our code is available \href{https://github.com/HUST-AI-HYZ/FARMS}{here}\footnote{https://github.com/HUST-AI-HYZ/FARMS}. Our key contributions are summarized below. 

\vspace{10pt}

\begin{itemize}
    \item \textbf{Mitigating Aspect Ratio Bias in Eigenspectrum Analysis}. \ourmethod addresses the aspect ratio bias of existing HT-SR eigenspectrum analysis, and it enables better computation of HT metrics on DNN models that have heterogeneous layer types.
    This improvement is achieved through a subsampling-based HT estimation method independent of the aspect ratio of weight matrices. In particular, we use a numerical example in Figure~\ref{fig:model_init_different_width}  and multiple real-data experiments in Section~\ref{wwsampling: cv_tempbalance} to convincingly demonstrate that \ourmethod mitigates the aspect ratio bias in training.

    \item \textbf{Improved Layer-wise Hyperparameter Tuning}. Since HT-SR Theory has been recently applied to various layer-wise hyperparameter tuning methods, \ourmethod thus improves these methods by offering a more accurate evaluation of HT metrics.
    In particular, we conduct experiments on DNN training and pruning across various model architectures. In CV models like ResNet and VGG, integrating \ourmethod into learning rate assignments yields improved accuracy compared to \TB, a recently proposed layer-wise learning rate assignment method~\citep{zhou2024temperature}.
    In LLMs pruning, \ourmethod improves \Alphapruning~\citep{lu2024alphapruning}, the SOTA method in assigning layer-wise pruning ratios.
    Specifically, \ourmethod can reduce the perplexity of the LLaMA-13B model from 2029.20 to 413.76 using the magnitude pruning method at a 0.7 sparsity ratio. As another example in LLaMA-7B, it reduces the perplexity from 96.02 to 79.42 using the SparseGPT pruning method~\citep{frantar2023sparsegpt} at a 0.8 sparsity ratio.
    In SciML, \ourmethod helps scientific models achieve up to a 5.66\% error reduction during fine-tuning compared to the HT-SR based method \TBSigmoid~\citep{zhou2024temperature}, even at relatively low L2 relative error (L2RE) levels.
\end{itemize}

\section{Related Work}

\subsection{Prior Work on HT-SR Theory} 
\label{wwsampling:htsr_related_work}
In this section, we provide an overview of prior work on HT-SR theory, a framework derived from RMT that is relevant to understanding modern, practical DNNs.
HT-SR theory~\citep{martin2021implicit, martin2019htsr} originates from RMT but extends well beyond it. The theory was proposed based on the observation that well-trained, state-of-the-art DNNs often exhibit HTed structures in the ESD of each layer. In the meantime, several rigorous theories in SGD relating HT phenomena to generalization performance were established, providing further theoretical support for HT-SR theory~\citep{hodgkinson2021multiplicative, hodgkinson2022generalization, gurbuzbalaban2021heavy, simsekli2019tail, simsekli2020hausdorff}.
Building on the theoretical foundation of HT-SR, a model analysis tool called WeightWatcher~\citep{martin2021implicit} was developed. Without accessing any training or test data, methods based on HT-SR Theory can be used to assess the training quality of models across various domains, such as CV and NLP~\citep{martin2021implicit, martin2021trends,yang2023generalization,liu2025lift}.

\subsection{Optimization Methods Based on Eigenspectrum Analysis} 

Our paper is mainly motivated by HT-SR theory, focusing on the eigenspectrum analysis of weight matrices.
Our paper connects to several DNN optimization methods that use eigenspectrum analysis to improve model performance.
For example, spectral norm regularization has been applied to improve generalizability of trained models~\citep{yoshida2017spectral,miyato2018spectral, farnia2018generalizable}. Stable rank normalization (SRN)~\citep{sanyal2020Stable}, a normalization technique, scales matrices using their stable rank to enhance training stability in GANs and generalization in DNNs.
TE-NAS~\citep{chen2021neural} is a training-free neural architecture search framework that identifies high-performing networks by analyzing the neural tangent kernel spectrum and input space linear~regions.

Although the aforementioned optimization methods based on eigenspectrum analysis have achieved a certain degree of model improvement, they do not provide fine-grained layer-wise optimization within DNNs. However, eigenspectrum analysis based on HT-SR theory offers a novel perspective by analyzing the shape of the HT ESDs, enabling a more precise estimate of the training quality for each layer.
\TB~\citep{zhou2024temperature} performs eigenspectrum analysis on ESDs of the weight matrices of DNNs to assess the training progress of each layer. It enables a balanced adjustment of the learning rate across layers during training—an important parameter that functions as a ``temperature-like’’ parameter within the language of statistical mechanics of learning. \Alphapruning~\citep{lu2024alphapruning}, on the other hand, uses a similar weight analysis method to evaluate the training quality of each layer in LLMs. It then uses this information to strategically allocate layer-wise sparsity ratios to maintain model performance after pruning. 

\newlength{\methodfigw}
\ificml
  \setlength{\methodfigw}{.98\linewidth}
\else
  \setlength{\methodfigw}{.66\linewidth}
\fi

\section{Method}
In this section, we first discuss the motivation behind our method \ourmethod, which is to address the aspect ratio bias of typical HT-SR methods. Then, we describe \ourmethod in detail and show how \ourmethod can reduce the aspect ratio bias.

\subsection{Typical HT-SR Analysis}\label{sec:typical_HT_analysis}

Before introducing \ourmethod, we first review the commonly adopted procedures in HT-SR weight analysis.
HT-SR analysis relies on estimating the layer quality based on the HT characteristic of the layer ESDs, which is quantified by HT metric \Alpha, introduced below. The ESD of a weight matrix is the histogram of eigenvalues. The ESD of weight matrices evolves during training, transitioning from a bulk-dominated regime to an HTed regime \cite{martin2021trends}. The HTed portion can be modeled by a power-law (PL) distribution within an interval ($\lambda_{\text{min}}$, $\lambda_{\text{max}}$):
\setlength{\abovedisplayskip}{10pt}
\setlength{\belowdisplayskip}{10pt}
\begin{equation}\label{eqn:ALPHA}
    p(\lambda) \propto \lambda^{-\alpha}, \lambda_{min} < \lambda < \lambda_{max},
\end{equation} 
where $\alpha$, the PL exponent, is a critical metric for analyzing training quality. To fit a PL distribution to the ESD, methods in HT-SR often use the Hill Estimator~\citep{hill1975simple,zhou2024temperature,lu2024alphapruning,liu2024model} \footnote{One important point to note is that estimating PLs is inherently a challenging task~\citep{clauset2009power}. Previous research suggests using the maximum likelihood estimate~\citep{alstott2014powerlaw,martin2021implicit}. However, empirical evidence indicates that the Hill estimator performs more reliably in DNN optimization applications~\citep{zhou2024temperature,lu2024alphapruning,liu2024model}.}. 
For the $i$-th layer, suppose the weight matrix is $\mathbf{W}_i$ and the correlation matrix $\mathbf{W}_i^\top\mathbf{W}_i$ has ascending eigenvalues $\lbrace \lambda_{i} \rbrace_{i=1}^n$. 
The Hill estimator calculates \AlphaHill (the PL exponent \Alpha estimated using the Hill estimator) as: 
\setlength{\abovedisplayskip}{10pt}
\setlength{\belowdisplayskip}{10pt}
\begin{equation}\label{eqn:hill_estimator}
\AlphaHill= 1+\frac{k}{\sum_{i=1}^k \ln\frac{\lambda_{n-i+1}}{\lambda_{n-k}}},
\end{equation}
where $k$ is an adjustable parameter. Changing $k$ essentially changes the lower eigenvalue threshold $\lambda_\text{min}$ for (truncated) PL estimation.
Various metrics have been proposed to analyze the properties of ESDs, among which shape metrics, which characterize the distributional shape of ESDs, have been shown to effectively predict the training quality of individual layers~\citep{yang2023generalization}. The \AlphaHill metric is one such shape metrics.

There are several metrics used to quantify the structure of ESDs in HT-SR theory. In this work, we mainly consider the \AlphaHill, which is empirically shown to be effective for training tasks~\citep{zhou2024temperature,lu2024alphapruning,liu2024model,qing2024alphalora}. In general, undertrained layers in DNNs tend to exhibit larger \AlphaHill values, whereas well-trained or over-trained layers typically have smaller \AlphaHill values. This metric can also be used for a comprehensive analysis across a series of DNN models with similar architectures to find the best model. Well-trained models tend to exhibit a lower average 
\AlphaHill across all layers~\citep{martin2021trends,yang2023generalization}. Moreover, in fine-tuning tasks, a larger amount of data generally results in a lower standard deviation (STD) of \AlphaHill across the model, indicating a more balanced training progression across different layers~\citep{liu2024model}.

\subsection{Rationale of \ourmethod: Why Typical HT-SR Methods Are Insufficient}

\begin{figure}[!h]
    \centering
        \includegraphics[width=\methodfigw]{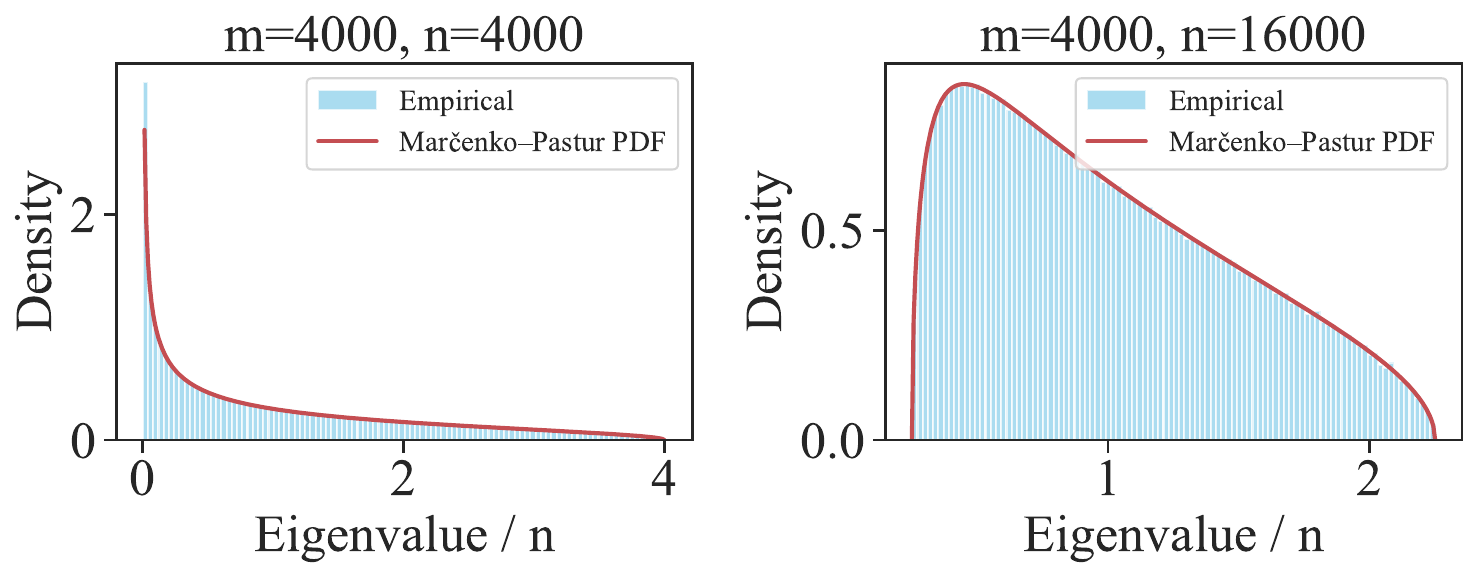}
    \caption{ESD shapes can be biased by the aspect ratio. Here we visualize the eigenvalues of the symmetric matrix $\frac{1}{n} \mathbf{X}_m \mathbf{X}_m^\top$. The left ESD is more HTed, while the right ESD is more concentrated. However, this change in the shape of the ESD is entirely due to the different aspect ratios of these two matrices and is not caused by differences in training quality. See detailed settings and supplementary results in the Figure \ref{fig:mp_law_different_mn} in Appendix \ref{appendix:mp_law}. } 
    \label{fig:mp_law_twocases}
\end{figure}

Here, we explain the rationale behind our method \ourmethod. In Section \ref{sec:typical_HT_analysis}, we mentioned that layers exhibiting more HTed ESDs tend to be more well-trained. This observation is the key to using HT-SR methods to measure model quality. However, beyond the training quality of weights, the aspect ratio of a weight matrix also influences the heavytailness of its ESD.
Specifically, in RMT, the Marchenko-Pastur (MP) distribution describes the limiting behavior of singular value distributions in large rectangular random matrices. According to the MP distribution, the ESD of the correlation matrix $\mathbf{W}^\top \mathbf{W}$ of an i.i.d. Gaussian random matrix $\mathbf{W}_{m \times n}$ becomes more concentrated as the aspect ratio $m/n$ deviates from 1.
An example is presented in Figure \ref{fig:mp_law_twocases}. Consequently, weight matrices with different aspect ratios naturally exhibit varying ESD shapes, leading to different degrees of heavytailness, which, if not analyzed carefully, can interfere with the quality assessment of model layers. We refer to this issue as the \emph{\textbf{aspect ratio bias}} in HT-SR methods.

As another example, consider a well-trained layer with a large aspect ratio, such as the final layer of a ResNet 18 model trained on the CIFAR 100 dataset. This weight matrix has size $512\times 100$ and a large aspect ratio of $512/100$. Thus, the ESD may not exhibit an obvious enough HT structure (See a related example in Figure \ref{fig:final_layer_analysis_resnet_vgg} in Appendix \ref{appendix:matrix_analysis}). If previous HT-SR methods are used, such layers may be quantified as poorly trained, but that conclusion results from the aspect ratio bias and is an inaccurate assessments of their training quality. Such misjudgment not only interferes with the evaluation of model performance but also leads to inaccurate tuning of hyperparameters in layer-wise optimization methods~\citep{zhou2024temperature,lu2024alphapruning}, as these methods all assign different layer-wise hyperparameters based on the HT estimates. Therefore, eliminating such aspect ratio bias is crucial for improving HT-SR eigenspectrum analysis.

\subsection{Analyzing ESDs Using \ourmethod}
\label{section:esd_farms}

To mitigate the aspect ratio bias in analyzing ESDs, we use a block-wise sampling method when processing weight matrices. We partition each weight matrix into overlapping sub-matrices with a fixed aspect ratio (across all layers) following a predefined scheme described below. 

\begin{figure}[t]
  \centering
  \includegraphics[width=\methodfigw]{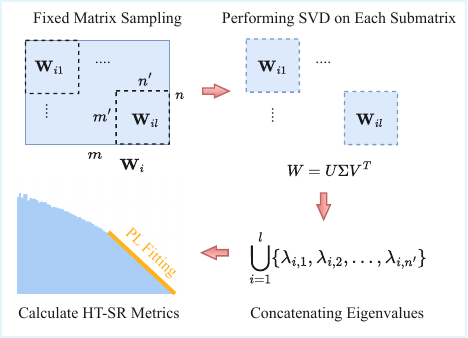}
  \caption{Main Steps in \ourmethod.}
  \label{fig:method_detailed}
\end{figure}

Consider the weight matrix $\mathbf{W}_i$ of the $i$-th layer, which has a shape of $m \times n$. Without loss of generality, consider the case when $m \geq n$. Note that the shape $m \times n$ changes across layers. To process this weight matrix, we apply a sliding window approach to partition it into several equally sized submatrices (potentially with overlap), denoted as $\mathbf{W}_{i1}, \mathbf{W}_{i2}, \dots, \mathbf{W}_{il}$, where $l$ represents the number of submatrices. Each submatrix has a shape of $m' \times n'$ and satisfies a fixed aspect ratio $Q=m'/n'$. The parameters $m'$, $n'$, $l$, and $Q$ are tunable hyperparameters (on which we will provide ablation studies in Section~\ref{wwsampling: ablation_study}), allowing flexibility in the partitioning strategy. The key requirement is that even if the aspect ratio of the whole weight matrix $m/n$ changes for different layer index $i$, that of the submatrices $Q = m'/n'$ is fixed across all layers.

Subsequently, for 2D Linear layers we compute the eigenvalues of the correlation matrices $\mathbf{X}_{ij}=\mathbf{W}_{ij}^\top \mathbf{W}_{ij}$ for each submatrix in the $i$-th layer. 
We then average the ESDs of $\mathbf{W}_{i1}, \mathbf{W}_{i2}, \dots, \mathbf{W}_{il}$ (by merging their corresponding eigenvalue series) and measure the HT metrics of the averaged ESD instead.
We merge the eigenvalue series because that's equivalent to averaging the empirical densities. The average of the ESDs leads to a less variant estimate of HT characteristics. 

For CNN layers, since they have 4D tensor weight matrices with four dimensions $[C_{1}, C_{2}, k_H, k_W]$ (which represent input channels, output channels, height, and width, respectively), we employ a slightly different method for subsampling and measuring.
We first flatten the two dimensions representing the convolution kernel size, $[k_H, k_W]$, in the 4D tensor into a single dimension. This results in $l' = k_H \times k_W$ two-dimensional matrices of shape $[C_{1}, C_{2}]$.
We assume here that $C_1 \geq C_2$ for convenience.
Then, for the 3D tensor of shape $l' \times C_{1} \times C_{2}$, we perform subsampling of size $m' \times n'$ on each $C_{1} \times C_{2}$ matrix corresponding to each $l'$. This results in $[\frac{C_{1}}{m'}] \times [\frac{C_{2}}{n'}] \times l'$ submatrices, each of size $m' \times n'$. Next, we average the ESDs by concatenating the rooted singular values for each submatrix along the $l'$ dimension, and then measure the HT metrics of the ESDs. In this way, we obtain $[\frac{C_{1}}{m'}] \times [\frac{C_{2}}{n'}]$ HT metrics values. Finally, we take the average of these metrics values to obtain the HT metrics corresponding to the original 2D CNN. We also considered concatenating all the rooted singular values into a single list and calculating the corresponding ESD to measure the degree of heavy-tailedness. However, our experimental results in Table~\ref{table:imgcls_cnn_methods_comp} in Appendix~\ref{appendix:cnn_methods_comp} show that averaging the results separately leads to better model performance. The detailed procedures of \ourmethod are shown in Figure~\ref{fig:method_detailed}.

\newlength{\methodtablew}
\ificml
  \setlength{\methodtablew}{.98\linewidth}
\else
  \setlength{\methodtablew}{.60\linewidth}
\fi

\section{Empirical Results}
\label{section:empirical_results}

In this section, we apply \ourmethod to measure HT metrics. To demonstrate the effectiveness of the new approach, we use these metrics across various optimization methods and tasks from different machine learning subfields.

In Section \ref{wwsampling: exp_setup}, we give full details of the experimental setup.
In Section \ref{wwsampling: LLM_Pruning}, we applied \ourmethod to LLM layer-wise pruning. In Section \ref{wwsampling: cv_tempbalance}, we employ \ourmethod on training VGGs and ResNets on image classification tasks. In Section \ref{wwsampling: sciml_tb_sig}, we validate that \ourmethod achieves better performance in SciML fine-tuning experiments. In Section \ref{wwsampling: HT_metrics_analysis}, we apply \ourmethod to analyze model quality, such as the balance of different layers. Finally, we perform ablation studies in Section  \ref{wwsampling: ablation_study}.

\begin{table*}[!thb]
    \centering
    \caption{WikiText validation perplexity for pruned LLaMA-7B and LLaMA-13B models at different sparsity settings. Our method is compared to \Alphapruning, each paired with magnitude based pruning, Wanda, and SparseGPT. Lower perplexity indicates improved model performance. For calculating the standard deviation(STD) and demonstrating the stability of our method, we sample different calibration sets with 128 samples using six different seeds [0, 1, 2, 3, 4, 5] in Wanda and SparseGPT.}  \vspace{0.25cm}
    \resizebox{0.95\linewidth}{!}{
    \begin{tabular}{c|c|ccc|ccc}
        \toprule
        \bf{Sparsity Ratio} & \bf{Layer-wise} & \multicolumn{3}{c|}{\bf{LLaMA-7B}}  & \multicolumn{3}{c}{\bf{LLaMA-13B}} \\
        
        & \bf{Method} & \bf{Magnitude} & \bf{Wanda} & \bf{SparseGPT} & \bf{Magnitude} & \bf{Wanda} & \bf{SparseGPT} \\
        
        \midrule
        0.7 
        & AlphaPruning & 231.76  & 24.30{\scriptsize$\pm$ 0.25}  & 18.66{\scriptsize$\pm$ 0.49}  & 2029.20  & 14.47{\scriptsize$\pm$ 0.08}  & 13.29{\scriptsize$\pm$ 0.17} \\
        \rowcolor{LightGrey} 
        & Ours & \textbf{173.49}  & \textbf{22.61{\scriptsize$\pm$ 0.18}}  & \textbf{18.53{\scriptsize$\pm$ 0.40}} & \textbf{413.76}  & \textbf{14.20{\scriptsize$\pm$ 0.09}}  & \textbf{13.06{\scriptsize$\pm$ 0.14}} \\
        
        \midrule
        0.75 
        & AlphaPruning & 2046.22  & 104.53{\scriptsize$\pm$4.49}  & 36.52{\scriptsize$\pm$1.13} & 2710.49  & 32.18{\scriptsize$\pm$0.31}  & 22.26{\scriptsize$\pm$0.59} \\
        \rowcolor{LightGrey} 
        & Ours & \textbf{1704.56}  & \textbf{71.67{\scriptsize$\pm$1.71}} & \textbf{35.47{\scriptsize$\pm$1.01}} & \textbf{2634.82}  & \textbf{29.56{\scriptsize$\pm$0.33}}  & \textbf{20.80{\scriptsize$\pm$0.43}} \\
        
        \midrule
        0.8 
        & AlphaPruning & 28865.67 & 772.20{\scriptsize$\pm$78.70}  & 96.02{\scriptsize$\pm$1.59} & 5399.87  & 160.59{\scriptsize$\pm$4.05}  & 47.57{\scriptsize$\pm$2.64} \\
        \rowcolor{LightGrey} 
        & Ours & \textbf{12799.58}  & \textbf{504.58{\scriptsize$\pm$23.05}}  & \textbf{79.42{\scriptsize$\pm$3.86}} & \textbf{5026.86}  & \textbf{127.49{\scriptsize$\pm$2.12}}  & \textbf{41.44{\scriptsize$\pm$1.58}}\\
        
        \midrule
        0.85 
        & AlphaPruning & 71710.96 & 4609.70{\scriptsize$\pm$978.39}  & 272.84{\scriptsize$\pm$30.84} & 38140.95  & 3144.01{\scriptsize$\pm$597.79} & 122.38{\scriptsize$\pm$8.88} \\
        \rowcolor{LightGrey} 
        & Ours & \textbf{66808.51} & \textbf{3595.54{\scriptsize$\pm$810.54}} & \textbf{234.46{\scriptsize$\pm$16.42}} & \textbf{37453.06}  & \textbf{2847.85{\scriptsize$\pm$368.10}}  & \textbf{101.06{\scriptsize$\pm$4.48}} \\
        \bottomrule
    \end{tabular}
    }
    \label{table:llama_pruning_wiki}
\end{table*}

\subsection{Experimental Setup}
\label{wwsampling: exp_setup}

\textbf{Datasets.} For image classification, we consider the CIFAR100 dataset~\citep{krizhevsky2012learning}. CIFAR100 consists of 50K pictures for the training set and 10K pictures for the testing set with 100 categories. For evaluating LLM pruning methods, we calculate model perplexity on the held-out WikiText~\citep{merity2017pointer} validation set and use seven tasks, including BoolQ~\citep{clark2019boolq}, RTE~\citep{wang2018glue}, HellaSwag~\citep{zellers2019hellaswag}, WinoGrande~\citep{sakaguchi2021winogrande}, ARC Easy and Challenge~\citep{clark2018think} and OpenbookQA~\citep{mihaylov2018can} for downstream zero-shot evaluation~\citep{gao2021framework}. For SciML, we fine-tune the models on simulated solutions of time-dependent PDE dataset 2D Compressible Navier-Stokes (CFD) \footnote{CFD means compressible fluid dynamics or, equivalently, the compressible Navier-Stokes equations.} from PDEBench~\citep{takamoto2022pdebench}. All datasets considered in this paper are standard and widely studied.

\textbf{Models.} We consider different types of NNs in various research fields:
VGG~\citep{simonyan2015very}, ResNet~\citep{he2016deep}, OPT~\citep{zhang2022opt}, LLaMA~\citep{touvron2023llama, touvron2023llamav2,grattafiori2024llama} and DPOT~\citep{hao2024dpot}. For VGG series, we consider VGG16 and VGG19. For ResNet series, we consider ResNet18 and ResNet34. For OPT series, we consider four different model size: OPT-125M/350M/1.3B/6.7B. For LLaMA series, we consider six different moodels: LLaMA-7B/13B, LLaMA-V2-7B/13B and LLaMA-V3-3B/8B. For DPOT series, we consider DPOT-Tiny and DPOT-Small. 

\textbf{Baseline.} We considered several model diagnostic and layer-wise hyperparameter scheduling methods as baselines for comparison.
\TB ~\citep{zhou2024temperature} is a layer-wise learning rate allocation algorithm based on HT-SR theory, designed to analyze the training progress of each layer and allocate learning rates accordingly. \TB measures the HT metrics of all layers, and it assigns a larger learning rate to weight matrices with a more lighted-tailed ESD.
\Alphapruning ~\citep{lu2024alphapruning}, on the other hand, is a layer-wise pruning method for LLMs based on HT-SR. It assigns a larger pruning ratio to Transformer weight matrices with a more lighted-tailed ESD.
Additionally, we compared our method with the \TBSigmoid ~\citep{liu2024model} approach, which was applied to experiments on SciML models. All three optimization methods adopt ESD analysis techniques, which are susceptible to aspect ratio bias because the HT metrics are measured on the whole weight matrix. Therefore, one can replace the HT measurement procedures in these methods with \ourmethod. The goal is to evaluate the improvement of \ourmethod compared to the existing HT-SR analysis used in these optimization methods.

\subsection{Improving LLM Pruning with \ourmethod}
\label{wwsampling: LLM_Pruning}

To explore the effectiveness of \ourmethod, we conduct experiments in the task of LLM pruning.
As we have mentioned, we can replace the HT measurement procedures in \Alphapruning with \ourmethod. Therefore, we can compare \Alphapruning with \ourmethod and without. We make the comparison with different sparsity ratios in the range \{0.7, 0.75, 0.8, 0.85\} and different LLM pruning methods, including Magnitude-based~\citep{NIPS2015_ae0eb3ee}, SparseGPT~\citep{frantar2023sparsegpt} and Wanda~\citep{sun2023simple}.
We use different LLM pruning methods because \Alphapruning assigns only layer-wise ratios, not specific pruning locations, and thus can be applied to various LLM pruning methods.  We followed the experimental setup from~\citet{lu2024alphapruning} and our hyperparameter ranges are reported in Table \ref{table:llama_dpot_hyper} of Appendix~\ref{appendix:Hyperparamter_settings}.

\begin{table*}
    \centering
     \caption{Comparison of mean zero-shot accuracies (\%) for pruned LLaMA-7B and LLaMA-13B models at different sparsity settings. We evaluate our method against \Alphapruning, each integrated with magnitude-based pruning, Wanda, and SparseGPT. Higher accuracy values indicate better zero-shot ability.}
    \vspace{0.25cm}
    \resizebox{0.88\linewidth}{!}{  
    \begin{tabular}{c|c|ccc|ccc}
        \toprule
        \bf{Sparsity Ratio} & \bf{Layer-wise}  & \multicolumn{3}{c|}{\bf{LLaMA-7B}}  & \multicolumn{3}{c}{\bf{LLaMA-13B}} \\
        
        &  \bf{Method} & \bf{Magnitude} & \bf{Wanda} & \bf{SparseGPT} & \bf{Magnitude} & \bf{Wanda} & \bf{SparseGPT} \\
        
        \midrule
        0.7 
        & AlphaPruning   & 35.67  & 43.67  & 44.79 & 38.23 & 47.46  & 49.07 \\
        \rowcolor{LightGrey} 
        & Ours & \textbf{35.96} & \textbf{44.26} & \textbf{44.84} & \textbf{38.87} & \textbf{47.75}  & \textbf{49.61} \\

        \midrule
        0.75
        & AlphaPruning   & 34.59 & 37.99  & 40.89 & \textbf{38.16} & \textbf{42.73}  & 44.17 \\
        \rowcolor{LightGrey} 
        & Ours & \textbf{35.88} & \textbf{39.66} & \textbf{40.93}  & 37.68 & 42.66  & \textbf{44.47} \\

        \midrule
        0.8
        & AlphaPruning  & 33.79 & 34.06 & 36.63 & 35.59 & 38.42  & 39.07 \\
        \rowcolor{LightGrey} 
        & Ours & \textbf{34.33} & \textbf{35.76} & \textbf{37.50} & \textbf{36.94} &  \textbf{39.23} & \textbf{40.18} \\

        \midrule
        0.85
        & AlphaPruning  & 33.29 & 31.69 & 34.86 & 33.11 &  32.05 & 36.73 \\
        \rowcolor{LightGrey} 
        & Ours  & \textbf{34.27} & \textbf{33.09} & \textbf{35.25} & \textbf{33.29} & \textbf{32.41}  & \textbf{37.04} \\

        \bottomrule
    \end{tabular}
    }
    \label{table:llama_pruning_zeroshot_mean}
\end{table*}
 
\paragraph{Language Modeling.} The results in Table \ref{table:llama_pruning_wiki} illustrate that using \ourmethod helps \Alphapruning achieve better performance under different sparsity ratios and LLM pruning methods. Specifically, when using the Magnitude method, our approach reduces the perplexity of LLaMA-7B from 231.76 to 173.49 at a sparsity ratio of 0.7. Similarly, when employing more advanced pruning methods such as Wanda and SparseGPT, our approach further reduces the perplexity of LLaMA-7B from 96.02 to 79.42 and LLaMA-13B from 47.57 to 41.44 at a sparsity ratio of 0.8.
One notable advantage of \Alphapruning is that it allows pruning LLMs to a sparsity ratio of 0.8 without significantly impacting perplexity. With \ourmethod, this performance can be further improved. We can also find that \ourmethod achieves lower perplexity STD in most settings. In Table \ref{table:opt_pruning_wiki} and Table \ref{table:llamav2v3_pruning_wiki} in Appendix \ref{appendix:more_results}, we provide additional experiment results on OPT series models and more recent LLaMA models.

\paragraph{Zero-Shot Tasks.} We test the zero-shot ability of pruned LLMs on seven zero-shot downstream tasks mentioned in Section \ref{wwsampling: exp_setup} with prompting. Results are shown in Table~\ref{table:llama_pruning_zeroshot_mean}, where we report the mean zero-shot accuracy on seven zero-shot tasks of pruned LLaMA-7B and LLaMA-13B models. \ourmethod can outperform \Alphapruning in improving accuracy across most settings. For example, although \Alphapruning achieved significant improvements over the uniform method, \ourmethod further improved accuracy by values of 1.11\% in SparseGPT pruning over \Alphapruning at a sparsity ratio of 0.8. We report the detailed performance for each individual task in Table \ref{table:llama7b_pruning_zeroshot_details} and Table \ref{table:llama13b_pruning_zeroshot_details} in Appendix \ref{appendix:more_results}.

\subsection{Improving Image Classification with \ourmethod}
\label{wwsampling: cv_tempbalance}

\begin{figure*}[!tbh]
    \centering
    \begin{subfigure}{0.66\linewidth}
        \includegraphics[width=\linewidth]{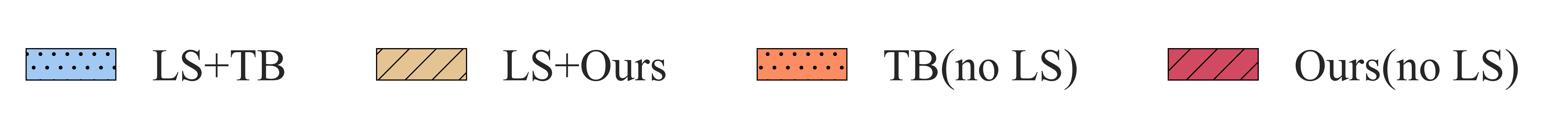}
    \end{subfigure}

    \begin{subfigure}{0.24\linewidth}
        \includegraphics[width=\linewidth]{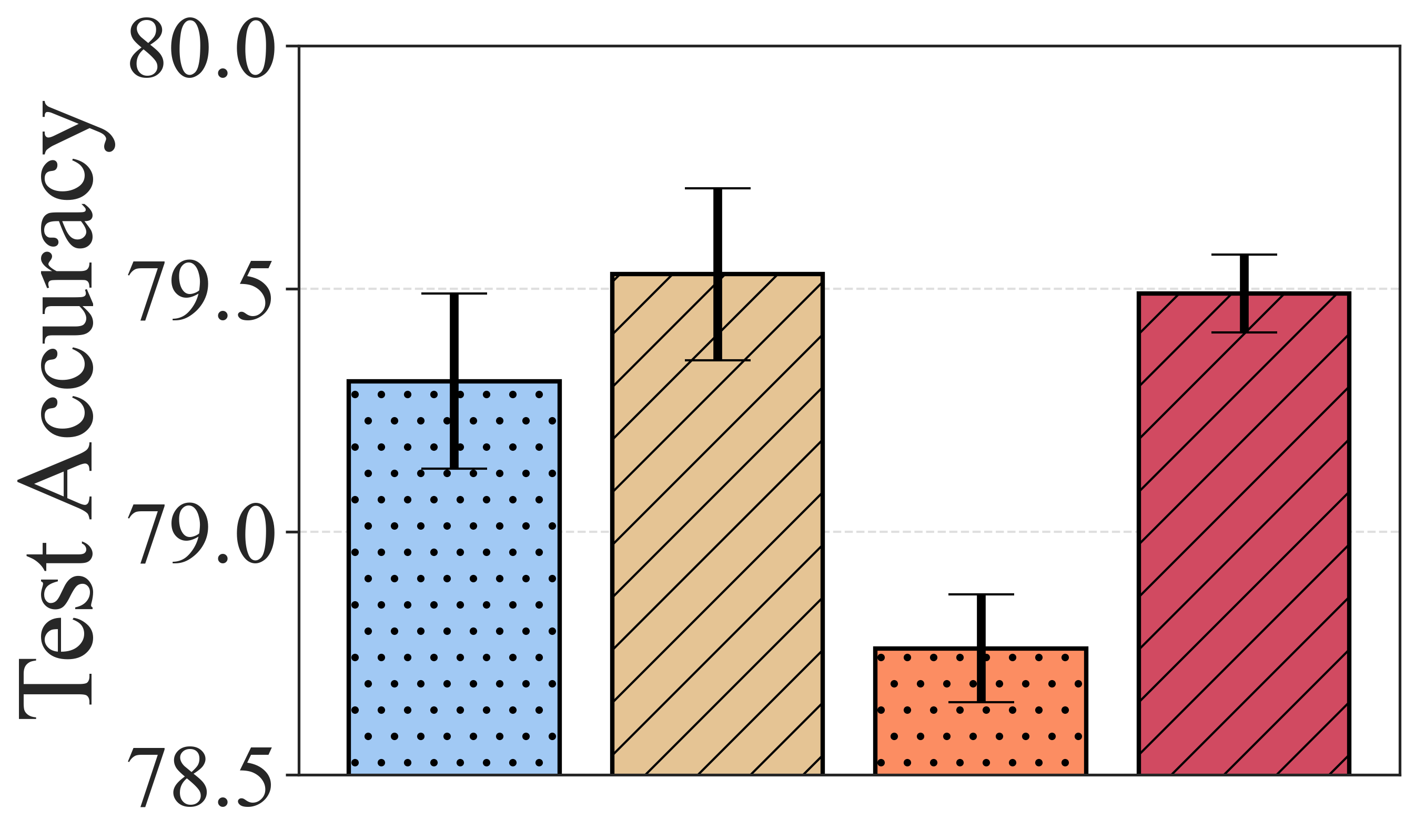}
        \caption{ResNet 18}
    \end{subfigure}
    \begin{subfigure}{0.24\linewidth}
        \includegraphics[width=\textwidth]{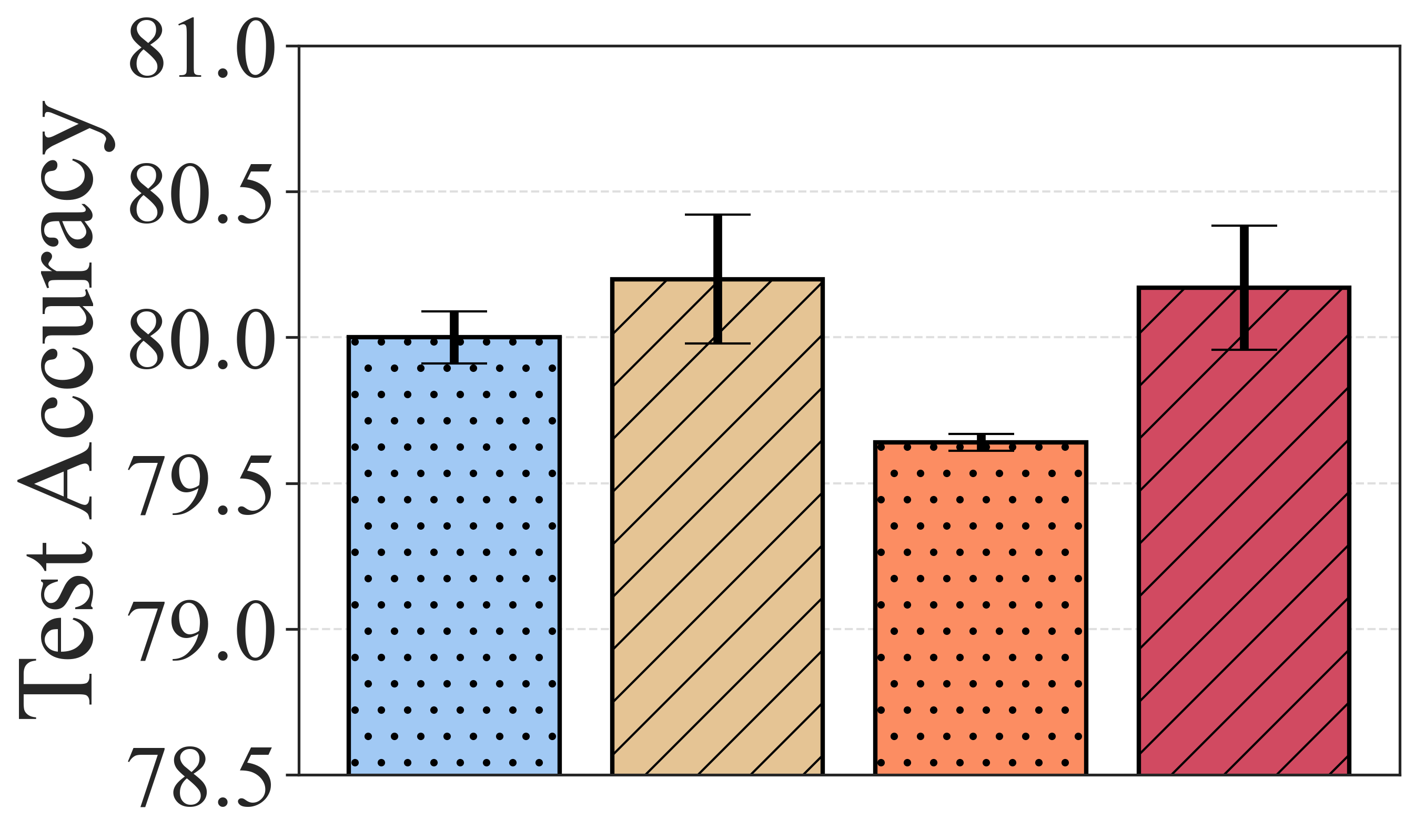}
        \caption{ResNet 34}
    \end{subfigure}
    \begin{subfigure}{0.24\linewidth}
        \includegraphics[width=\linewidth]{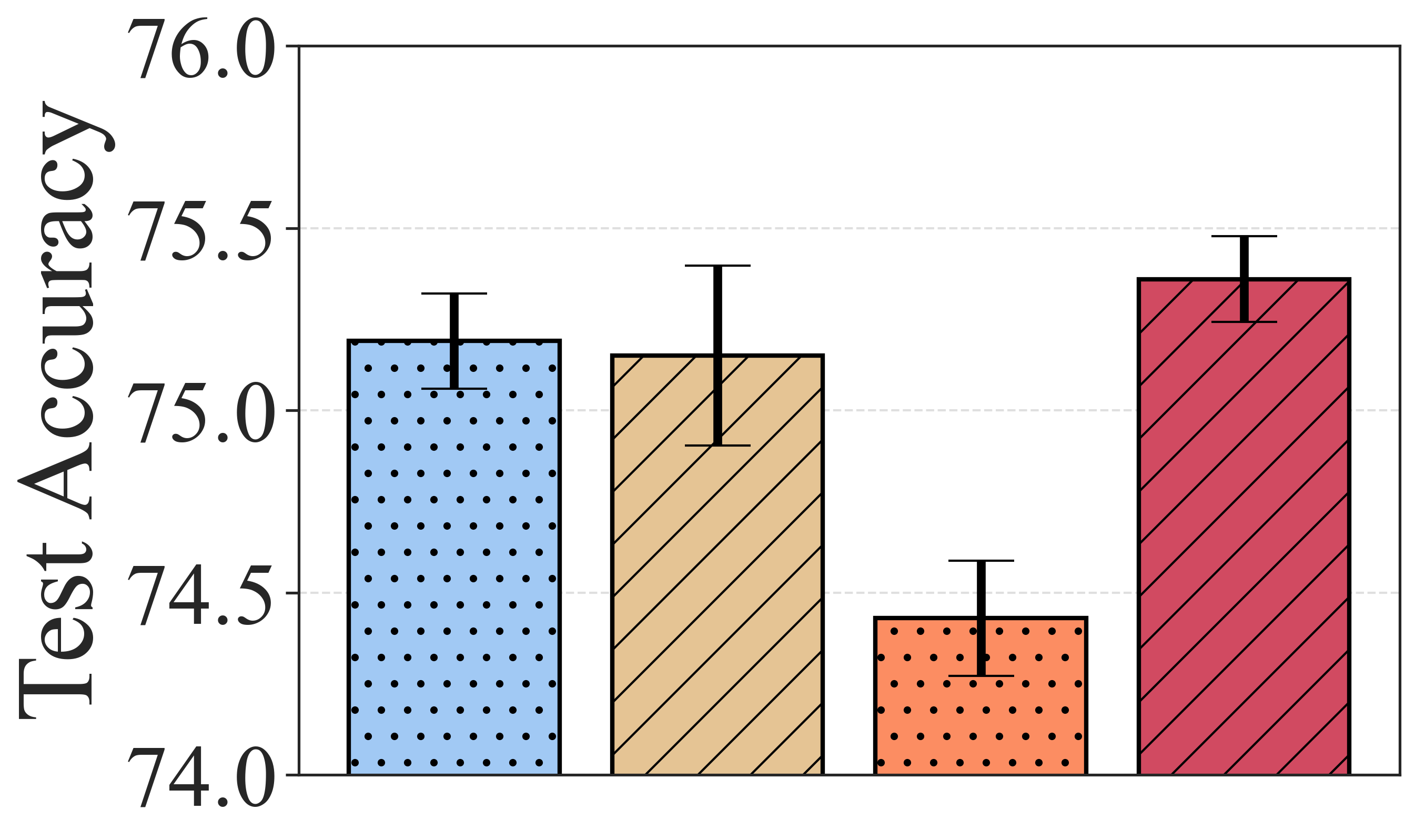}
        \caption{VGG 16}
    \end{subfigure}
    \begin{subfigure}{0.24\linewidth}
        \includegraphics[width=\textwidth]{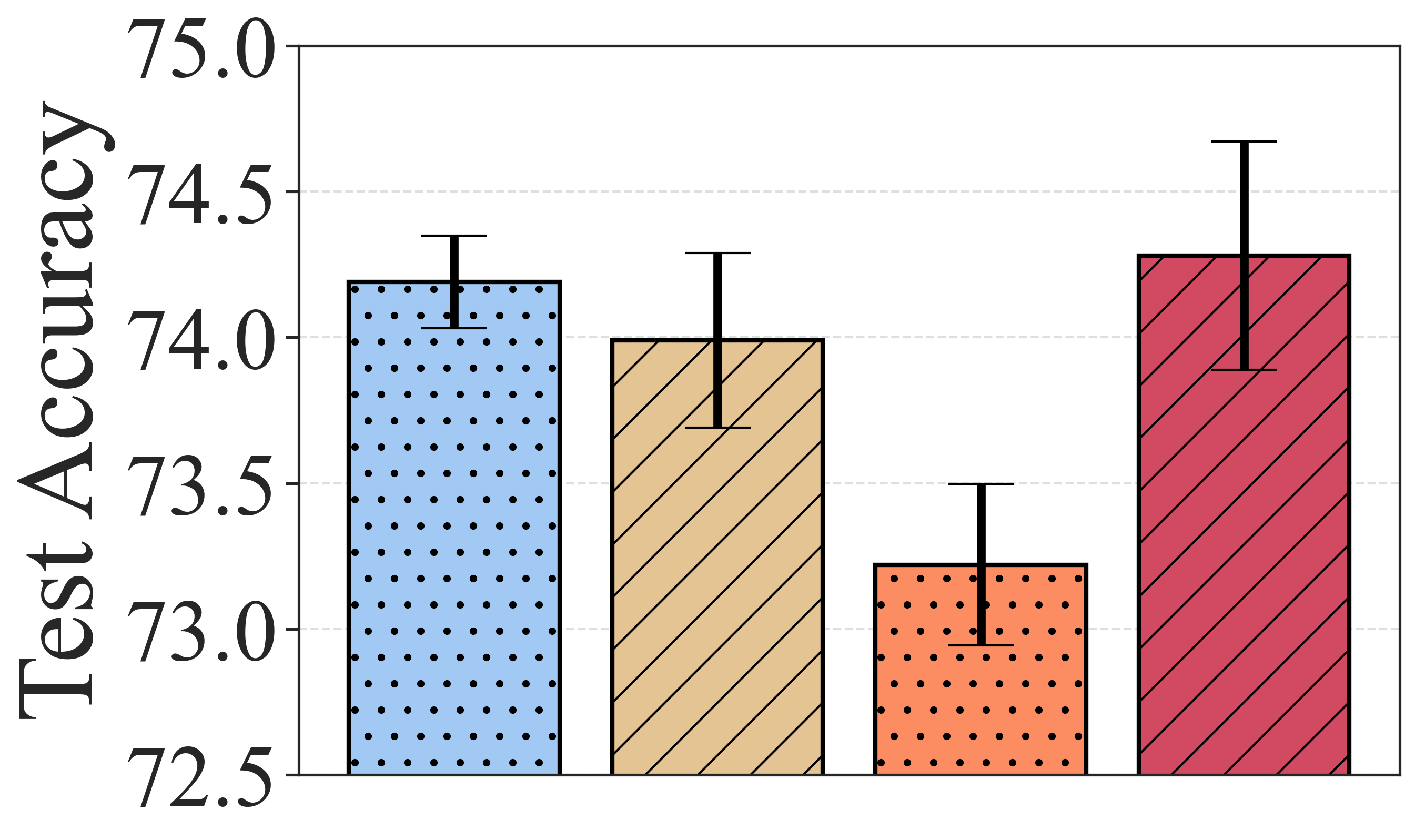}
        \caption{VGG 19}
    \end{subfigure}
    
    \caption{Comparing our method to \TB at different layer usage settings in training ResNet and VGG series models on CIFAR100. Higher test accuracy values indicate better model performance. See Table \ref{table:resnet_vgg_hyper} in Appendix \ref{appendix:Hyperparamter_settings} for the details in all hyperparameters.}
    \label{fig: imgcls_main_resnet_vgg_cifar100}
\end{figure*}

In this subsection, we compare \ourmethod to \TB on image classification tasks. As we mentioned, \TB is a layer-wise learning rate assignment method using HT metrics.
In Figure \ref{fig:  imgcls_main_resnet_vgg_cifar100}, we report the evaluation results of training the ResNet and VGG series models under different optimizing settings.
Again, we compare \TB with or without \ourmethod. ``Ours'' means replacing the method to measure heavytailness previously employed in the \TB with \ourmethod.

An important configuration used in the \TB method is the exclusion of certain layers from the learning rate assignment process.
Instead of receiving layer-wise learning rate adjustments as in \TB, these layers are assigned a global learning rate that decays following the cosine annealing schedule.
These excluded layers have ``tall-and-skinny matrices,'' which are matrices with a large aspect ratio and a relatively small number of eigenvalues. For example, the final layer of the ResNet 18 model has dimensions of $512 \times 100$, resulting in a large aspect ratio of $512/100=5.12$. This kind of extreme size makes it difficult for previous methods to accurately estimate the heavytailness of the ESDs. 
This issue further exacerbates the inability of the previous \TB method to accurately allocate per-layer learning rates, resulting in certain layers remaining at either excessively high or extremely low learning rates. We illustrate this phenomenon in more detail in Figure~\ref{fig:alpha_lr_change_byepochs_acrosslayers} of Appendix \ref{appendix:more_results}. Consequently, this imbalance leads to suboptimal model performance after training.

Therefore, according to the experimental setup described in \TB, the first and last layers of ResNet and VGG models, along with certain layers whose correlation matrix of weight matrix contains a relatively small number of eigenvalues, will be excluded from the layer-wise learning rate scheduling. 
To ensure a comprehensive study, we compare \ourmethod and \TB with and without this ``layer selection'' (LS) heuristic. 
See Figure \ref{fig:  imgcls_main_resnet_vgg_cifar100}, when the LS heuristic is not applied (and all layers will be included in the adjustable learning rate scheduling), the performance of the original \TB method (shown as ``TB (no LS)'') deteriorates significantly. For example, compared to the LS-enabled setup (shown as ``LS+TB''), the test accuracy of the VGG 19 model trained on CIFAR100 without LS using the \TB method drops from 74.19\% to 73.22\%, while the standard deviation increases from 0.159 to 0.277. In contrast, when training models using the \TB method with \ourmethod, performance remains stable or even improves when LS is disabled. This observation demonstrates that our method is more robust to weight matrices of extreme sizes, and the reason is that our method provides a more accurate assessment of the HTed ESDs independent of the matrix size.

In \TB algorithm, the layer-wise adjustable learning rate is scaled in the range of ($s_1\eta_t$, $s_2\eta_t$), where $\eta_t$ is the global learning rate at time $t$, and $s_1$ and $s_2$ are two tunable lower and upper limits of learning rates. To demonstrate that \ourmethod can more accurately evaluate the training quality of each model layer, we present the test performance of models trained using the different learning rate scaling ratios ($s_1$, $s_2$). in which we consider five different settings for ($s_1$, $s_2$): [(0.1, 1.9), (0.2, 1.8), (0.3, 1.7), (0.4, 1.6), (0.5, 1.5)]. Additionally, we consider that these five scaling ratios are generally close to the optimal scaling ratio. We run tasks on CIFAR100 with four VGG and ResNet architectures. Our results in Table \ref{table: imgcls_vggresnet_cifar100_scalingratios} show that \ourmethod can improve the test accuracy of models among the five scaling ratios and help models achieve a lower overall standard deviation compared to \TB. Detailed results can be found in Figure \ref{fig:lr_range_imgcls}, Appendix \ref{appendix:more_results}.

\begin{table}[!tbh]
    \centering
    \caption{Comparing our method to \TB among the five scaling ratios. All results are reported as the mean test accuracy and standard deviation obtained across five different scaling ratios. In this experiment, we allowed \TB baseline to use the LS heuristic to achieve slightly better test accuracy.} 
    \vspace{0.25cm}
    \resizebox{\methodtablew}{!}{
    \begin{tabular}{c|cc|cc}
        \toprule
        \bf{Method} & \bf{ResNet 18} & \bf{ResNet 34} &  \bf{VGG 16} &   \bf{VGG 19}\\
        \midrule
        
        TB  & 79.03{\scriptsize$\pm$0.169}  & 79.81{\scriptsize$\pm$0.145} 
            & 74.87{\scriptsize$\pm$0.214}  & 73.89{\scriptsize$\pm$0.199}  \\

        \rowcolor{LightGrey}
        Ours  & \bf{79.35{\scriptsize$\pm$0.126}}  & \bf{80.07{\scriptsize$\pm$0.097}} 
              & \bf{75.16{\scriptsize$\pm$0.212}}  & \bf{74.03{\scriptsize$\pm$0.163}}  \\
        
        \midrule
    \end{tabular}
    }
    \label{table: imgcls_vggresnet_cifar100_scalingratios} 
\end{table}

\subsection{Improving SciML Fine-tuning with \ourmethod}
\label{wwsampling: sciml_tb_sig}

To explore the potential applications of \ourmethod in multiple domains of machine learning research, we also performed SciML fine-tuning experiments using the 2DCFD dataset with DPOT-Tiny and DPOT-Small models and compared \ourmethod with the \TBSigmoid method~\citep{liu2024model}. We followed the experimental setup from~\citet{liu2024model}, and our hyperparameter ranges are detailed in Table \ref{table:llama_dpot_hyper} of Appendix \ref{appendix:Hyperparamter_settings}. In Table \ref{table:tb_nrmse_ft_sciml}, we show the results of comparing \ourmethod with the \TBSigmoid (shown as ``TB\_Sig'') with layer selection and the baseline fine-tuning (shown as ``FT'') with a uniform learning rate for each layer in different data subsampling ratios in the range \{5\%, 10\%, 25\%, 50\%, 100\%\}. Our method consistently outperformed \TBSigmoid across all subsampling ratios. For instance, with a full dataset (100\% of the data), our approach reduced the model's L2RE to 1.017e-2. Comparing to the \TBSigmoid, the L2RE decreased by 5.66\%.

\newlength{\methodtablewsicml}
\ificml
  \setlength{\methodtablewsicml}{.98\linewidth}
\else
  \setlength{\methodtablewsicml}{.50\linewidth}
\fi
\begin{table}[!tb]
    \centering
    \caption{\ourmethod achieves lower L2RE(↓) on the test dataset than \TBSigmoid and baseline fine-tuning method on SciML tasks.}
    \vspace{0.25cm}
    \resizebox{\methodtablewsicml}{!}{
    \begin{tabular}{c|c|cc}
        \toprule
        \bf{Subsampling} &  &  &  \\
        \bf{Ratio} & \bf{Method} & \bf{DPOT-Tiny} & \bf{DPOT-Small}  \\
        
        \midrule
        5\%  
        & FT       & 1.863e-2 & 1.546e-2 \\ 
        & TB\_Sig  & 1.856e-2 & 1.539e-2 \\
        \rowcolor{LightGrey} 
        & Ours     & \textbf{1.842e-2} & \textbf{1.536e-2} \\
        
        \midrule
        10\%  
        & FT       & 1.747e-2 & 1.426e-2 \\
        & TB\_Sig  & 1.730e-2 & 1.415e-2 \\
        \rowcolor{LightGrey} 
        & Ours     & \textbf{1.706e-2} & \textbf{1.407e-2} \\
        
        \midrule
        25\% 
        & FT       & 1.543e-2 & 1.226e-2 \\
        & TB\_Sig  & 1.517e-2 & 1.203e-2 \\
        \rowcolor{LightGrey} 
        & Ours     & \textbf{1.499e-2} & \textbf{1.189e-2} \\
        
        \midrule
        50\%  
        & FT       & 1.309e-2 & 1.025e-2 \\
        & TB\_Sig  & 1.283e-2 & 1.005e-2 \\
        \rowcolor{LightGrey} 
        & Ours     & \textbf{1.242e-2} & \textbf{9.822e-3} \\
        
        \midrule
        100\% 
        & FT       & 1.096e-2 & 8.400e-3 \\
        & TB\_Sig  & 1.078e-2 & 8.193e-3 \\
        \rowcolor{LightGrey} 
        & Ours     & \textbf{1.017e-2} & \textbf{7.949e-3} \\
        \bottomrule
    \end{tabular}
    }
     \label{table:tb_nrmse_ft_sciml} 
\end{table}

\begin{figure*}
    \centering
    \begin{subfigure}{0.2\linewidth}
        \includegraphics[width=\linewidth]{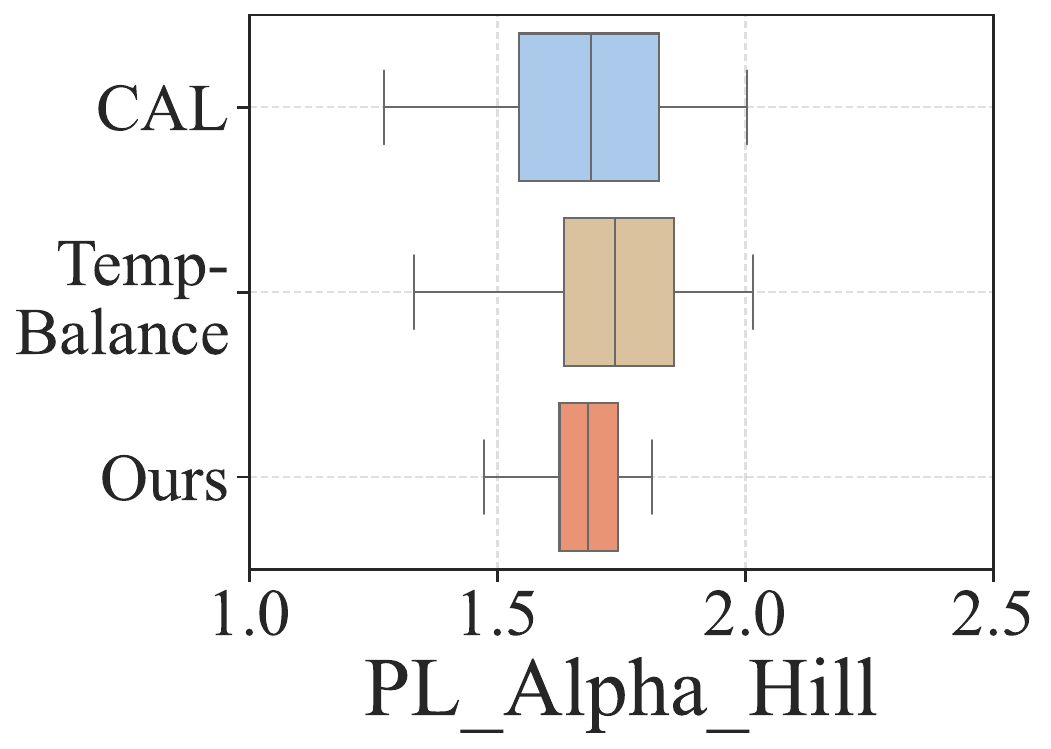}
        \caption{ResNet 18, Cifar100}
        \label{subfig:alpha_bar_resnet18}
    \end{subfigure}
    \begin{subfigure}{0.2\linewidth}
        \includegraphics[width=\textwidth]{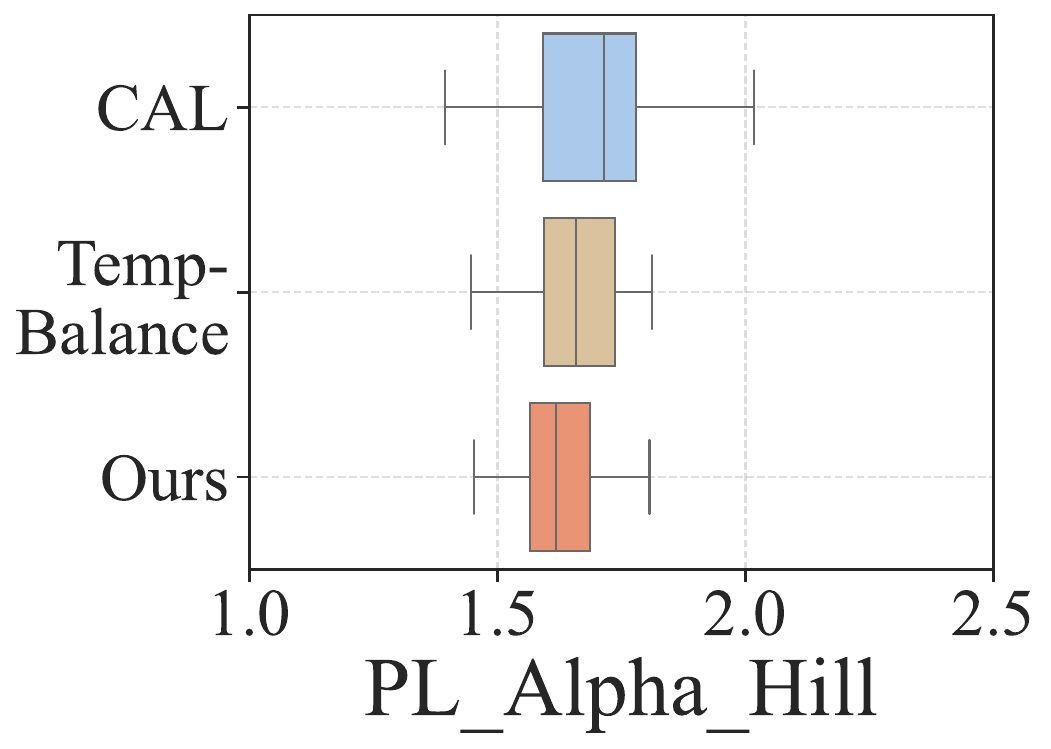}
        \caption{ResNet 34, Cifar100}
        \label{subfig:alpha_bar_resnet34}
    \end{subfigure}
    \begin{subfigure}{0.2\linewidth}
        \includegraphics[width=\linewidth]{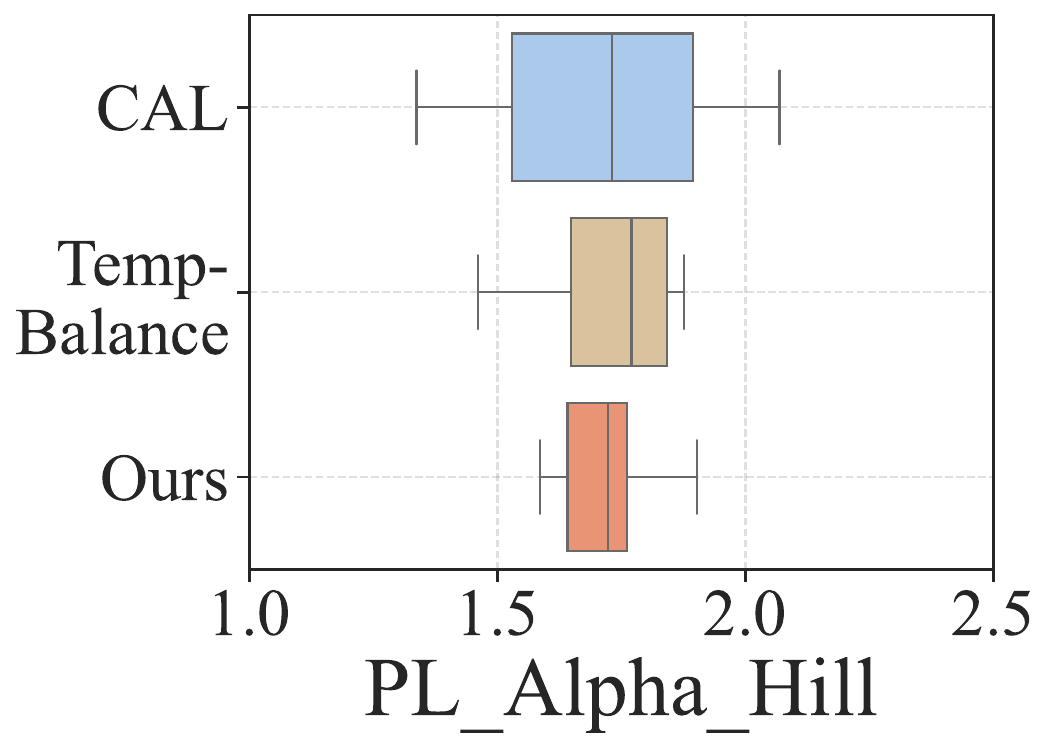}
        \caption{VGG 16, Cifar100}
        \label{subfig:alpha_bar_vgg16}
    \end{subfigure}
    \begin{subfigure}{0.2\linewidth}
        \includegraphics[width=\textwidth]{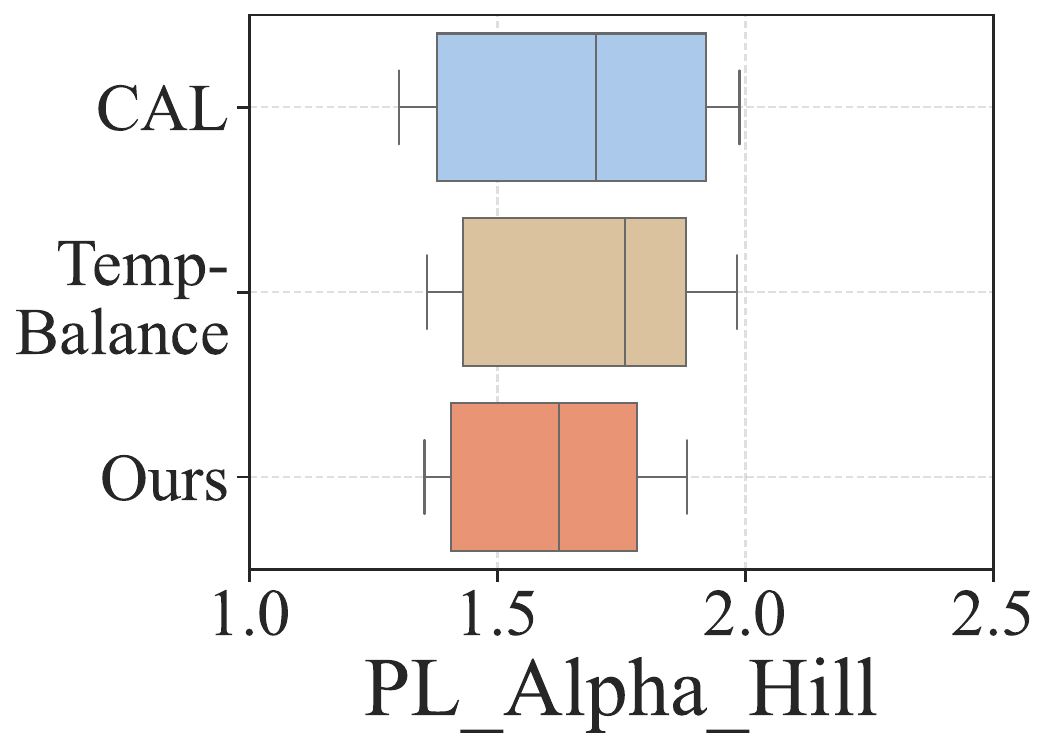}
        \caption{VGG 19, Cifar100}
        \label{subfig:alpha_bar_vgg19}
    \end{subfigure}

    \caption{Comparing the distribution of \AlphaHill of ResNet and VGG series models. The top blue-shaded section in each subplot of the experiments (shown as "CAL") represents models trained without employing layer-wise optimization methods. The middle brown-shaded section in each subplot indicates models trained using \TB; Lastly, the bottom orange-shaded section in each subplot corresponds to the results obtained using \ourmethod. Figures \ref{subfig:alpha_bar_resnet18}, \ref{subfig:alpha_bar_resnet34}, \ref{subfig:alpha_bar_vgg16} and \ref{subfig:alpha_bar_vgg19} present the averaged results obtained from multiple experiments within the optimal parameter range.} 
    \label{fig:alpha_bar_plot_main}
\end{figure*}

\subsection{HT Metrics Analysis}
\label{wwsampling: HT_metrics_analysis}

In this subsection, we demonstrate that \ourmethod can effectively control the shape of ESDs, resulting in a more balanced distribution of \AlphaHill values among the layers of NNs compared to the previous methods.
We assess the models presented in Section~\ref{wwsampling: cv_tempbalance}, which include 2D Convolution and 2D Linear layers of ResNets and VGG networks.
Results in Figure \ref{fig:alpha_bar_plot_main} show the distribution of \AlphaHill of models trained using baseline cosine annealing (CAL), \TB, and \TB using our method \ourmethod.
It can be seen that \ourmethod generally leads to a more concentrated distribution and, in most cases, reduces the average \AlphaHill value.
These experimental results suggest that our method \ourmethod enables a more balanced training progression across different layers of the model. 
These observations align well with the findings reported by~\citet{martin2021trends,liu2024model}, which suggest that a decrease in both average \AlphaHill values and the variance across layers indicate better training quality.

\subsection{Ablation Study}
\label{wwsampling: ablation_study}

\textbf{Different Aspect Ratios of Weight Matrix.} Here, we study the effect of the aspect ratio of the subsampled matrices in \ourmethod. We consider image classification tasks and use \ourmethod to replace the HT metrics in \TB for assigning layer-wise learning rates.
We consider five different aspect ratios in the range \{0.5, 0.75, 1.0, 1.5, 2.0\} and keep other hyperparameters optimal. We again use ResNet18 and VGG16 as our architectures and show results on CIFAR100. Results in Table \ref{table:ablation_q_ratios} show that \ourmethod achieves a higher test accuracy when the aspect ratio is 1.0.

\begin{table}[!h]
    \centering
    \caption{Comparing the different Aspect ratios settings for submatrices on ResNet 18 and VGG 16 models trained on CIFAR100 Dataset.} 
    \vspace{0.25cm}
    \resizebox{\methodtablew}{!}{
    \begin{tabular}{c|ccccc}
        \toprule
        \bf{Q Ratio} & 0.5 & 0.75 & \textbf{1.0} & 1.5 & 2.0 \\
        \midrule
         ResNet 18 & 
         79.13{\scriptsize$\pm$0.158} & 79.33{\scriptsize$\pm$0.122} &
         \bf{79.53{\scriptsize$\pm$0.177}} & 	79.13{\scriptsize$\pm$0.282} & 79.31{\scriptsize$\pm$0.046}  \\
         \midrule
         VGG 16 & 
         75.05{\scriptsize$\pm$0.285} &
         75.19{\scriptsize$\pm$0.517} &
         \bf{75.36{\scriptsize$\pm$0.118}} &
         75.21{\scriptsize$\pm$0.345} & 
         75.10{\scriptsize$\pm$0.360} \\
        \bottomrule
    \end{tabular}
    }
    \label{table:ablation_q_ratios}
\end{table}

\textbf{Subsampling Window Sizes and Sampling Steps.}
Here, we study different subsampling window sizes and sampling steps (i.e., the number of subsampled submatrices) in the task of LLaMA-7B pruning. The aspect ratio $Q$ is fixed to be 1.0. For all weight matrices in LLaMA-7B, the smallest dimension is 4096. Therefore, when using a window size of 4096, sliding window downsampling is applied only along the larger dimension. This process generates multiple $4096 \times 4096$ submatrices. We report our results in Table \ref{tb:window_size_step_size_ablation_study}. We find that using an appropriately sized sampling window, such as $2000 \times 2000$, the model achieves a lower perplexity. As a comparison, baseline perplexity achieved by \Alphapruning without using \ourmethod is 96.02.

\begin{table}[!h]
    \centering
    \caption{Different Window Size and Sampling Steps for pruning the LLaMA-7B at 0.8 sparsity ratio. The pruning method is SparseGPT. The model is evaluated on WikiText dataset with perplexity $(\downarrow)$.} 
    \vspace{0.25cm}
    \resizebox{\methodtablew}{!}{
    \begin{tabular}{c|ccccc}
        \toprule
         & \multicolumn{4}{c}{\bf{Sampling Steps}}  \\
        \midrule
        \bf{Window size} & 5 & 10 & 15 & 20 \\
        \midrule
         500  & 96.34{\scriptsize$\pm$9.15}  & 95.93{\scriptsize$\pm$2.80}  & 99.23{\scriptsize$\pm$3.53}  & 88.08{\scriptsize$\pm$4.23}   \\
         
         \midrule
         1000 & 84.71{\scriptsize$\pm$4.97}  & 81.96{\scriptsize$\pm$4.22}  & 89.20{\scriptsize$\pm$3.91}   & 84.04{\scriptsize$\pm$2.48}   \\

          \midrule
         2000 & 82.33{\scriptsize$\pm$3.18}  & \bf{79.42{\scriptsize$\pm$3.86}}  & 80.14{\scriptsize$\pm$2.32}   & 86.54{\scriptsize$\pm$3.28}  \\

          \midrule
         4096 & 82.63{\scriptsize$\pm$2.45}   & 89.61{\scriptsize$\pm$5.25}   &  84.19{\scriptsize$\pm$6.05}   & 84.07{\scriptsize$\pm$5.64}   \\
        \bottomrule
    \end{tabular}
    }
    \label{tb:window_size_step_size_ablation_study} 
\end{table}
\vspace{-5pt}
\section{Conclusion}

We propose a subsampling strategy to address the measurement bias introduced by varying aspect ratios of weight matrices in HT-SR eigenspectrum analysis. The main idea of our method is to extract submatrices of the original matrix with a fixed aspect ratio, followed by averaging the ESDs of these submatrices to accurately assess the training quality of each weight matrix. Our extensive experiments demonstrate that our method can precisely estimate layer-wise training quality, improve the performance of layer-wise optimization algorithms, and contribute to more balanced training and efficient pruning. Furthermore, results on visual analytics confirm the effectiveness of our approach, showing that it successfully eliminates the measurement bias caused by aspect ratio variations.

\clearpage

\section*{Impact Statement}

This paper presents research aimed at advancing the fields of Machine Learning and Deep Learning, particularly in training using information from eigenspectrum analysis. While our work has various potential societal implications, we do not find it necessary to highlight any specific ones~here.

\section*{Acknowledgments}
This work is supported by DOE under Award Number DE-SC0025584 and Dartmouth College.


\bibliography{main}
\bibliographystyle{ICML/icml2025}

\newpage
\appendix
\onecolumn
\section*{Appendix}

\section{Detailed Matrix Analysis}
\label{appendix:matrix_analysis}
\subsection{Random Initialization}

In this section, we do weight analysis for ResNet 34 and VGG 19 models that are randomly initialized using methods proposed in~\citet{he2015delving}. The weights \( w \) are drawn from a normal distribution with mean 0 and variance \( {2}/{n_{in}} \), where \( n_{in} \) represents the number of input neurons. This is represented as:
\setlength{\abovedisplayskip}{10pt}
\setlength{\belowdisplayskip}{10pt}
\[
w \sim \mathcal{N}\left(0, \frac{2}{n_{in}}\right)
\]

We present our results of comparing \ourmethod with the previous HT estimation methods in measuring randomly initialized models in Figure \ref{fig:model_init_different_width}. In Figure \ref{subfig:resnet34_alllayers} and Figure \ref{subfig:vgg19_alllayers}, all layers in these models should be similarly under-trained (because they are all just initialized). However, previous methods tend to report significantly higher values for \AlphaHill in certain layers, suggesting that these layers are much more under-trained. In contrast, \ourmethod produces results where the training quality of all initialized layers is nearly identical, aligning more closely with the expected result. Similarly, in Figure \ref{subfig:resnet34_diffwidth} and Figure \ref{subfig:vgg19_diffwidth}, for randomly initialized ResNet-34 and VGG-19 models with varying widths, the previous method measures \AlphaHill values that increase as model width increases. In contrast, \ourmethod shows \AlphaHill values that remain unaffected by the aspect ratio bias introduced by model width variations. 

\ificml
\begin{figure}[ht]
    \centering
    \begin{subfigure}{0.24\linewidth}
        \includegraphics[width=\linewidth]{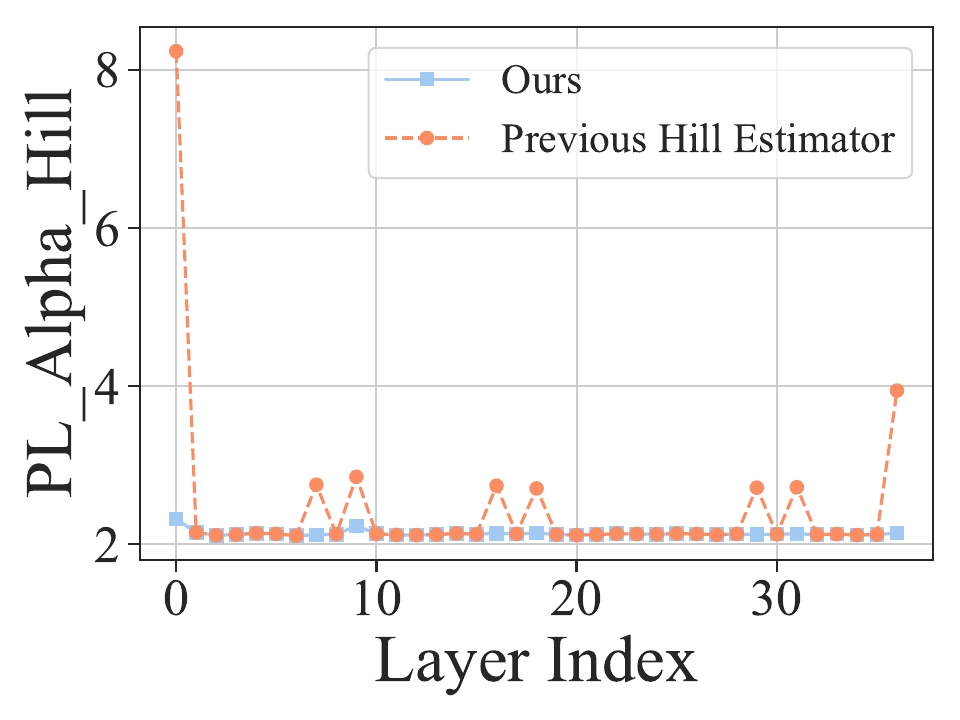}
        \caption{ResNet 34, All Layers}
        \label{subfig:resnet34_alllayers}
    \end{subfigure}
    \begin{subfigure}{0.24\linewidth}
        \includegraphics[width=\linewidth]{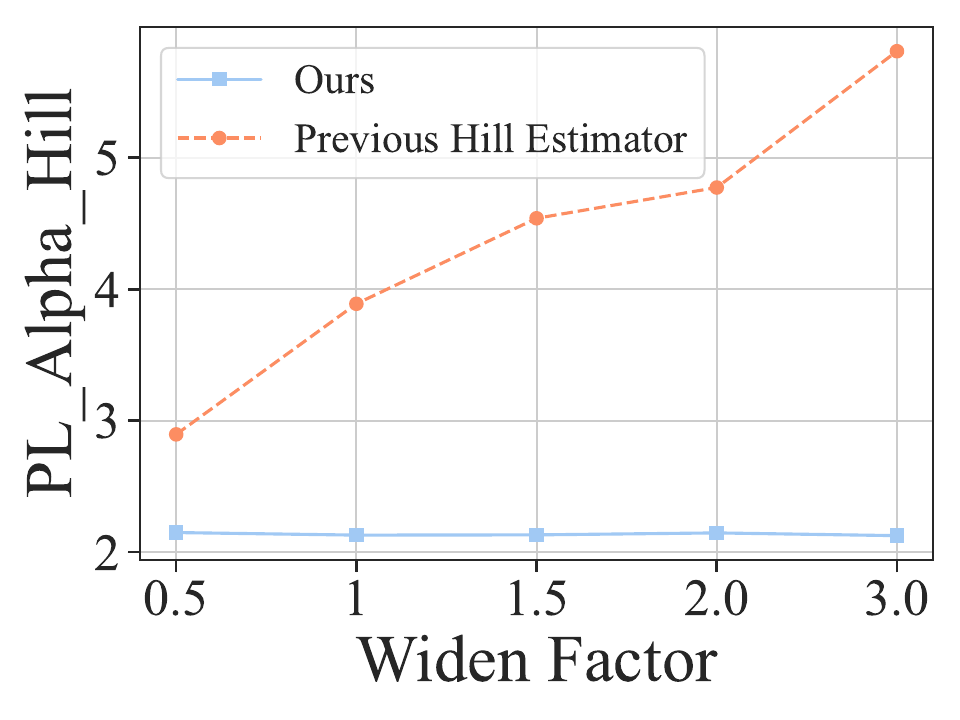}
        \caption{ResNet 34, Different Width}
        \label{subfig:resnet34_diffwidth}
    \end{subfigure}
    \begin{subfigure}{0.24\linewidth}
        \includegraphics[width=\linewidth]{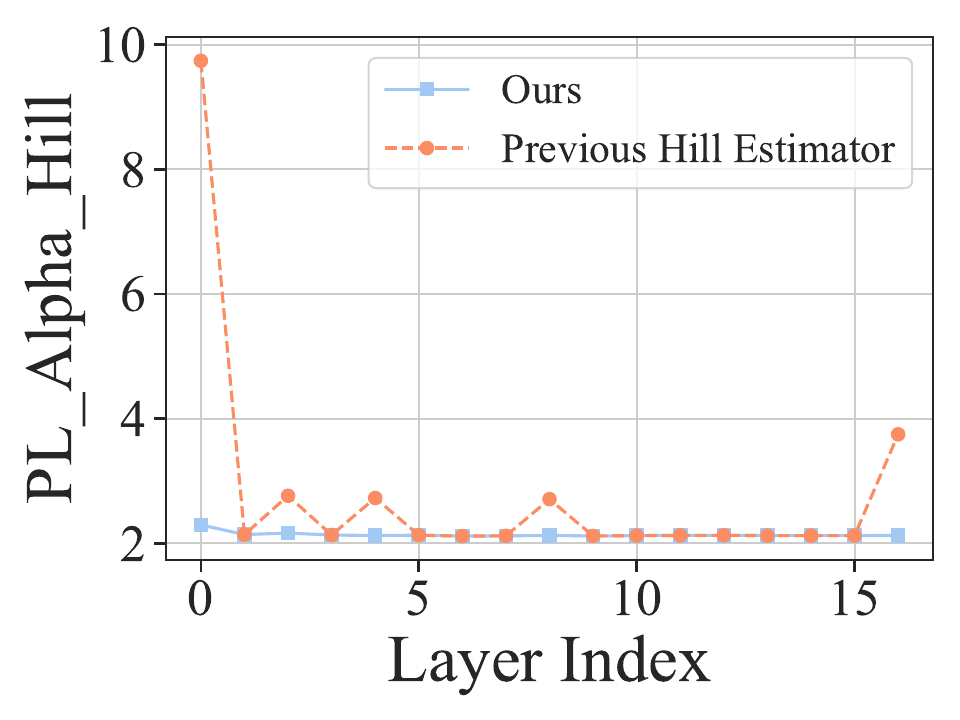}
        \caption{VGG 19, All Layers}
        \label{subfig:vgg19_alllayers}
    \end{subfigure}
    \begin{subfigure}{0.24\linewidth}
        \includegraphics[width=\linewidth]{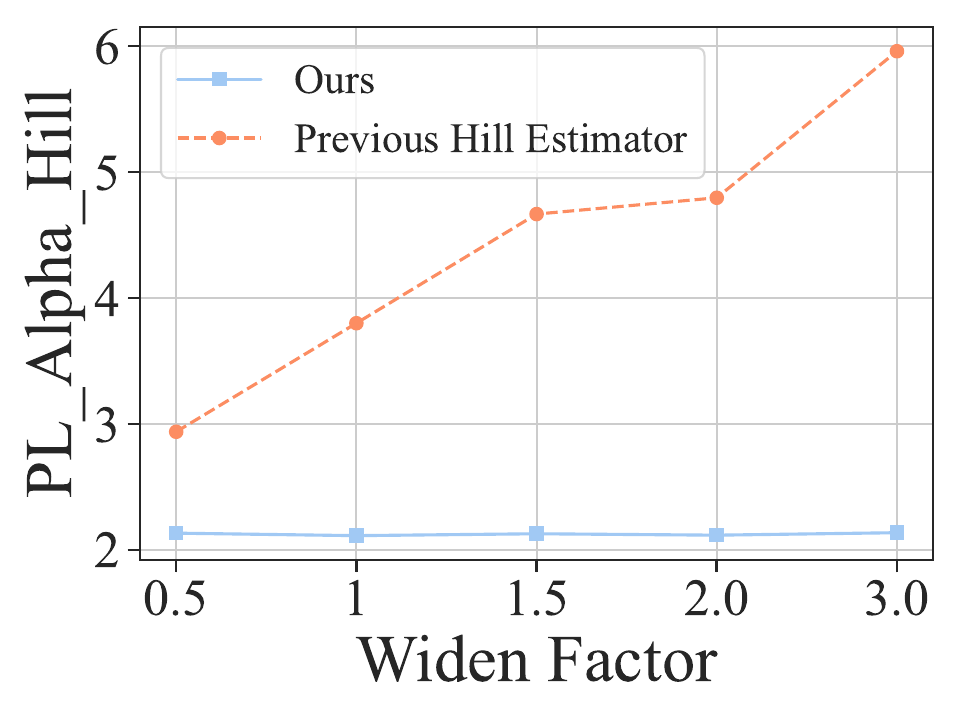}
        \caption{VGG 19, Different Width}
        \label{subfig:vgg19_diffwidth}
    \end{subfigure}

    \caption{Comparing \ourmethod and previous HT-SR methods for measuring the randomly initialized ResNet 34 and VGG 19 weights. Figure \ref{subfig:resnet34_alllayers} and Figure \ref{subfig:vgg19_alllayers} show the \AlphaHill values for each layer in models (widen factor is 1.0) by using different methods. Figure \ref{subfig:resnet34_diffwidth} and Figure \ref{subfig:vgg19_diffwidth} show the measured \AlphaHill values of the final linear layer across different model width factors. As the width increases, the aspect ratio of the weight matrix also becomes larger, leading to bias in previous HT-SR methods.}
    \label{fig:model_init_different_width}
\end{figure}
\else
\begin{figure}[!tbh]
    \centering
    \begin{subfigure}{0.24\linewidth}
        \includegraphics[width=\linewidth]{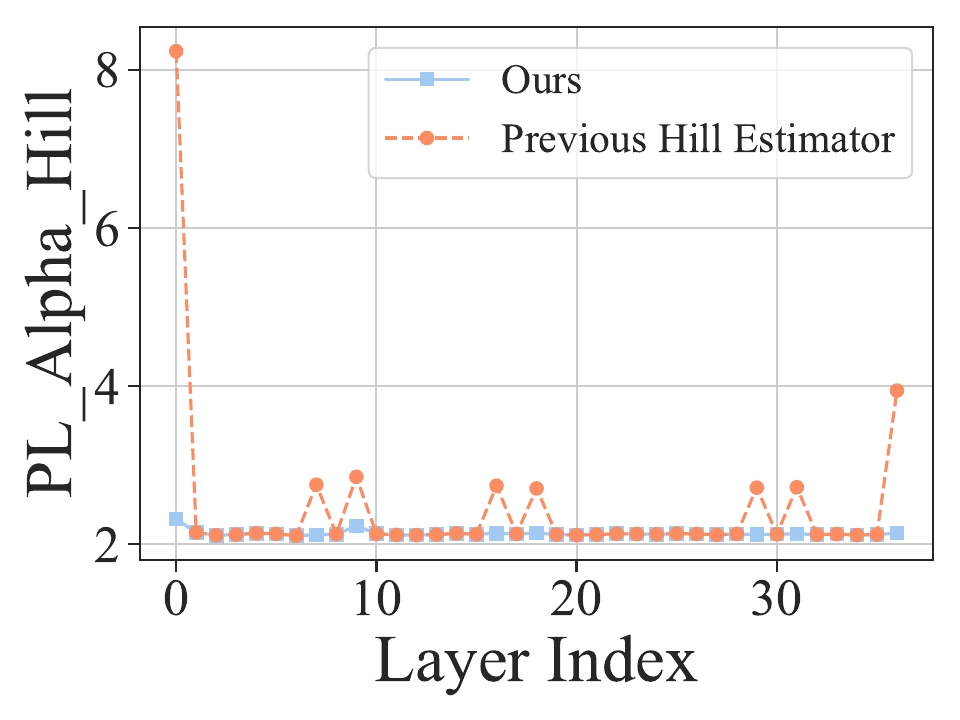}
        \caption{ResNet 34, All Layers}
        \label{subfig:resnet34_alllayers}
    \end{subfigure}
    \begin{subfigure}{0.24\linewidth}
        \includegraphics[width=\linewidth]{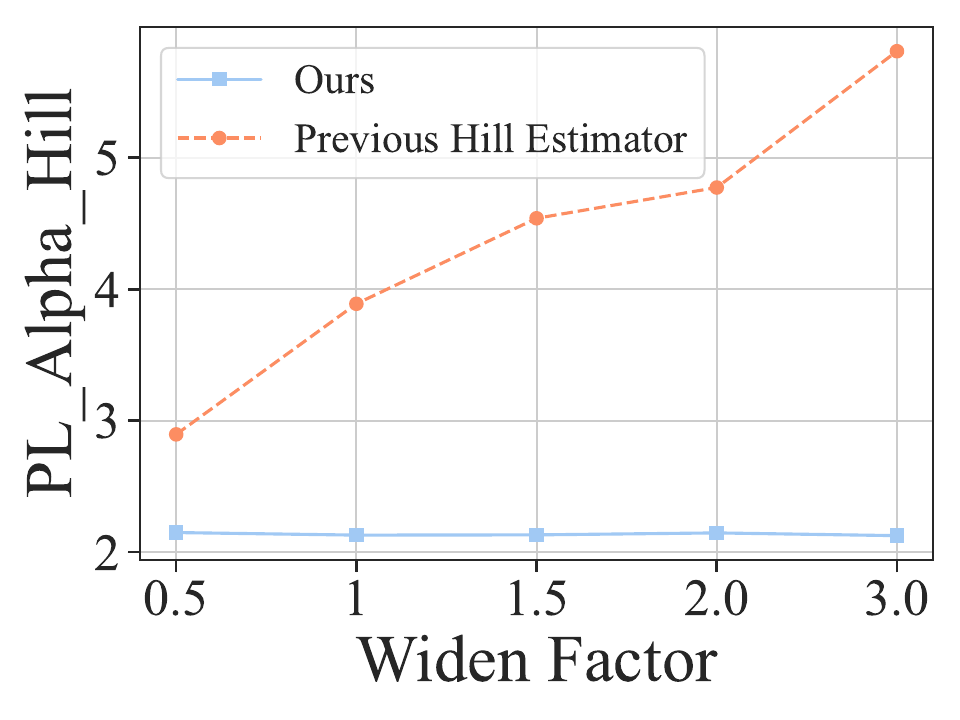}
        \caption{ResNet 34, Width}
        \label{subfig:resnet34_diffwidth}
    \end{subfigure}
    \begin{subfigure}{0.24\linewidth}
        \includegraphics[width=\linewidth]{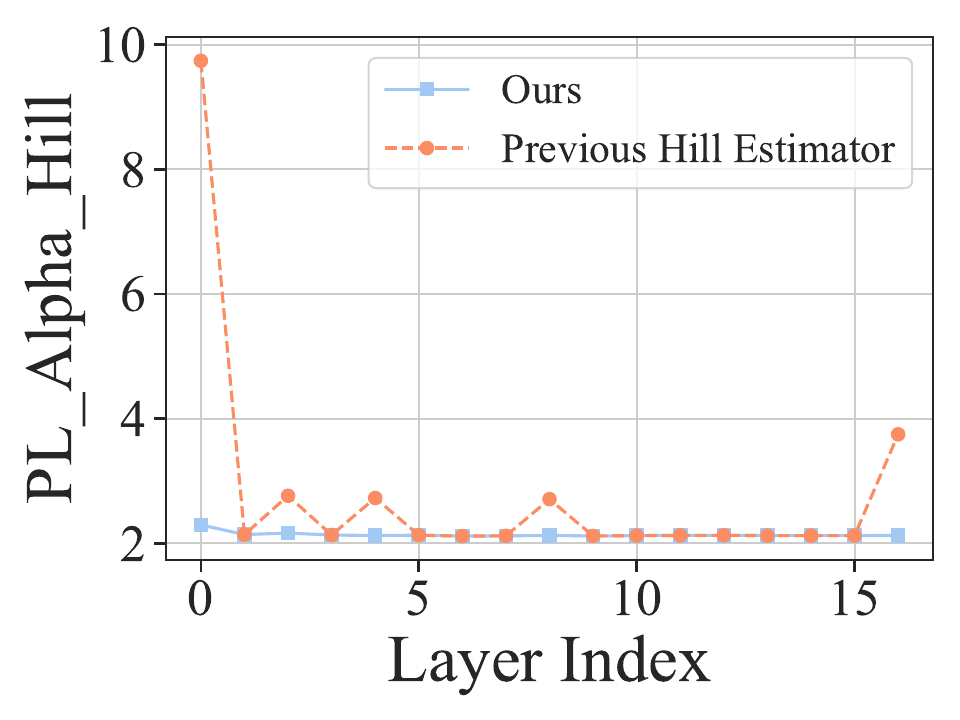}
        \caption{VGG 19, All Layers}
        \label{subfig:vgg19_alllayers}
    \end{subfigure}
    \begin{subfigure}{0.24\linewidth}
        \includegraphics[width=\linewidth]{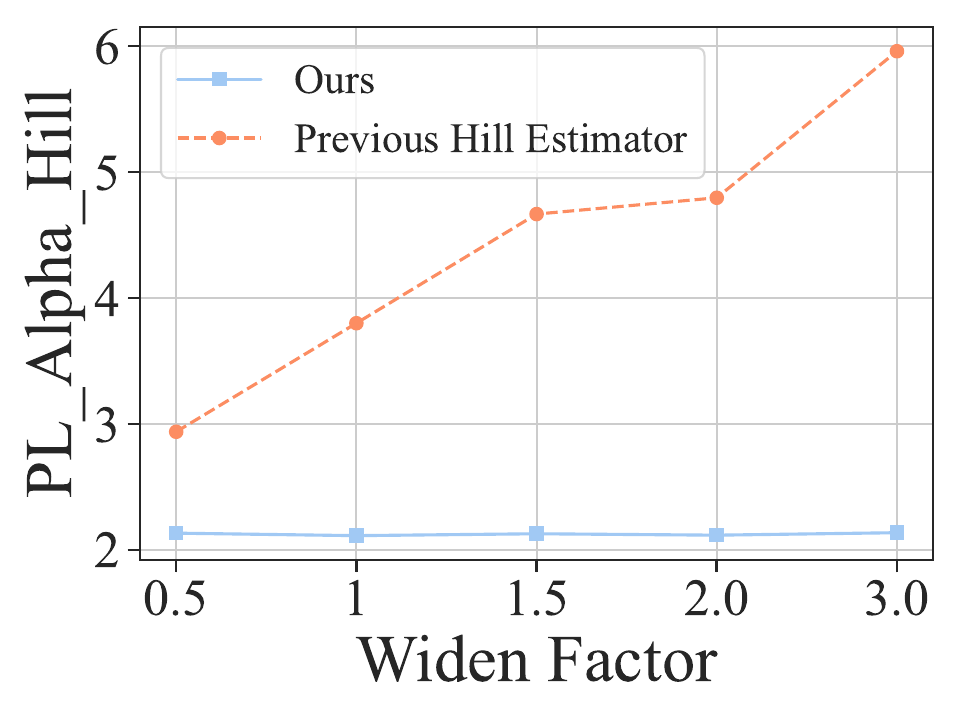}
        \caption{VGG 19, Width}
        \label{subfig:vgg19_diffwidth}
    \end{subfigure}

    \caption{Comparing \ourmethod and previous HT-SR methods for measuring the randomly initialized ResNet 34 and VGG 19 weights. Figure \ref{subfig:resnet34_alllayers} and Figure \ref{subfig:vgg19_alllayers} show the \AlphaHill values for each layer in models (widen factor is 1.0) by using different methods. Figure \ref{subfig:resnet34_diffwidth} and Figure \ref{subfig:vgg19_diffwidth} show the measured \AlphaHill values of the final linear layer across different model width factors. As the width increases, the aspect ratio of the weight matrix also becomes larger, leading to bias in previous HT-SR methods.}
    \label{fig:model_init_different_width}
\end{figure}
\fi

\subsection{Mitigating Aspect Ratio Bias in Specific Layers}

In this section, we do weight analysis for the last layer from the ResNet 34 and VGG 16 models trained on CIFAR100. These models are trained with hyperparameters according to the Appendix \ref{appendix:Hyperparamter_settings}. The weight matrices of the final layers in both models have dimensions of $512 \times 100$ (aspect ratio = 5.12). This means that the previous weight matrix analysis method will be significantly affected by aspect ratio bias, leading to inaccurate assessments of the layer's training quality. 

In Figure \ref{fig:final_layer_analysis_resnet_vgg}, the presented experimental results validate this observation. Figure \ref{subfig:resnet34_lastlayer_baseline} and Figure \ref{subfig:vgg16_lastlayer_baseline} show the ESD fitting results obtained using the previous weight analysis method, where the measured \AlphaHill values are larger than those typically observed in well-trained layers. This leads to the erroneous classification of these layers as poorly trained, ultimately affecting both model performance evaluation and optimization. In contrast, Figure \ref{subfig:resnet34_lastlayer_ourmethod} and Figure \ref{subfig:vgg16_lastlayer_ourmethod} present the ESD fitting results obtained using \ourmethod. Our method produces measurements that align more closely with the expected ESD of well-trained layers, providing a more accurate assessment of training quality.

\ificml
\begin{figure}[!tbh]
    \centering
    \begin{subfigure}{0.24\linewidth}
        \includegraphics[width=\linewidth]{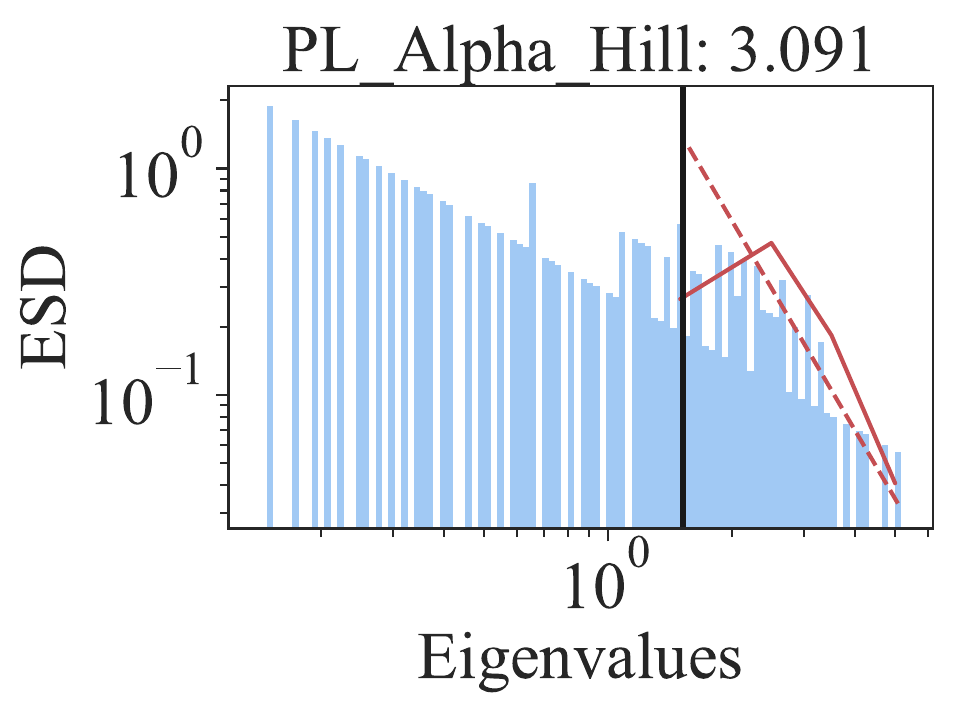}
        \caption{Previous Method, ResNet 34}
        \label{subfig:resnet34_lastlayer_baseline}
    \end{subfigure}
    \begin{subfigure}{0.24\linewidth}
        \includegraphics[width=\linewidth]{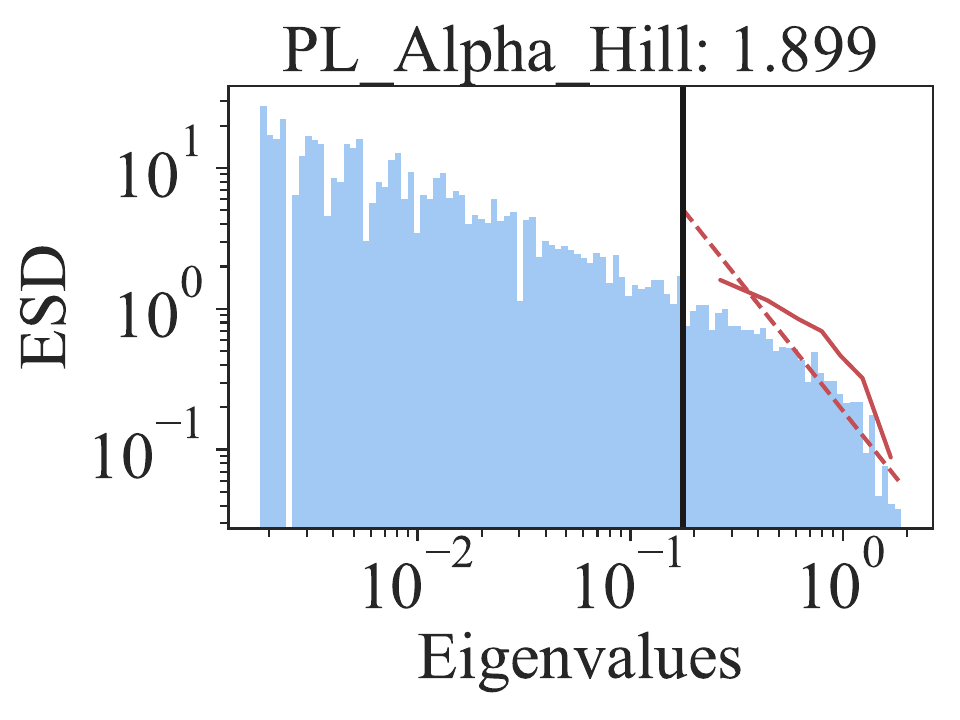}
        \caption{\ourmethod, ResNet 34}
        \label{subfig:resnet34_lastlayer_ourmethod}
    \end{subfigure}
    \begin{subfigure}{0.24\linewidth}
        \includegraphics[width=\linewidth]{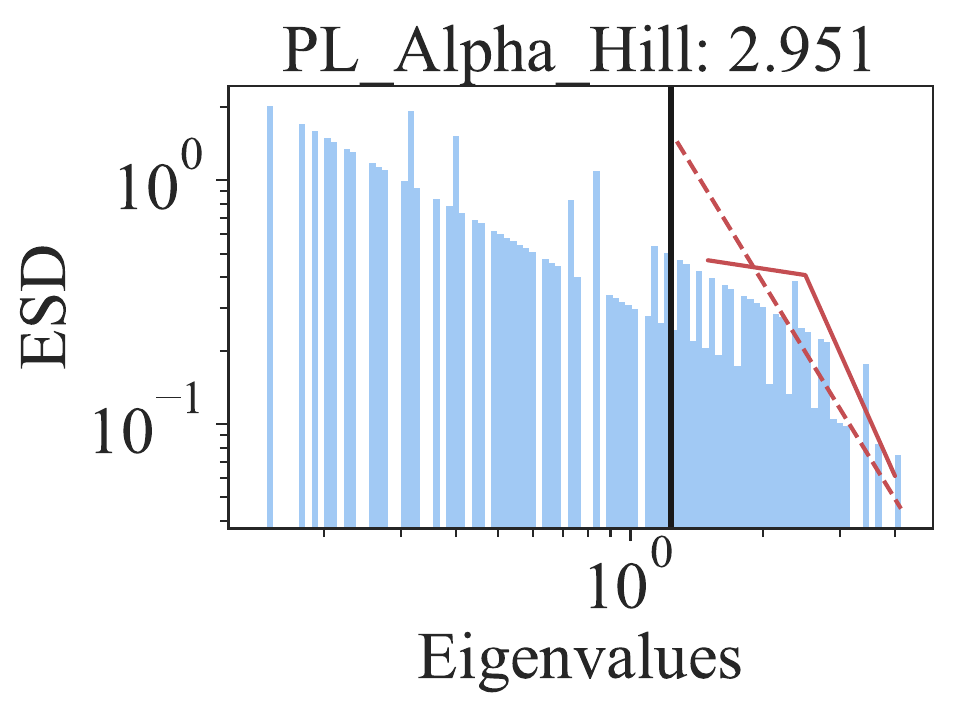}
        \caption{Previous Method, VGG 16}
        \label{subfig:vgg16_lastlayer_baseline}
    \end{subfigure}
    \begin{subfigure}{0.24\linewidth}
        \includegraphics[width=\linewidth]{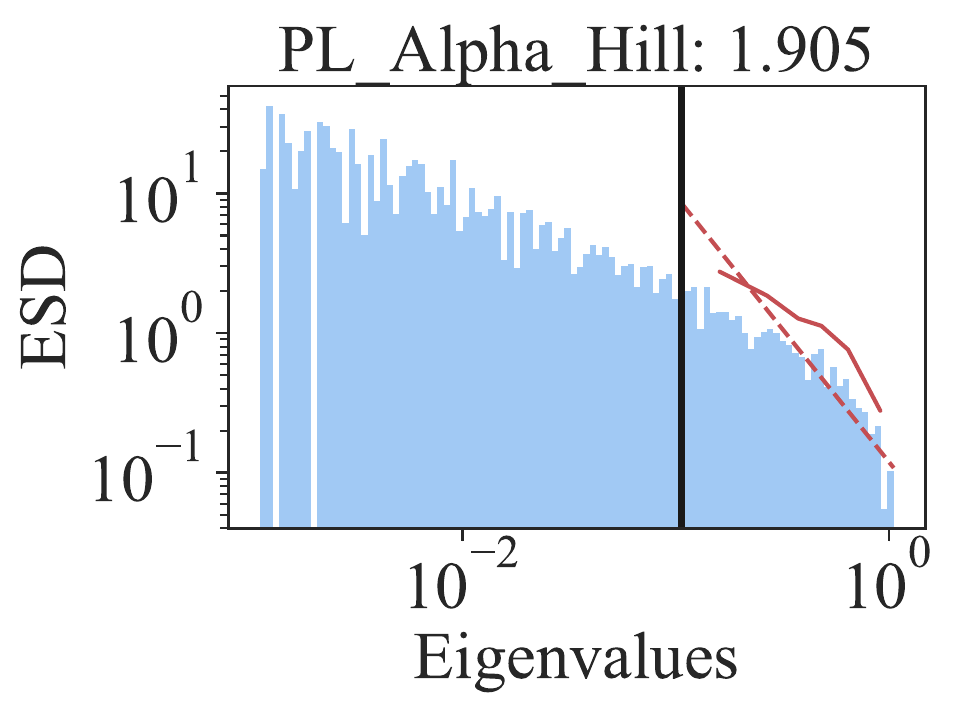}
        \caption{\ourmethod, VGG 16}
        \label{subfig:vgg16_lastlayer_ourmethod}
    \end{subfigure}

    \caption{Comparing \ourmethod and the previous method for measuring the ESD of final layers in ResNet 34 and VGG 16 trained on CIFAR 100.}
    \label{fig:final_layer_analysis_resnet_vgg}
\end{figure}
\else
\begin{figure}[!tbh]
    \centering
    \begin{subfigure}{0.24\linewidth}
        \includegraphics[width=\linewidth]{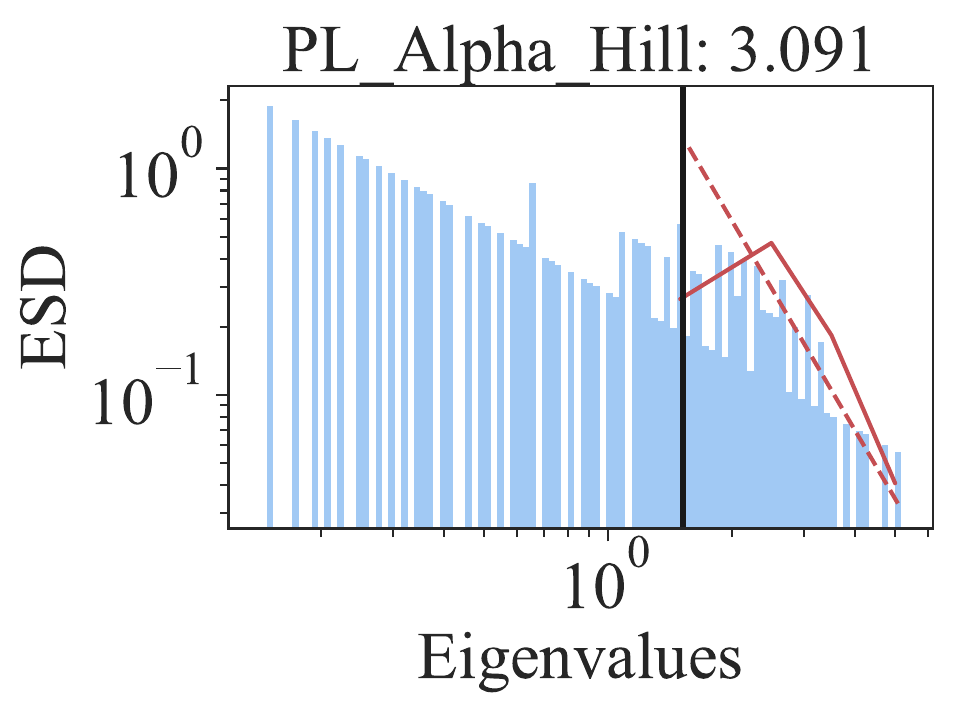}
        \caption{Previous, ResNet 34}
        \label{subfig:resnet34_lastlayer_baseline}
    \end{subfigure}
    \begin{subfigure}{0.24\linewidth}
        \includegraphics[width=\linewidth]{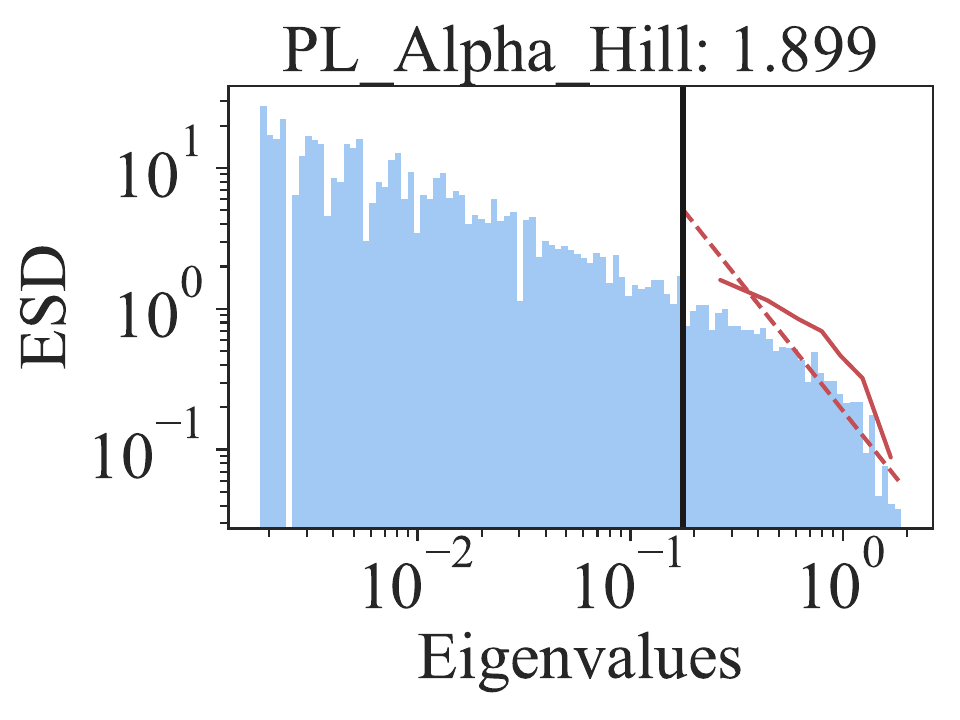}
        \caption{\ourmethod, ResNet 34}
        \label{subfig:resnet34_lastlayer_ourmethod}
    \end{subfigure}
    \begin{subfigure}{0.24\linewidth}
        \includegraphics[width=\linewidth]{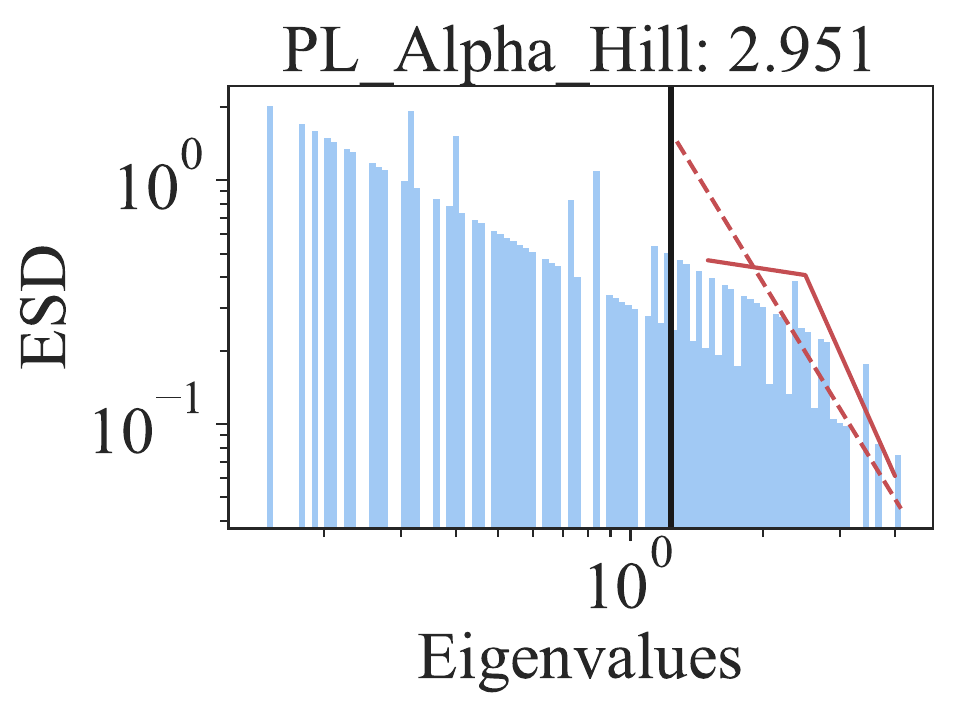}
        \caption{Previous, VGG 16}
        \label{subfig:vgg16_lastlayer_baseline}
    \end{subfigure}
    \begin{subfigure}{0.24\linewidth}
        \includegraphics[width=\linewidth]{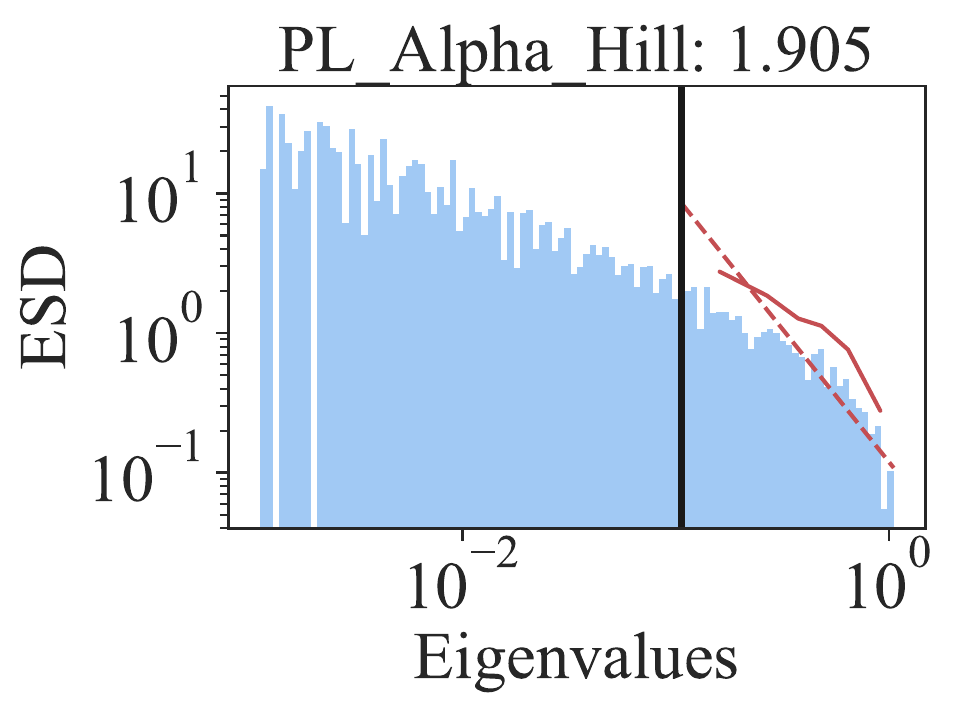}
        \caption{\ourmethod, VGG 16}
        \label{subfig:vgg16_lastlayer_ourmethod}
    \end{subfigure}

    \caption{Comparing \ourmethod and the previous method for measuring the ESD of final layers in ResNet 34 and VGG 16 trained on CIFAR 100.}
    \label{fig:final_layer_analysis_resnet_vgg}
\end{figure}
\fi

\section{Marchenko–Pastur Distribution}
\label{appendix:mp_law}

\begin{figure}[!tbh]
    \centering
    \begin{subfigure}{0.8\linewidth}
        \includegraphics[width=\linewidth]{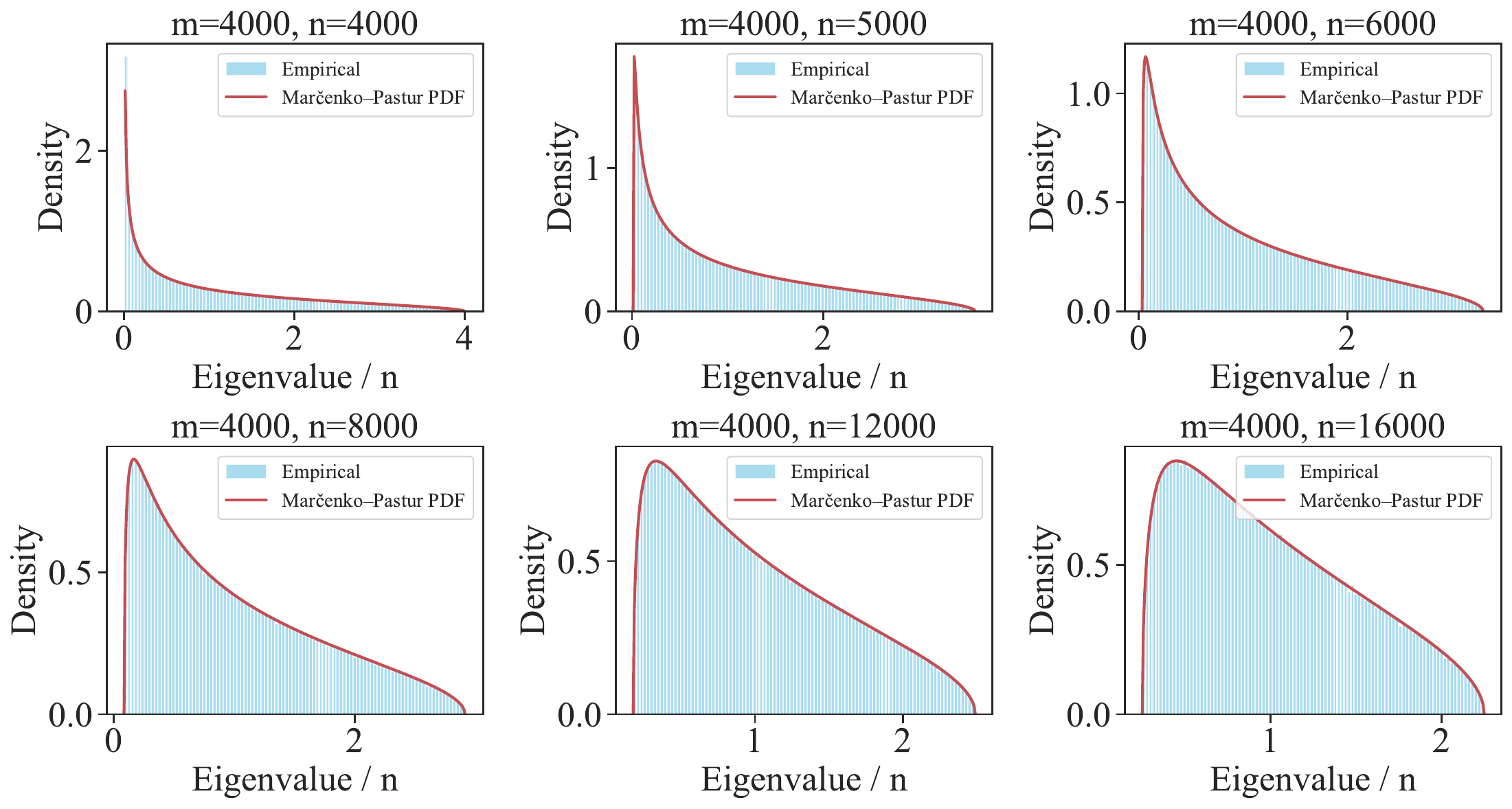}
    \end{subfigure}
        \caption{The Marchenko-Pastur (MP) Law for different values of $n$. The ESD (blue histogram) of the eigenvalues of the sample covariance matrix is compared with the theoretical Marchenko-Pastur probability density function (red curve) for various aspect ratios, where $m = 4000$ is fixed, and $n$ varies from 4000 to 16000.}
    \label{fig:mp_law_different_mn}
\end{figure}

In RMT, the Marchenko-Pastur distribution, also known as the Marchenko-Pastur law, characterizes the asymptotic behavior of singular values of large rectangular random matrices.

Let $X_{ij}, \, 1 \leq i \leq m, \, 1 \leq j \leq n$, be independent random variables with $\mathbb{E}X_{ij} = 0$ and $\mathbb{E}X_{ij}^2 = 1$, and $\mathbf{X}_m = (X_{ij})_{1 \leq i \leq m, 1 \leq j \leq n}$. Denote by $\lambda_1 \leq \dots \leq \lambda_m$ the eigenvalues of the symmetric matrix\footnote{\textbf{Note.} The transpose operation applies to the second (right-hand) matrix in this expression, which differs from the transpose placement used in HT-SR Theory, shown as $\mathbf{X}_{ij}=\mathbf{W}_{ij}^\top \mathbf{W}_{ij}$ in Section \ref{section:esd_farms}. Thus, the “large aspect ratio” regime discussed in HT-SR Theory corresponds to a large value of the ratio $n/m$ in Appendix \ref{appendix:mp_law}. }

\[
\mathbf{W} := \mathbf{W}_m := \frac{1}{n} \mathbf{X}_m \mathbf{X}_m^\top
\]

and defined its empirical distribution by
\[
F_m(x) = \frac{1}{m} \sum_{k=1}^m I_{\{\lambda_k \leq x\}},
\]
where $I_{\{B\}}$ denotes the indicator of an event $B$. One often investigates the rate of convergence of the expected spectral distribution $\mathbb{E}F_m(x)$ as well as $F_m(x)$ to the Marchenko–Pastur distribution function $F_y(x)$ with density
\[
f_y(x) = \frac{1}{2xy\pi} \sqrt{(b - x)(x - a)} I_{\{[a,b]\}}(x) + I_{\{[1,\infty)\}}(y)(1 - y^{-1})\delta(x),
\]
where $y=m/n$, $y \in (0, \infty)$ and $a = (1 - \sqrt{y})^2, \, b = (1 + \sqrt{y})^2$. Here we denote by $\delta(x)$ the Dirac delta function and by $I_{\{[a,b]\}}(x)$ the indicator function of the interval $[a, b]$.


We visualize the MP distribution and the ESD of the weight matrices in Figure \ref{fig:mp_law_different_mn}. In this figure, we can observe that the empirical distribution converges to the theoretical MP distribution. Additionally, as $n$ increases, the ESD distribution exhibits an increasingly concentrated shape with a reduced degree of HT. Therefore, ignoring the impact of aspect ratio and directly measuring the HT degree of the ESD to estimate a layer's training quality may lead to inaccurate results.

\section{Additional Experiment Results}
\label{appendix:more_results}

\subsection{Detailed Performance on Different Scaling Ratios}

\begin{figure*}[!tbh]
    \centering
    \begin{subfigure}{0.28\linewidth}
        \includegraphics[width=\linewidth]{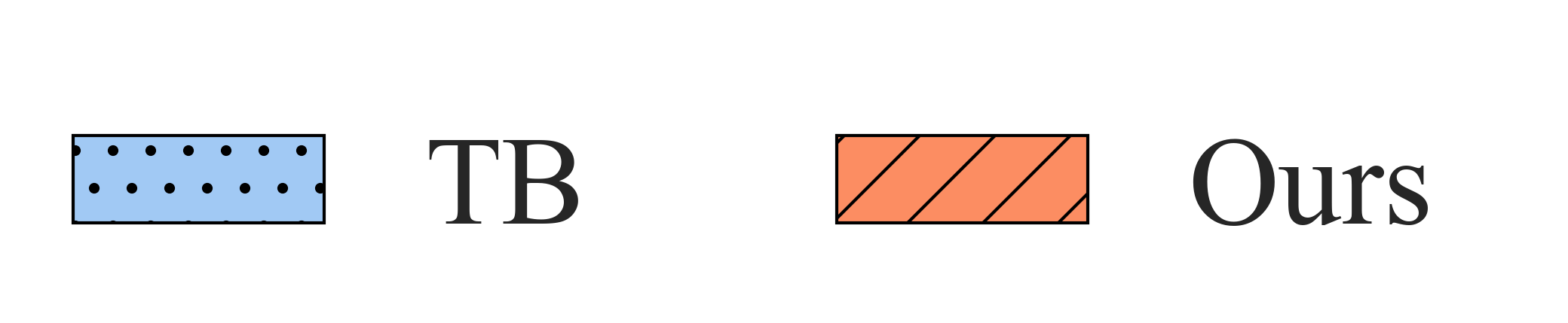}
    \end{subfigure}

    \begin{subfigure}{0.24\linewidth}
        \includegraphics[width=\linewidth]{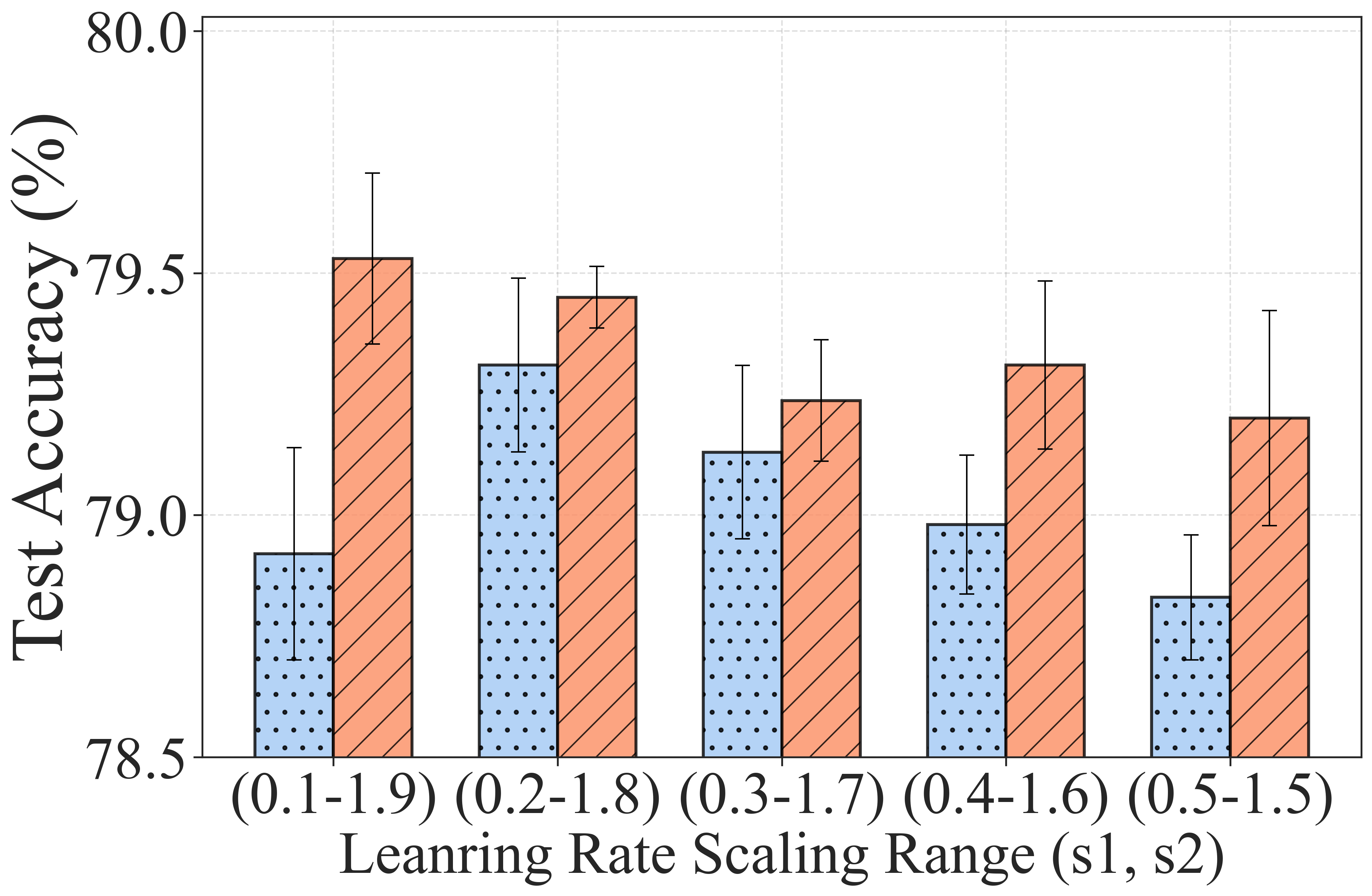}
        \caption{ResNet 18, CIFAR 100}
    \end{subfigure}
    \begin{subfigure}{0.24\linewidth}
        \includegraphics[width=\linewidth]{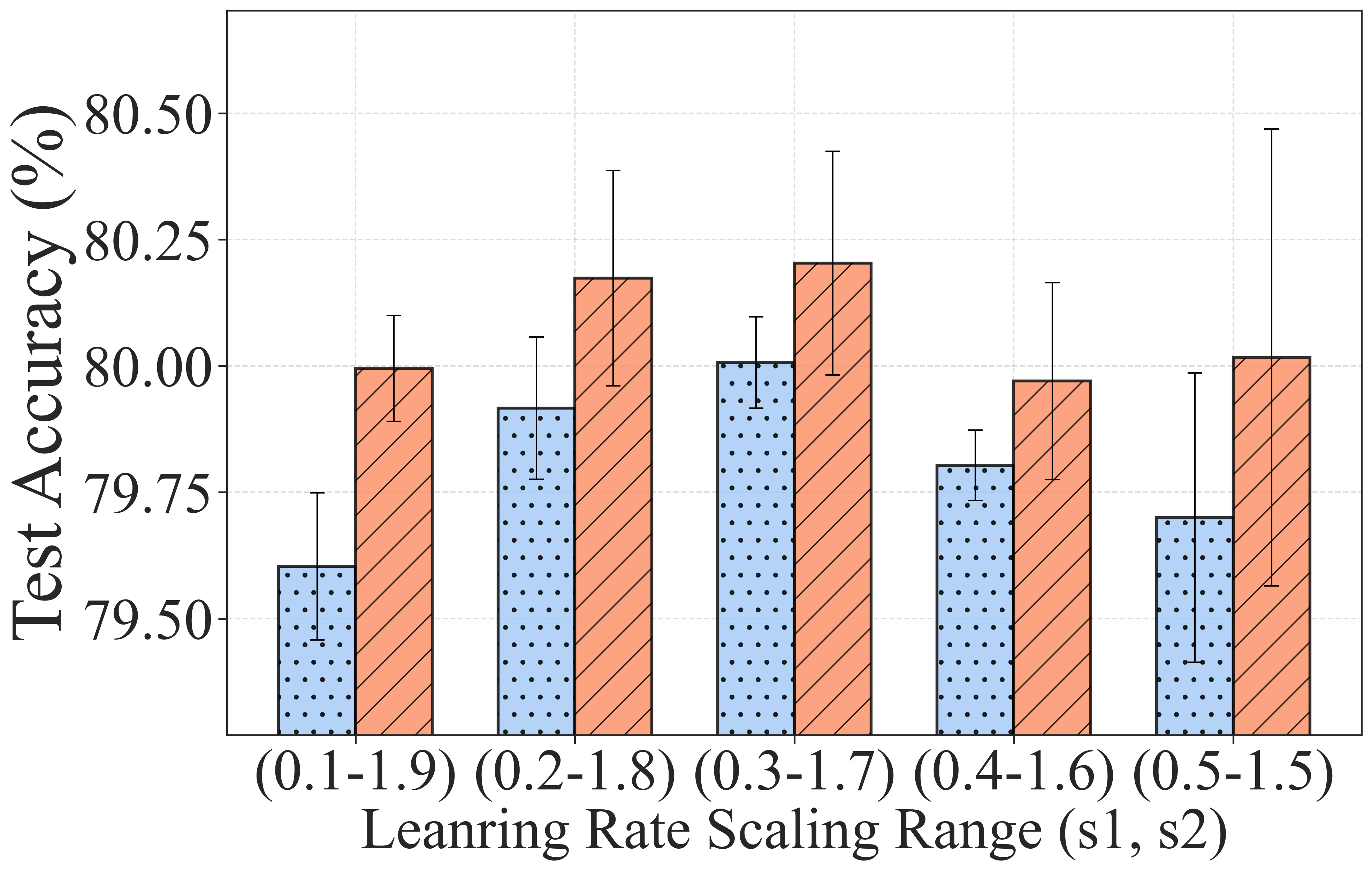}
        \caption{ResNet 34, CIFAR 100}
    \end{subfigure}
    \begin{subfigure}{0.24\linewidth}
        \includegraphics[width=\linewidth]
        {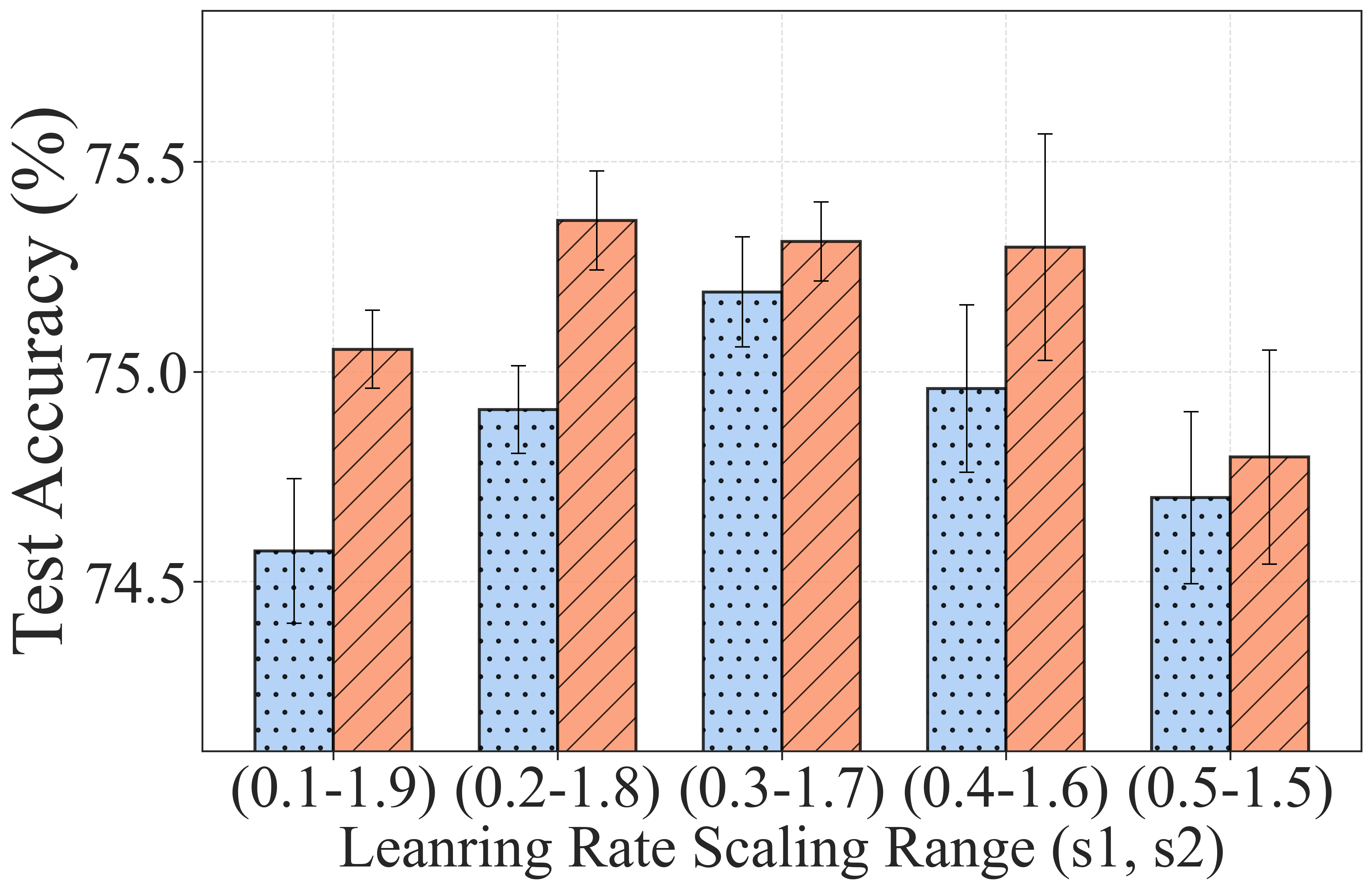}
        \caption{VGG 16, CIFAR 100}
    \end{subfigure}
    \begin{subfigure}{0.24\linewidth}
        \includegraphics[width=\linewidth]
        {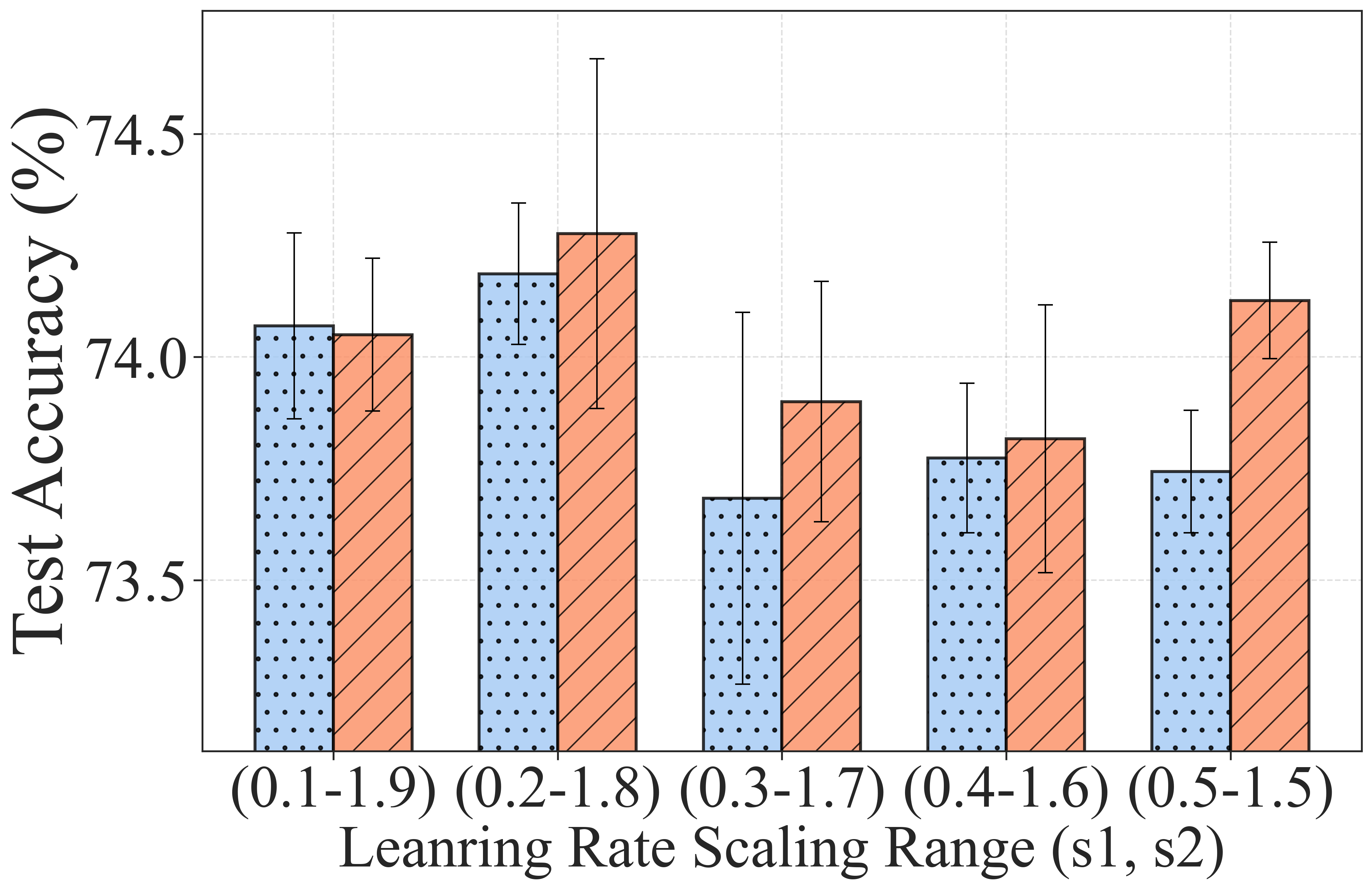}
        \caption{VGG 19, CIFAR 100}
    \end{subfigure}

    \caption{Comparison of test accuracy across different architectures and learning rate scaling ranges. The figure presents the test accuracy (\%) of \TB (blue, dotted) and \ourmethod (orange, hatched). These two methods are evaluated on the CIFAR 100 dataset using ResNet and VGG series architectures. The x-axis represents different learning rate scaling ranges $(s_1, s_2)$, while the y-axis indicates test accuracy. Error bars denote standard deviations across multiple runs.}

    \label{fig:lr_range_imgcls}
\end{figure*}

We provide detailed results of a hyperparameter study on learning rate scaling ratio $(s_{1}, s_{2})$. We set other hyperparameters (including layer selection, initial learning rate, and so on) as the optimal value for each training set. The results in Figure \ref{fig:lr_range_imgcls} show that across all tested architectures and most scaling ratios, \ourmethod consistently outperforms \TB. This demonstrates that \ourmethod provides a more accurate measurement of each layer's training quality. As a result, it helps the model achieve better training performance when using different learning rate scaling ratios.

\subsection{Comparison of Different CNN Processing Methods}
\label{appendix:cnn_methods_comp}

We compared two different methods for processing the ESD of CNNs. The first method concatenates the rooted singular values of all submatrices and then measures the HT metrics of the resulting ESD. The second method measures the HT metrics of the ESD formed by concatenating the rooted singular values of each of the $C_{1} \times C_{2}$ submatrices separately, and then averages these HT metrics from each group. In Table \ref{table:imgcls_cnn_methods_comp}, we provide a performance comparison of training ResNet-18 and VGG-16 models using these two methods. For each experiment, all other hyperparameters are set to their optimal values. We find that although the performance difference between the two methods is not very large, the second method brings more performance improvement.

\begin{table}[!thb]
    \centering
    \caption{Comparing Two Stratgy in CNN Processing with \ourmethod.} 
    \vspace{0.25cm}
    \resizebox{0.5\linewidth}{!}{
    \begin{tabular}{c|cc}
        \toprule
        \bf{Method} & \bf{ResNet 18} & \bf{VGG 16} \\
        \midrule
        
        Calculating the Overall ESD  & 79.36{\scriptsize$\pm$0.101}  & 75.27{\scriptsize$\pm$0.201}  \\

        \rowcolor{LightGrey}
        Averaging ESDs Computed from Subsets  & \bf{79.53{\scriptsize$\pm$0.177}}  & \bf{75.36{\scriptsize$\pm$0.118}}  \\
        
        \midrule
    \end{tabular}
    }
    \label{table:imgcls_cnn_methods_comp} 
\end{table}

\subsection{Experiment Results for Additional Models}

We prune the OPT models ~\citep{zhang2022opt} and LLaMA-V2/V3 models ~\citep{touvron2023llamav2,grattafiori2024llama} with \Alphapruning and compare the model performance with \ourmethod. We report the additional experiment results on LLM pruning in Table ~\ref{table:opt_pruning_wiki} and Table ~\ref{table:llamav2v3_pruning_wiki}. For the OPT series models, we use four sparsity ratios: \{0.6, 0.7, 0.75, 0.8\}. For the LLaMA series models, we select \{0.7, 0.75, 0.8, 0.85\} as the sparsity ratios. Experimental results show that our method can help \Alphapruning achieve better model performance on models such as OPT.

\begin{table*}[!thb]
    \centering
    \caption{WikiText validation perplexity for pruned OPT models at different sparsity settings. Our method is compared to \Alphapruning, each paired with Wanda and SparseGPT. Lower perplexity indicates improved model performance.} 
    \vspace{0.25cm}
    \resizebox{0.98\linewidth}{!}{
    \begin{tabular}{c|c|cc|cc|cc|cc}
        \toprule
        \bf{Sparsity Ratio} & \bf{Layer-wise} & \multicolumn{2}{c|}{\bf{OPT-125M}}  & \multicolumn{2}{c|}{\bf{OPT-350M}} & \multicolumn{2}{c|}{\bf{OPT-1.3B}}  & \multicolumn{2}{c}{\bf{OPT-6.7B}} \\
        
        & \bf{Method}  & \bf{Wanda} & \bf{SparseGPT} & \bf{Wanda} & \bf{SparseGPT} & \bf{Wanda} & \bf{SparseGPT} & \bf{Wanda} & \bf{SparseGPT} \\
        
        \midrule
        0.6 
        & AlphaPruning & 67.12  & 59.18  & 87.95  & 50.94  & \textbf{27.09}  & 22.76 & 15.6  & 13.71 \\
        \rowcolor{LightGrey} 
        & Ours         & \textbf{66.71}  & \textbf{58.86}  & \textbf{82.55}  & \textbf{50.52}  & 27.15  & \textbf{22.69} & \textbf{15.59} & \textbf{13.67} \\

        \midrule
        0.7 
        & AlphaPruning & 263.71 & 196.64  & 595.26  & 147.20  &   101.13 & 50.30  &  44.92 & 20.76  \\
        \rowcolor{LightGrey} 
        & Ours         & \textbf{261.84} & \textbf{188.31}  & \textbf{489.75}  & \textbf{136.80}  &   \textbf{100.74} & \textbf{50.01}  &  \textbf{42.03} & \textbf{20.76} \\

        \midrule
        0.75 
        & AlphaPruning & \textbf{718.69}  & 515.55  & 1453.21 & 330.11  & 613.78  & 142.68  & 245.16 & 35.27  \\
        \rowcolor{LightGrey} 
        & Ours         & 721.52  & \textbf{491.98}  & \textbf{1298.22}  & \textbf{303.07}  & \textbf{601.24}  & \textbf{122.01}  & \textbf{181.92} & \textbf{34.88}  \\

        \midrule
        0.8 
        & AlphaPruning & 1713.45  & 1525.25  & 2869.8  & 921.17  & 2763.33 & 511.66 & 5781.0 & 98.3 \\
        \rowcolor{LightGrey} 
        & Ours         & \textbf{1609.36}  & \textbf{1412.59}  & \textbf{2551.46}  & \textbf{798.25}  & \textbf{2502.60} & \textbf{499.82} & \textbf{5212.51} & \textbf{90.88} \\  

        \bottomrule
    \end{tabular}
    }
    \label{table:opt_pruning_wiki}
\end{table*}

\begin{table*}[!thb]
    \centering
    \caption{WikiText validation perplexity for pruned LLaMA-V2 and LLaMA-V3 models at different sparsity settings. Our method is compared to \Alphapruning, each paired with Wanda and SparseGPT. Lower perplexity indicates improved model performance.} 
    \vspace{0.25cm}
    \resizebox{0.98\linewidth}{!}{
    \begin{tabular}{c|c|cc|cc|cc|cc}
        \toprule
        \bf{Sparsity Ratio} & \bf{Layer-wise} & \multicolumn{2}{c|}{\bf{LLaMA-V2-7B}}  & \multicolumn{2}{c|}{\bf{LLaMA-V2-13B}} & \multicolumn{2}{c|}{\bf{LLaMA-V3.2-3B}}  & \multicolumn{2}{c}{\bf{LLaMA-V3.1-8B}} \\
        
        & \bf{Method}  & \bf{Wanda} & \bf{SparseGPT} & \bf{Wanda} & \bf{SparseGPT} & \bf{Wanda} & \bf{SparseGPT} & \bf{Wanda} & \bf{SparseGPT} \\
        
        \midrule
        0.7 
        & AlphaPruning & 34.71  & 20.92  & 15.37  & 13.69  & 123.19  & 64.33 & \textbf{105.64}  & 38.43 \\
        \rowcolor{LightGrey} 
        & Ours         & \textbf{31.09}  & \textbf{20.49}  & \textbf{15.36}  & \textbf{13.64}  & \textbf{120.03}  & \textbf{63.78} & 107.00 & \textbf{37.81} \\

        \midrule
        0.75 
        & AlphaPruning & 170.74 & 41.39  & 35.26  & 23.21  &   356.69 & 121.53  &  240.17 & \textbf{81.30}  \\
        \rowcolor{LightGrey} 
        & Ours         & \textbf{161.58} & \textbf{39.98}  & \textbf{33.67}  & \textbf{22.98}  &   \textbf{311.73} & \textbf{117.19}  &  \textbf{229.63} & 81.93 \\

        \midrule
        0.8 
        & AlphaPruning & 868.66  & 89.73  & 150.01 & 46.32 & 1412.53  & 262.44  & 688.23 & 180.40  \\
        \rowcolor{LightGrey} 
        & Ours         & \textbf{831.37}  & \textbf{88.25}  & \textbf{127.69}  & \textbf{44.74}  & \textbf{1219.10}  & \textbf{212.65}  & \textbf{607.58} & \textbf{175.93}  \\

        \midrule
        0.85 
        & AlphaPruning & 6425.25  & 217.79  & 874.97  & 107.21  & 5857.39 & 505.8 & \textbf{3498.94} & \textbf{400.69} \\
        \rowcolor{LightGrey} 
        & Ours         & \textbf{4437.46}  & \textbf{204.44}  & \textbf{748.91}  & \textbf{98.99}  & \textbf{4407.94} & \textbf{463.34} & 3766.13 & 411.35 \\  

        \bottomrule
    \end{tabular}
    }
    \label{table:llamav2v3_pruning_wiki}
\end{table*}

\subsection{Zero-shot Tasks Performance}

We demonstrate the task-wise performance in detail in Table ~\ref{table:llama7b_pruning_zeroshot_details} and Table ~\ref{table:llama13b_pruning_zeroshot_details}.

\begin{table*}[!tbh]
    \centering
     \caption{Accuracies (\%) of LLaMA-7B for 7 zero-shot tasks with unstructured sparsity from 70\% to 85\%. We compare \ourmethod with uniform pruning ratios and \Alphapruning using Magnitude-based pruning, Wanda and SparseGPT.}
    \vspace{0.25cm}
    \resizebox{0.9\linewidth}{!}{
    \begin{tabular}{cccccccccc}
        \toprule
        Sparsity Ratio & Method & BoolQ & RTE & HellaSwag & WinoGrande & ARC-e & ARC-c & OBQA & Mean  \\
        
        \midrule
        & Magnitude                          & 38.29 & 52.71 & 25.59 & 51.22 & 26.73 & 19.62 & 11.60 & 32.25   \\
        & AlphaPruning $w$. Magnitude        & \bf{41.31} & \bf{52.71} & 30.37 & 51.46 & \bf{35.44} & 22.01 & 16.40 & 35.67     \\
        \rowcolor{LightGrey}
        & Ours $w$. Magnitude         & 40.09 & 52.71 & \bf{30.49} & \bf{53.83} & 34.22 & \bf{23.21} & \bf{17.20} & \bf{35.96}   \\ 
        \cline{2-10}
        & Wanda                          & 56.06 & 55.60 & 28.90 & 50.99 & 32.11 & 18.26 & 13.60 & 36.50  \\
        0.7 & AlphaPruning $w$. Wanda        & 64.34 & 57.40 & 35.57 & \bf{61.09} & \bf{45.58} & 24.49 & 17.20 & 43.67    \\
        \rowcolor{LightGrey}
        & Ours $w$. Wanda         & \bf{64.40} & \bf{59.93} & \bf{36.54} & 60.62 & 45.24 & \bf{24.66} & \bf{18.40} & \bf{44.26}  \\
        \cline{2-10}
        & SparseGPT                     & 65.14 & \bf{53.79} & 33.95 & 58.72 & 44.44 & 23.98 & 17.20 & 42.46  \\
        & AlphaPruning $w.$ SparseGPT   & \bf{65.74} & 53.07 & \bf{37.72} & 64.01 & 47.35 & \bf{26.88} & 18.80 & 44.79  \\
        \rowcolor{LightGrey}
        & Ours $w.$ SparseGPT         & 65.54 & 53.07 & 36.85 & \bf{64.33} & \bf{48.11} & 26.37 & \bf{19.60} & \bf{44.84}   \\

        \midrule
        & Magnitude                          & 42.26 & 52.35 & 25.88 & 48.54 & 26.68 & \bf{21.42} & 14.00 & 33.02   \\
        & AlphaPruning $w$. Magnitude        & 50.67 & 52.71 & 26.80 & \bf{49.96} & 27.19 & 21.42 & 13.40 & 34.59     \\
        \rowcolor{LightGrey}
        & Ours $w$. Magnitude         & \bf{55.54} & \bf{53.07} & \bf{27.80} & 49.25 & \bf{29.21} & 21.08 & \bf{15.20} & \bf{35.88}   \\ 
        \cline{2-10}
        & Wanda                          & 37.83 & 53.79 & 27.01 & 49.96 & 27.74 & 19.37 & 12.60 & 32.61  \\
        0.75 & AlphaPruning $w$. Wanda        & 62.17 & 53.43 & 29.52 & 53.75 & 33.04 & 20.82 & 13.20 & 37.99    \\
        \rowcolor{LightGrey}
        & Ours $w$. Wanda        & \bf{62.17} & \bf{58.12} & \bf{30.99} & \bf{55.49} & \bf{35.69} & \bf{21.76} & \bf{13.40} & \bf{39.66}  \\
        \cline{2-10}
        & SparseGPT                     & 62.14 & \bf{53.43} & 29.76 & 51.22 & 33.80 & 19.11 & 13.40 & 37.55  \\
        & AlphaPruning $w.$ SparseGPT   & 63.71 & 52.71 & \bf{33.11} & \bf{59.91} & \bf{38.68} & \bf{23.89} & 14.20 & 40.89  \\
        \rowcolor{LightGrey}
        & Ours $w.$ SparseGPT & \bf{64.07} & 53.07 & 32.74 & 59.83 & 38.43 & 23.55 & \bf{14.80} & \bf{40.93}   \\

       \midrule
        & Magnitude                        & \bf{48.81} & 49.10 & 25.59 & 48.78 & 24.92 & 22.01 & \bf{14.20} & 33.34   \\
        & AlphaPruning $w$. Magnitude      & 44.40 & 53.07 & \bf{26.17} & 50.75 & \bf{25.67} & 22.27 & 14.20 & 33.79\\
        \rowcolor{LightGrey}
        & Ours $w$. Magnitude         & 46.76 & \bf{53.79} & 26.08 & \bf{52.33} & 24.58 & \bf{22.95} & 13.80 & \bf{34.33}   \\ 
        \cline{2-10}
        & Wanda                          & 61.69 & 51.42 & 25.85 & \bf{50.24} & 26.23 & \bf{20.62} & \bf{14.00} & 35.72 \\
        0.8 & AlphaPruning $w$. Wanda        & 52.75 & 51.62 & 26.53 & 48.54 & 26.81 & 20.39 & 11.80 & 34.06    \\
        \rowcolor{LightGrey}
        & Ours $w$. Wanda         & \bf{62.14} & \bf{51.62} & \bf{26.81} & 50.20 & \bf{27.57} & 20.56 & 11.40 & \bf{35.76}  \\
        \cline{2-10}
        & SparseGPT                     & 43.55 & 52.71 & 27.87 & 48.86 & 29.34 & 18.34 & 13.40 & 33.44  \\
        & AlphaPruning $w.$ SparseGPT   & 61.62 & 52.35 & 28.29 & 52.25 & \bf{30.64} & 19.28 & 12.00 & 36.63  \\
        \rowcolor{LightGrey}
        & Ours $w.$ SparseGPT        & \bf{62.26} & \bf{53.43} & \bf{29.29} & \bf{54.22} & 30.18 & \bf{19.71} & \bf{13.40} & \bf{37.50} \\

        \midrule
        & Magnitude                          & 48.75 & 51.62 & 25.58 & 48.46 & \bf{25.76} & \bf{22.61} & \bf{14.60} & 33.91   \\
        & AlphaPruning $w$. Magnitude        & 43.18 & \bf{52.71} & \bf{25.87} & 49.49 & 25.00 & 22.35 & 14.40 & 33.29     \\
        \rowcolor{LightGrey}
        & Ours $w$. Magnitude         & \bf{56.64} & 48.01 & 25.60 & \bf{49.49} & 25.08 & 22.44 & 12.60 & \bf{34.27}   \\ 
        \cline{2-10}
        & Wanda                          & \bf{51.74} & 45.85 & \bf{26.12} & 47.59 & 25.88 & 20.48 & \bf{15.00} & \bf{33.24}  \\
        0.85 & AlphaPruning $w$. Wanda        & 38.23 & 47.29 & 25.87 & 50.12 & 25.76 & \bf{22.18} & 12.40 & 31.69    \\
        \rowcolor{LightGrey}
        & Ours $w$. Wanda        & 49.91 & \bf{48.38} & 25.92 & \bf{50.12} & \bf{26.09} & 20.22 & 11.00 & 33.09  \\ 
        \cline{2-10}
        & SparseGPT                     & 37.89 & \bf{53.07} & 26.69 & 50.36 & \bf{26.98} & 19.28 & 11.40 & 32.24  \\
        & AlphaPruning $w.$ SparseGPT   & \bf{57.61} & 52.35 & 27.08 & 48.70 & 26.64 & 19.45 & 12.20 & 34.86  \\
        \rowcolor{LightGrey}
        & Ours $w.$ SparseGPT        & 55.69 & 51.62 & \bf{27.19} & \bf{52.88} & 26.98 & \bf{19.97} & \bf{12.40} & \bf{35.25}    \\

        \bottomrule
    \end{tabular}
    }
    
    \label{table:llama7b_pruning_zeroshot_details}
\end{table*}

\begin{table*}[!tbh]
    \centering
    \caption{Accuracies (\%) of LLaMA-13B for 7 zero-shot tasks with unstructured sparsity from 70\% to 85\%. We compare \ourmethod with uniform pruning ratios and \Alphapruning using Magnitude-based pruning, Wanda and SparseGPT.}
    \vspace{0.25cm}
    \resizebox{0.9\linewidth}{!}{
    \begin{tabular}{cccccccccc}
        \toprule
        Sparsity Ratio & Method & BoolQ & RTE & HellaSwag & WinoGrande & ARC-e & ARC-c & OBQA & Mean  \\
        
        \midrule
        & Magnitude                          & 52.87 & \bf{50.54} & 26.57 & 50.83 & 28.45 & 20.56 & 14.80 & 34.95   \\
        & AlphaPruning $w$. Magnitude        & \bf{61.28} & 46.93 & 30.24 & 50.43 & 31.23 & 26.28 & 21.20 & 38.23     \\
        \rowcolor{LightGrey}
        & Ours $w$. Magnitude               & 57.16 & 43.32 & \bf{33.90} & \bf{53.43} & \bf{34.97} & \bf{26.88} & \bf{22.40} & \bf{38.87}   \\ 
        \cline{2-10}
        & Wanda                          & 62.08 & 52.71 & 30.38 & 52.41 & 40.53 & 17.49 & 16.00 & 38.80  \\
        0.7 & AlphaPruning $w$. Wanda        & 63.43 & \bf{53.43} & \bf{42.16} & 65.59 & 57.53 & \bf{28.67} & 21.40 & 47.46    \\
        \rowcolor{LightGrey}
        & Ours $w$. Wanda         & \bf{66.18} & 52.71 & 41.94 & \bf{64.80} & \bf{57.87} & 28.58 & \bf{22.20} & \bf{47.75}  \\ 
        \cline{2-10}
        & SparseGPT                     & 69.17 & 52.71 & 37.25 & 63.22 & 51.98 & 24.83 & 21.60 & 45.82  \\
        & AlphaPruning $w.$ SparseGPT   & 72.05 & 54.15 & \bf{42.44} & 68.35 & 55.60 & 28.50 & \bf{22.40} & 49.07   \\ 
        \rowcolor{LightGrey}
        & Ours $w.$ SparseGPT        & \bf{73.03} & \bf{54.51} & 41.98 & \bf{69.06} & \bf{57.45} & \bf{29.27} & 22.00 & \bf{49.61}  \\

        \midrule
        & Magnitude                          & 45.05 & \bf{50.90} & 25.93 & 50.43 & 25.93 & 20.82 & 13.60 & 33.24   \\
        & AlphaPruning $w$. Magnitude        & 60.03 & 50.90 & \bf{30.10} & \bf{52.88} & 28.20 & \bf{25.60} & 19.40 & \bf{38.16}     \\
        \rowcolor{LightGrey}
        & Ours $w$. Magnitude                & \bf{60.06} & 44.04 & 28.29 & 52.57 & \bf{32.53} & 23.89 & \bf{22.40} & 37.68   \\ 
        \cline{2-10}
        & Wanda                          & 42.51 & 52.71 & 27.88 & 49.64 & 29.17 & 17.83 & 12.00 & 33.11  \\
        0.75 & AlphaPruning $w$. Wanda        & 62.17 & 52.71 & \bf{36.09} & \bf{62.67} & 44.15 & 23.29 & \bf{18.00} & \bf{42.73}    \\
        \rowcolor{LightGrey}
        & Ours $w$. Wanda        & \bf{62.42} & \bf{52.71} & 35.41 & 61.33 & \bf{46.93} & \bf{23.46} & 16.40 & 42.66  \\ 
        \cline{2-10}
        & SparseGPT                     & 62.69 & 52.71 & 31.66 & 57.06 & 40.74 & 20.56 & 15.00 & 40.06  \\
        & AlphaPruning $w.$ SparseGPT   & 64.19 & \bf{53.07} & \bf{37.44} & \bf{64.88} & 45.79 & \bf{26.45} & 17.40 & 44.17  \\
        \rowcolor{LightGrey}
        & Ours $w.$ SparseGPT   & \bf{65.69} & 52.71 & 36.89 & 64.72 & \bf{47.43} & 26.02 & \bf{17.80} & \bf{44.47}   \\

        \midrule
        & Magnitude                          & 38.50 & \bf{53.43} & 25.95 & 48.86 & 25.51 & 22.10 & 13.40 & 32.53   \\
        & AlphaPruning $w$. Magnitude        & 53.30 & 51.26 & 26.67 & 51.70 & 26.60 & 23.38 & 16.20 & 35.59    \\
        \rowcolor{LightGrey}
        & Ours $w$. Magnitude         & \bf{59.79} & 48.38 & \bf{28.32} & \bf{54.38} & \bf{27.10} & \bf{24.23} & \bf{16.40} & \bf{36.94}  \\ 
        \cline{2-10}
        & Wanda                          & 37.86 & 52.71 & 26.64 & 48.30 & 27.15 & 19.97 & 13.20 & 32.26  \\
        0.8 & AlphaPruning $w$. Wanda        & 62.17 & 52.71 & 29.59 & 55.09 & 34.93 & 20.22 & 14.20 & 38.42    \\
        \rowcolor{LightGrey}
        & Ours $w$. Wanda         & \bf{62.17} & \bf{52.71} & \bf{30.45} & \bf{57.38} & \bf{36.41} & \bf{21.08} & \bf{14.40} & \bf{39.23}  \\ 
        \cline{2-10}
        & SparseGPT                     & 60.80 & 52.71 & 28.71 & 50.28 & 30.22 & 18.26 & 13.00 & 36.28  \\
        & AlphaPruning $w.$ SparseGPT   & 62.20 & 52.71 & 31.46 & 57.62 & 35.48 & 19.62 & 14.40 & 39.07  \\
        \rowcolor{LightGrey}
        & Ours $w.$ SparseGPT        & \bf{62.20} & \bf{52.71} & \bf{32.37} & \bf{59.19} & \bf{37.16} & \bf{22.18} & \bf{15.40} & \bf{40.18}   \\

        \midrule
        & Magnitude                          & 38.29 & \bf{54.51} & \bf{25.95} & 49.96 & 25.21 & \bf{23.12} & 14.00 & 33.01   \\
        & AlphaPruning $w$. Magnitude        & 39.11 & 51.26 & 25.47 & \bf{50.83} & \bf{26.39} & 22.10 & \bf{16.60} & 33.11     \\
        \rowcolor{LightGrey}
        & Ours $w$. Magnitude               & \bf{41.22} & 51.62 & 25.73 & 50.43 & 26.01 & 22.01 & 16.00 & \bf{33.29}    \\ 
        \cline{2-10}
        & Wanda                          & 37.83 & 52.71 & 26.09 & 47.91 & 25.88 & \bf{21.16} & \bf{12.80} & 32.05  \\
        0.85 & AlphaPruning $w$. Wanda        & 37.83 & 52.71 & \bf{26.72} & \bf{49.80} & 26.09 & 19.62 & 11.60 & 32.05    \\
        \rowcolor{LightGrey}
        & Ours $w$. Wanda        & \bf{38.65} & \bf{53.79} & 26.31 & 48.46 & \bf{26.77} & 20.31 & 12.60 & \bf{32.41}  \\ 
        \cline{2-10}
        & SparseGPT                     & 38.41 & 52.71 & 27.31 & 50.04 & 26.30 & 17.66 & 12.60 & 32.15  \\
        & AlphaPruning $w.$ SparseGPT   & \bf{62.17} & 52.71 & \bf{28.64} & 51.70 & 29.67 & \bf{18.86} & \bf{13.40} & 36.73  \\
        \rowcolor{LightGrey}
        & Ours $w.$ SparseGPT        & 62.17 & \bf{52.71} & 28.59 & \bf{54.14} & \bf{30.09} & 18.77 & 12.80 & \bf{37.04}   \\ 
        
        \bottomrule
    \end{tabular}
    }
    \label{table:llama13b_pruning_zeroshot_details}
\end{table*}

\subsection{Detailed \AlphaHill Distribution}

\begin{figure*}[!h]
    \centering

    \begin{subfigure}{0.19\linewidth}
        \includegraphics[width=\linewidth]{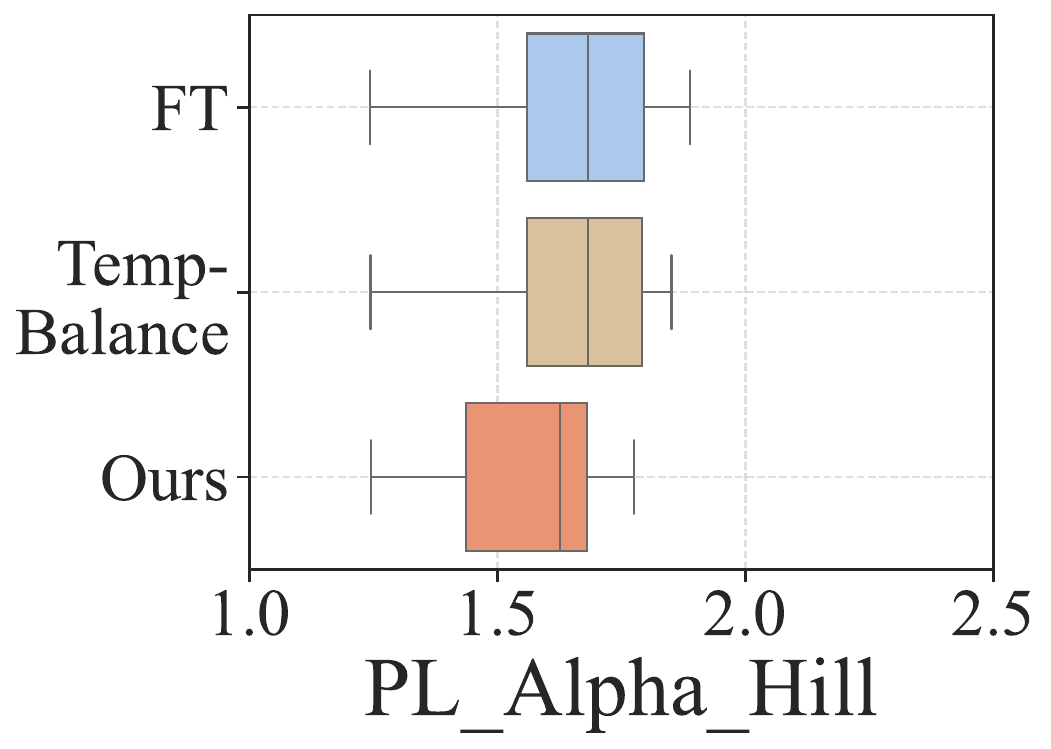}
        \caption{DPOT-Tiny, 5\%}
    \end{subfigure}
    \begin{subfigure}{0.19\linewidth}
        \includegraphics[width=\linewidth]{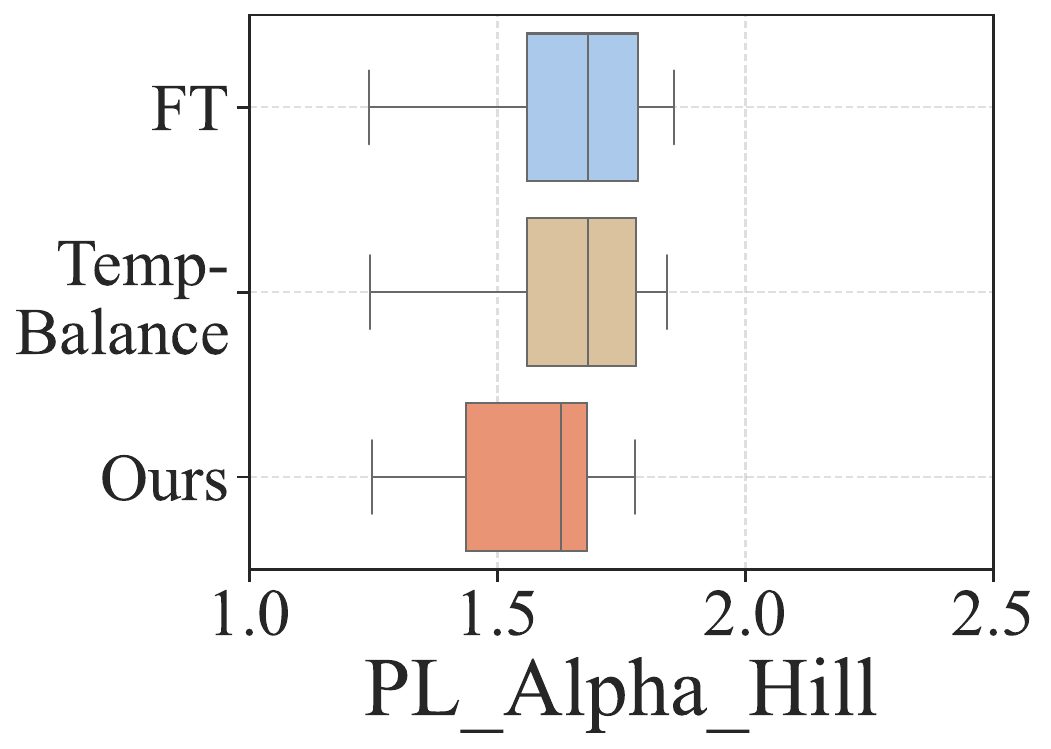}
        \caption{DPOT-Tiny, 10\%}
    \end{subfigure}
    \begin{subfigure}{0.19\linewidth}
        \includegraphics[width=\linewidth]{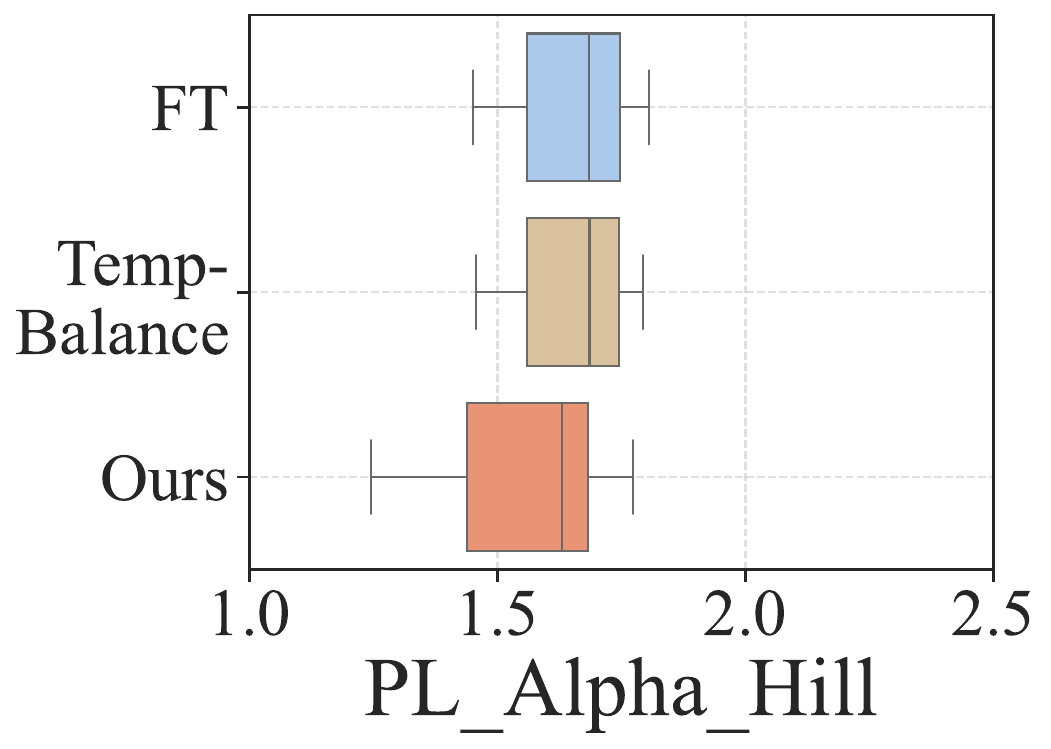}
        \caption{DPOT-Tiny, 25\%}
    \end{subfigure}
    \begin{subfigure}{0.19\linewidth}
        \includegraphics[width=\linewidth]{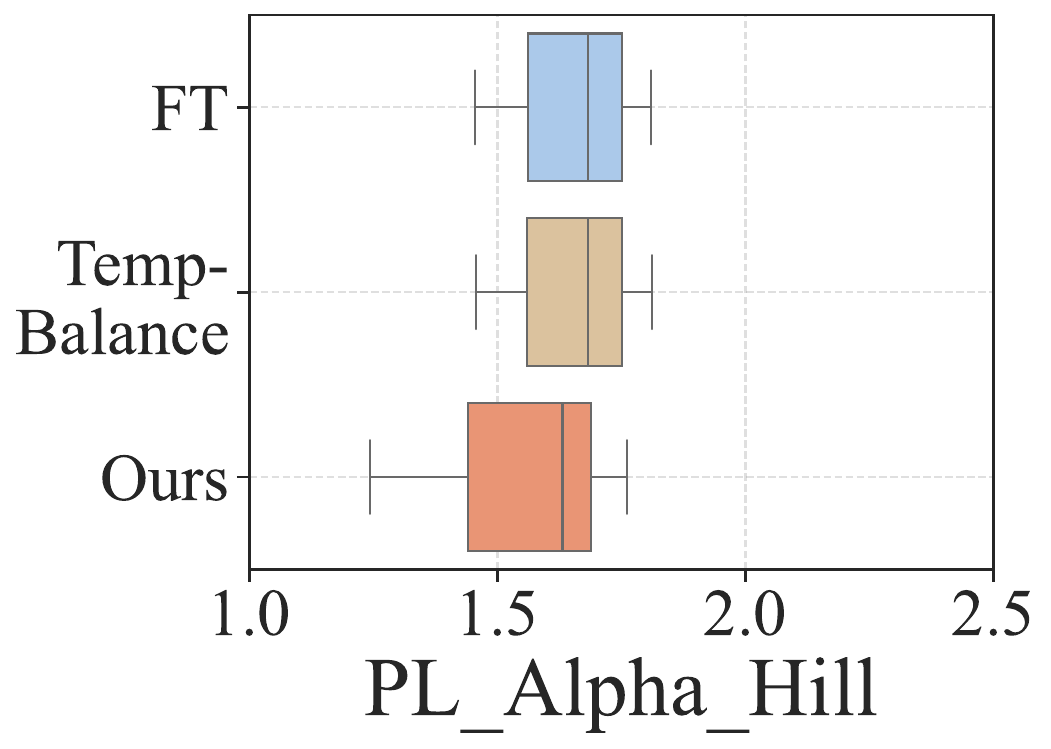}
        \caption{DPOT-Tiny, 50\%}
    \end{subfigure}
    \begin{subfigure}{0.19\linewidth}
        \includegraphics[width=\linewidth]{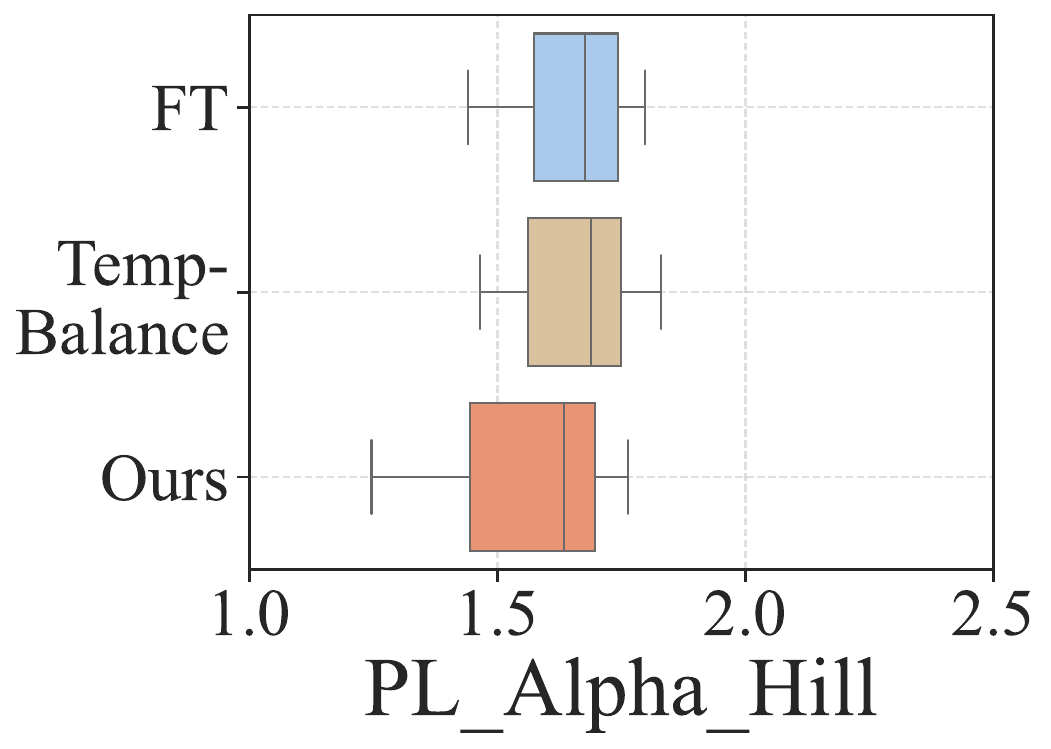}
        \caption{DPOT-Tiny, 100\%}
    \end{subfigure}

    \begin{subfigure}{0.19\linewidth}
        \includegraphics[width=\linewidth]{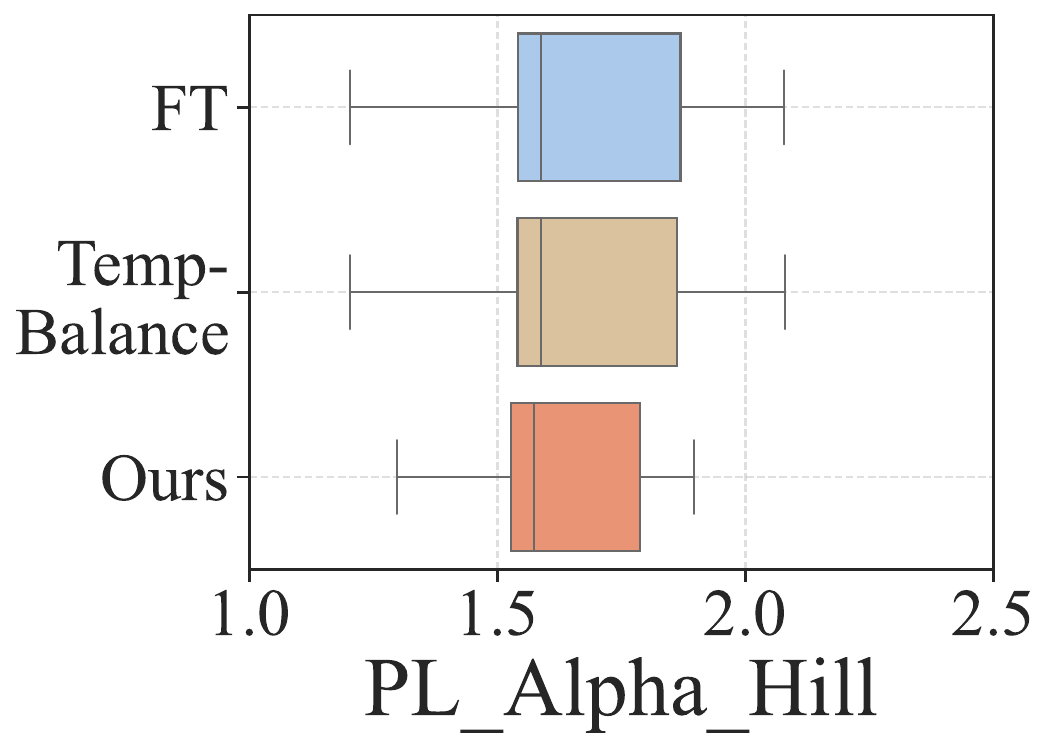}
        \caption{DPOT-Small, 5\%}
    \end{subfigure}
    \begin{subfigure}{0.19\linewidth}
        \includegraphics[width=\linewidth]{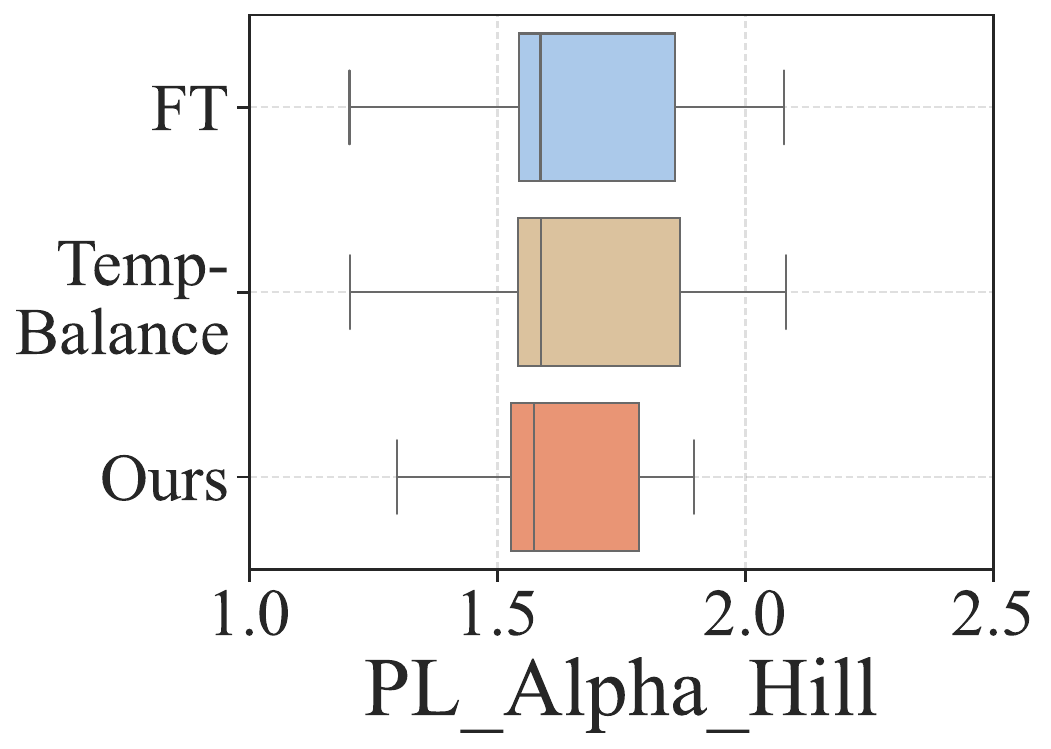}
        \caption{DPOT-Small, 10\%}
    \end{subfigure}
    \begin{subfigure}{0.19\linewidth}
        \includegraphics[width=\linewidth]{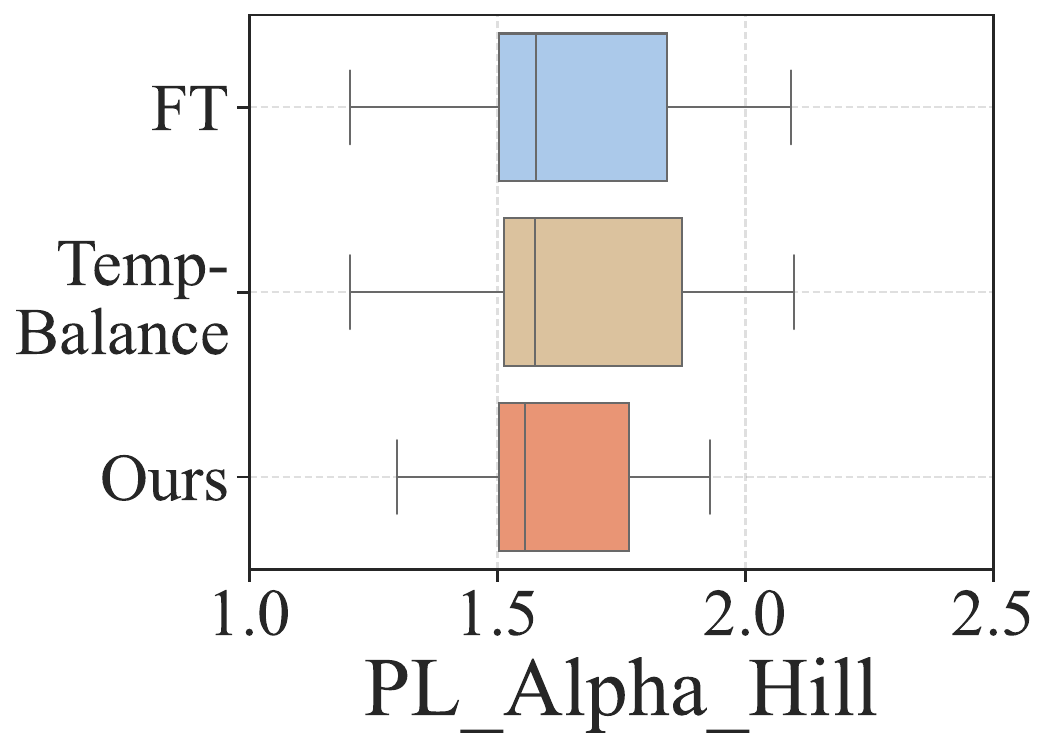}
        \caption{DPOT-Small, 25\%}
    \end{subfigure}
    \begin{subfigure}{0.19\linewidth}
        \includegraphics[width=\linewidth]{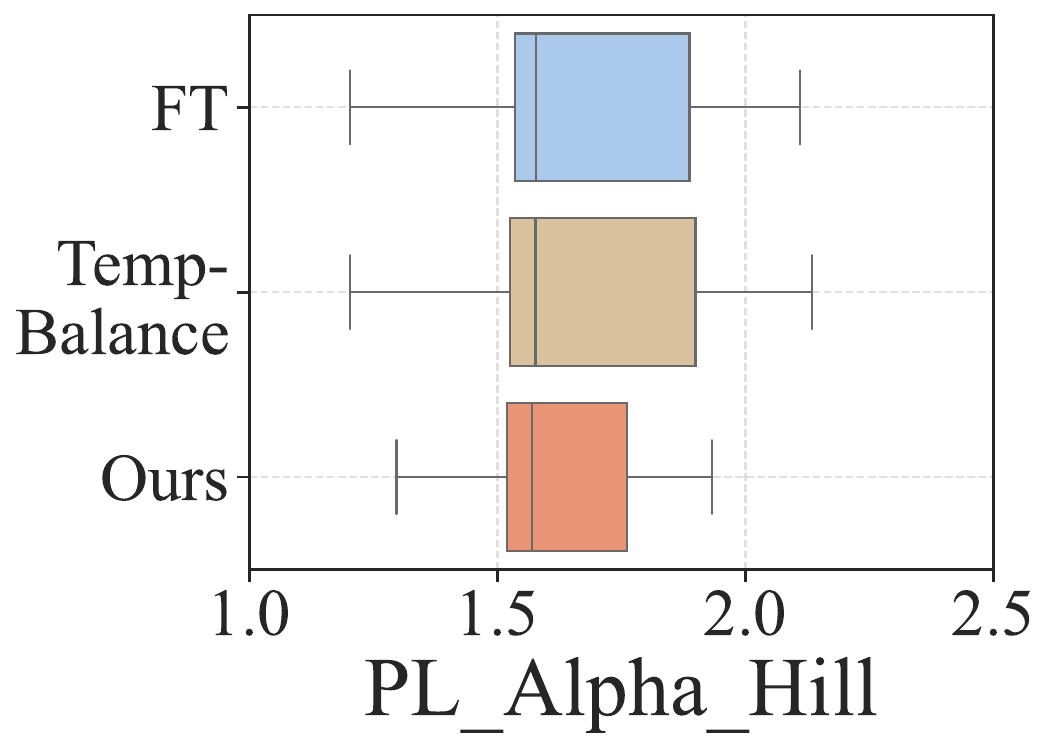}
        \caption{DPOT-Small, 50\%}
    \end{subfigure}
    \begin{subfigure}{0.19\linewidth}
        \includegraphics[width=\linewidth]{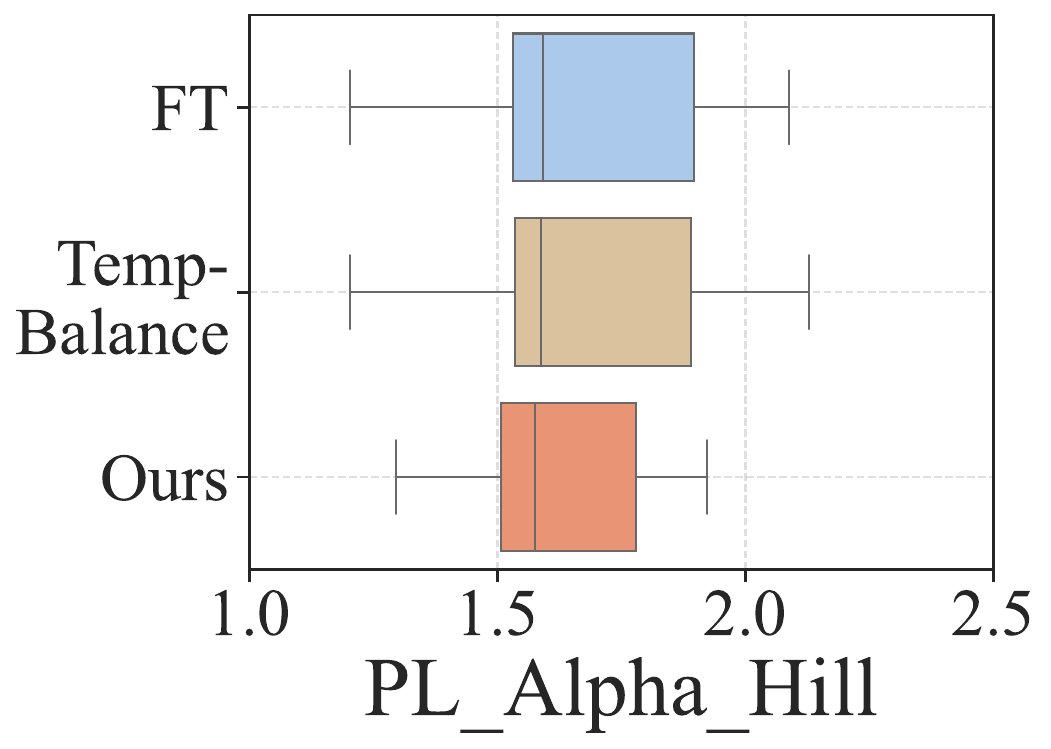}
        \caption{DPOT-Small, 100\%}
    \end{subfigure}
    \caption{Comparing the distribution of \AlphaHill  of DPOT-Tiny and DPOT-Small in different fine-tuning methods and data ratios.}
    \label{fig:sciml_alpha_bar_all}
\end{figure*}

We provide additional experimental results on LLM pruning and SciML fine-tuning to support the \AlphaHill distribution analysis presented in Section \ref{wwsampling: HT_metrics_analysis}. The results in Figure \ref{fig:sciml_alpha_bar_all} and \ref{fig:llama_alpha_bar_all} show that \ourmethod consistently helps models achieve better layer quality across different experimental settings. 

\ificml
\begin{figure*}[!h]
    \centering
    
    \begin{subfigure}{0.2\linewidth}
        \includegraphics[width=\linewidth]{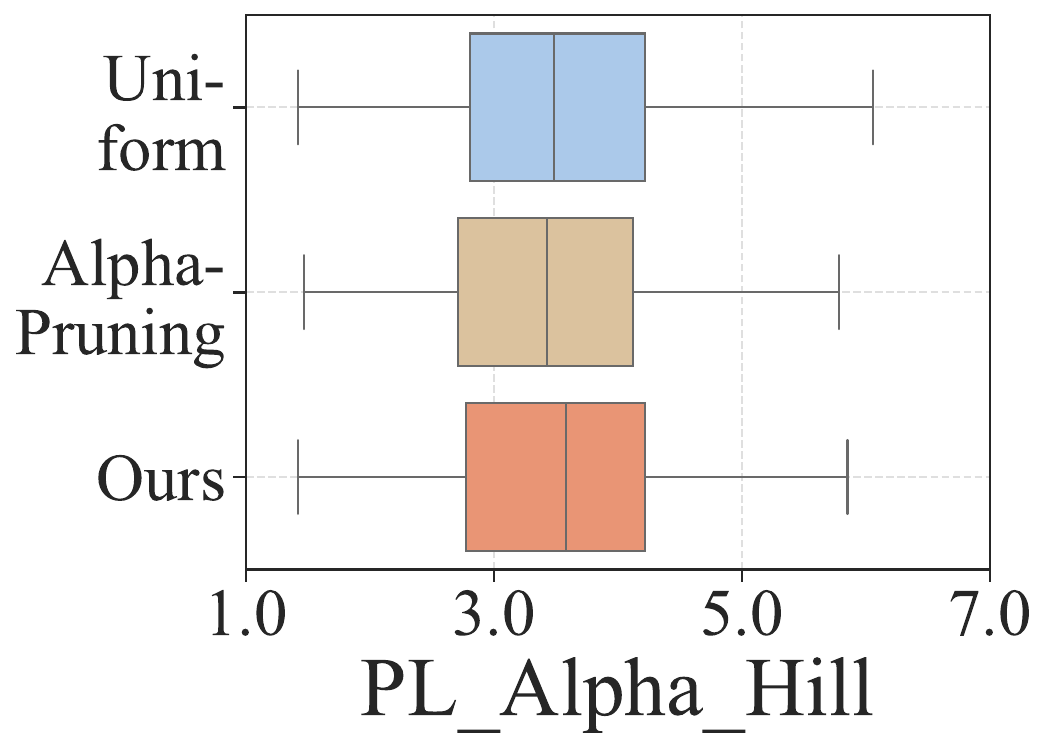}
        \caption{LLaMA-7B, SR=0.7}
    \end{subfigure}
    \begin{subfigure}{0.2\linewidth}
        \includegraphics[width=\linewidth]{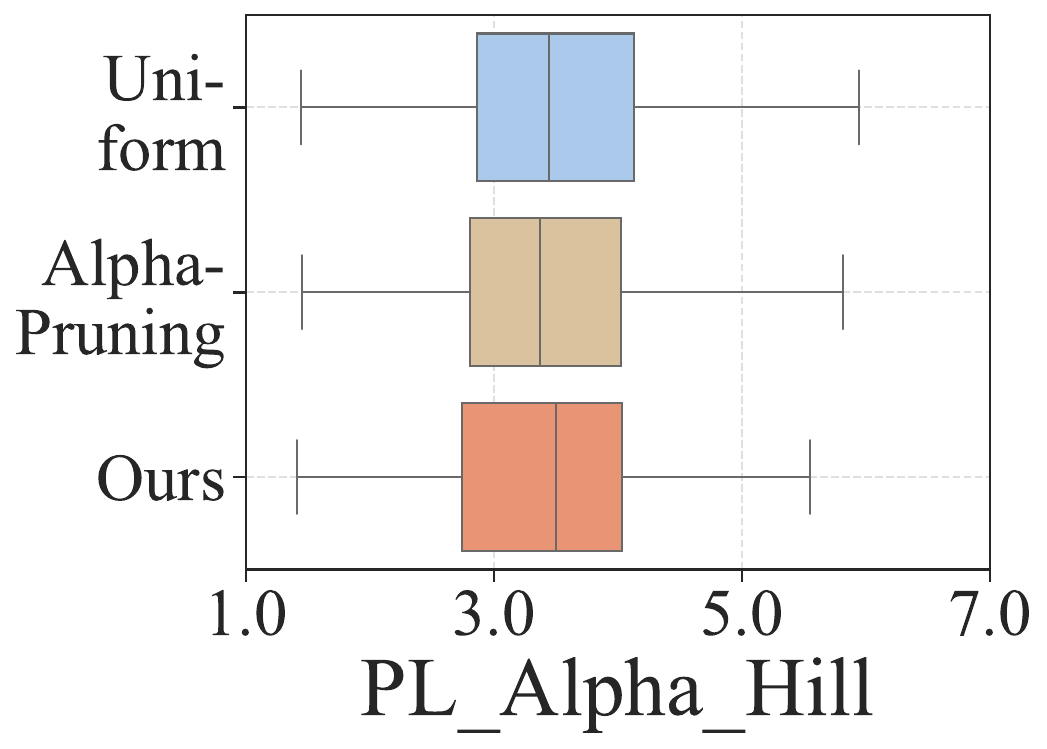}
        \caption{LLaMA-7B, SR=0.75}
    \end{subfigure}
    \begin{subfigure}{0.2\linewidth}
        \includegraphics[width=\linewidth]{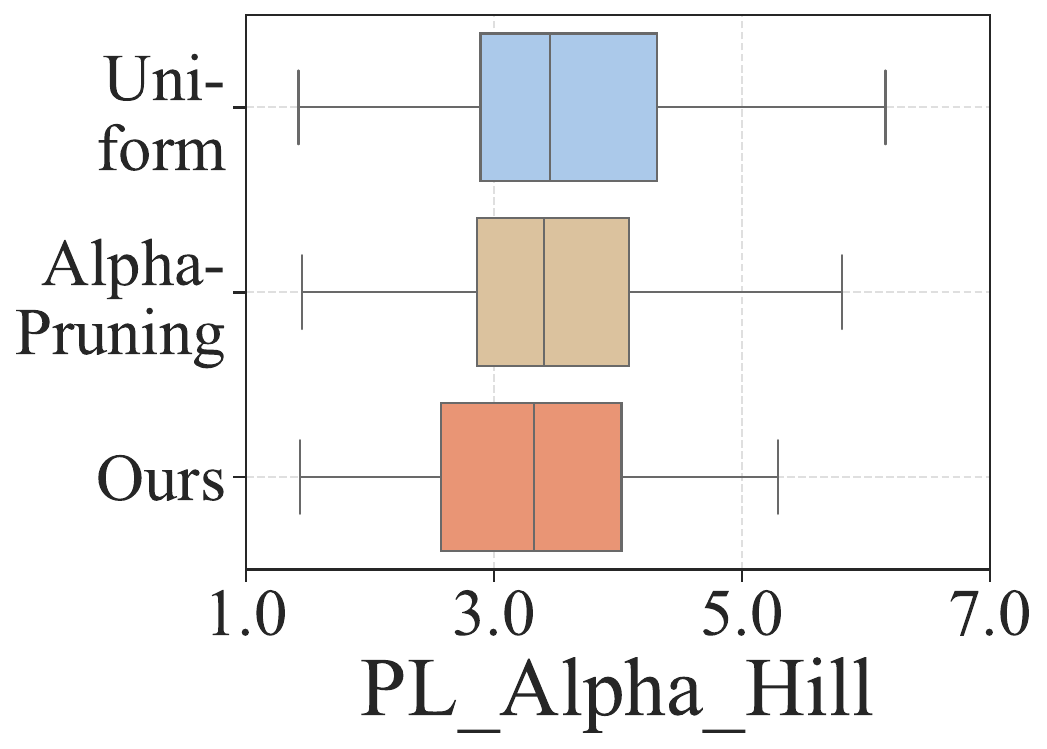}
        \caption{LLaMA-7B, SR=0.8}
    \end{subfigure}
    \begin{subfigure}{0.2\linewidth}
        \includegraphics[width=\linewidth]{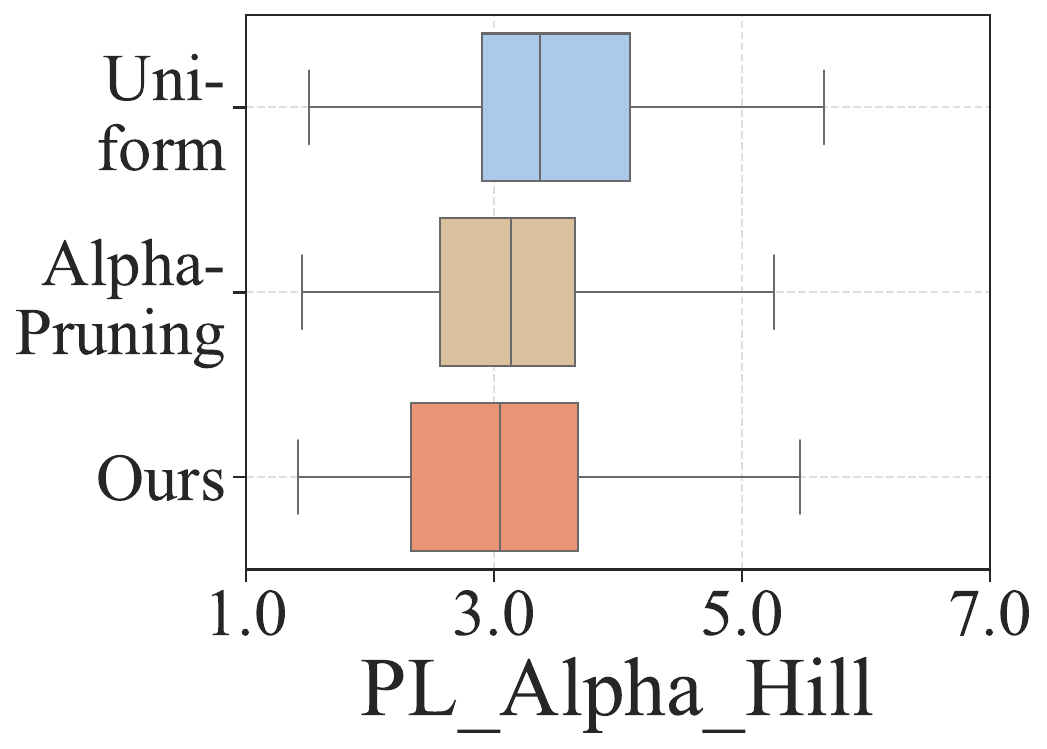}
        \caption{LLaMA-7B, SR=0.85}
    \end{subfigure}

    \begin{subfigure}{0.2\linewidth}
        \includegraphics[width=\linewidth]{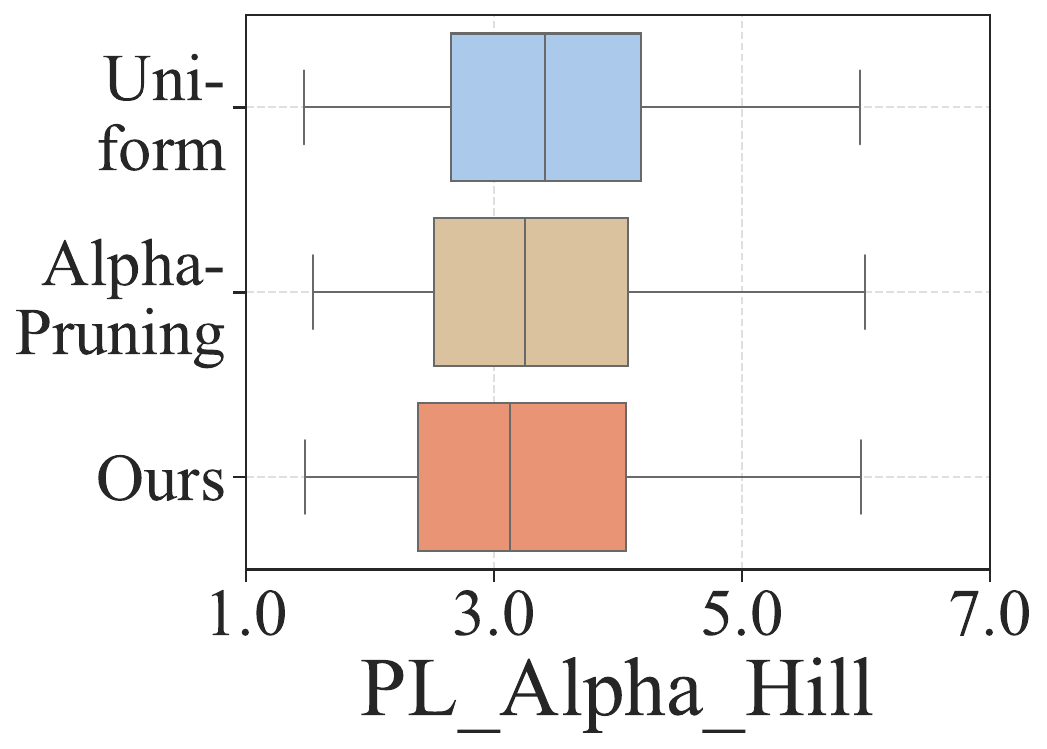}
        \caption{LLaMA-13B, SR=0.7}
    \end{subfigure}
    \begin{subfigure}{0.2\linewidth}
        \includegraphics[width=\linewidth]{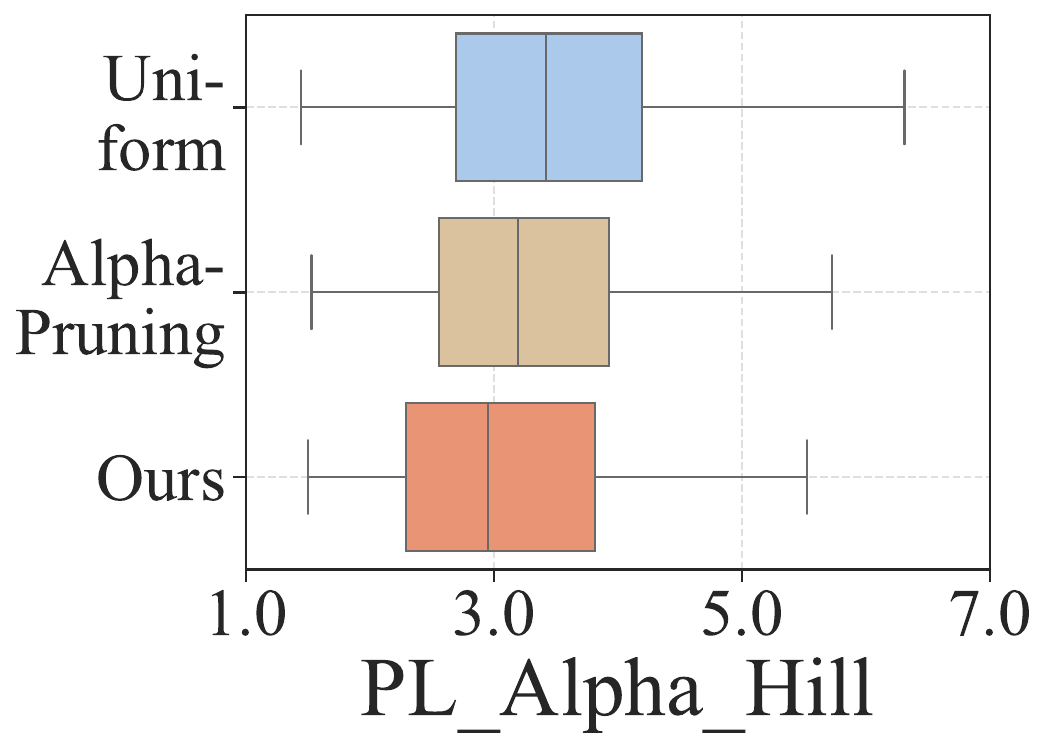}
        \caption{LLaMA-13B, SR=0.75}
    \end{subfigure}
    \begin{subfigure}{0.2\linewidth}
        \includegraphics[width=\linewidth]{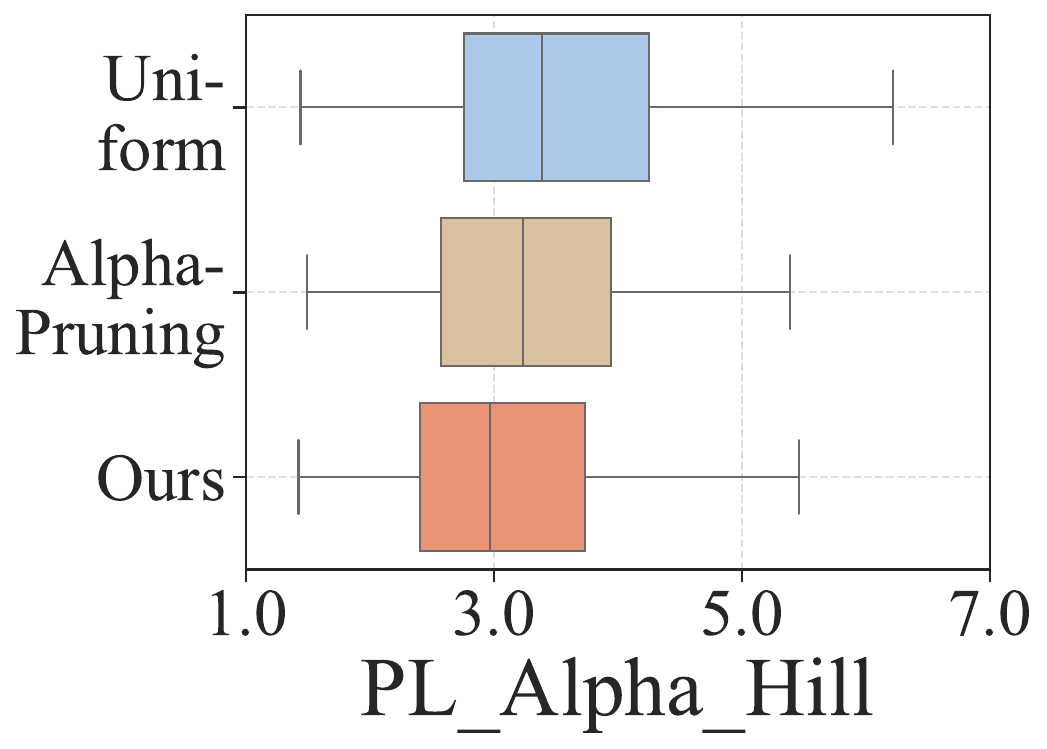}
        \caption{LLaMA-13B, SR=0.8}
    \end{subfigure}
    \begin{subfigure}{0.2\linewidth}
        \includegraphics[width=\linewidth]{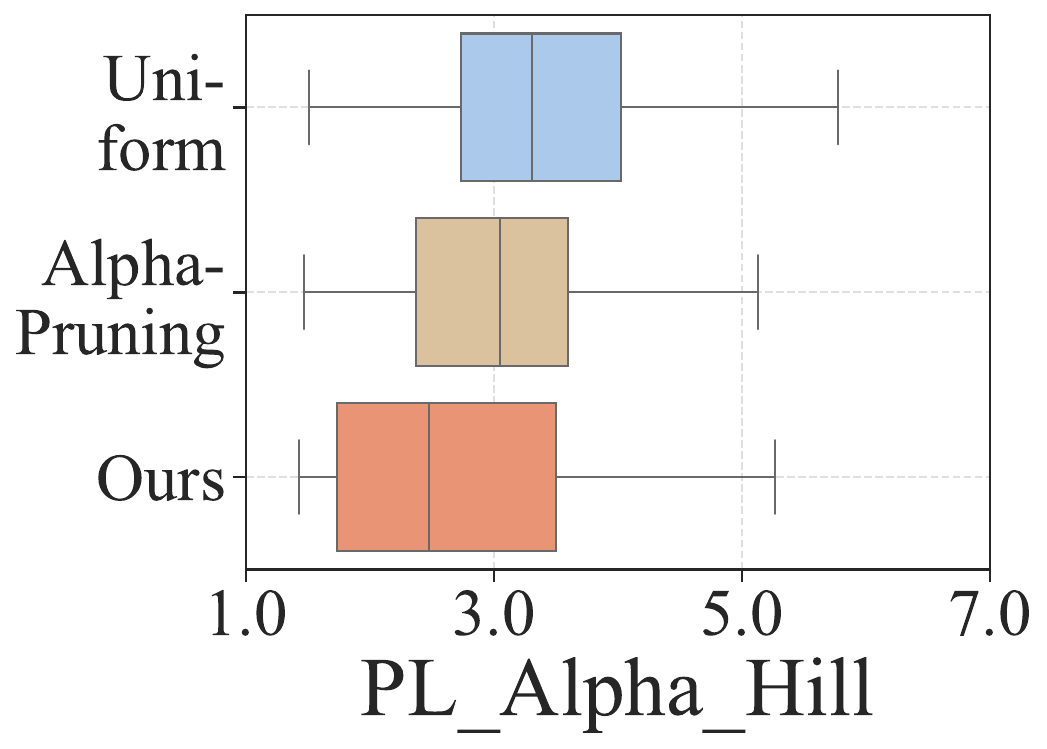}
        \caption{LLaMA-13B, SR=0.85}
    \end{subfigure}
    
    \caption{Comparing the distribution of \AlphaHill  of LLaMA-7B and LLaMA-13B in different layer-wise strategies and sparsity ratios (shown as "SR"). The pruning method is SparseGPT.} 
    \label{fig:llama_alpha_bar_all}
\end{figure*}
\else
\begin{figure*}[!h]
    \centering
    
    \begin{subfigure}{0.24\linewidth}
        \includegraphics[width=\linewidth]{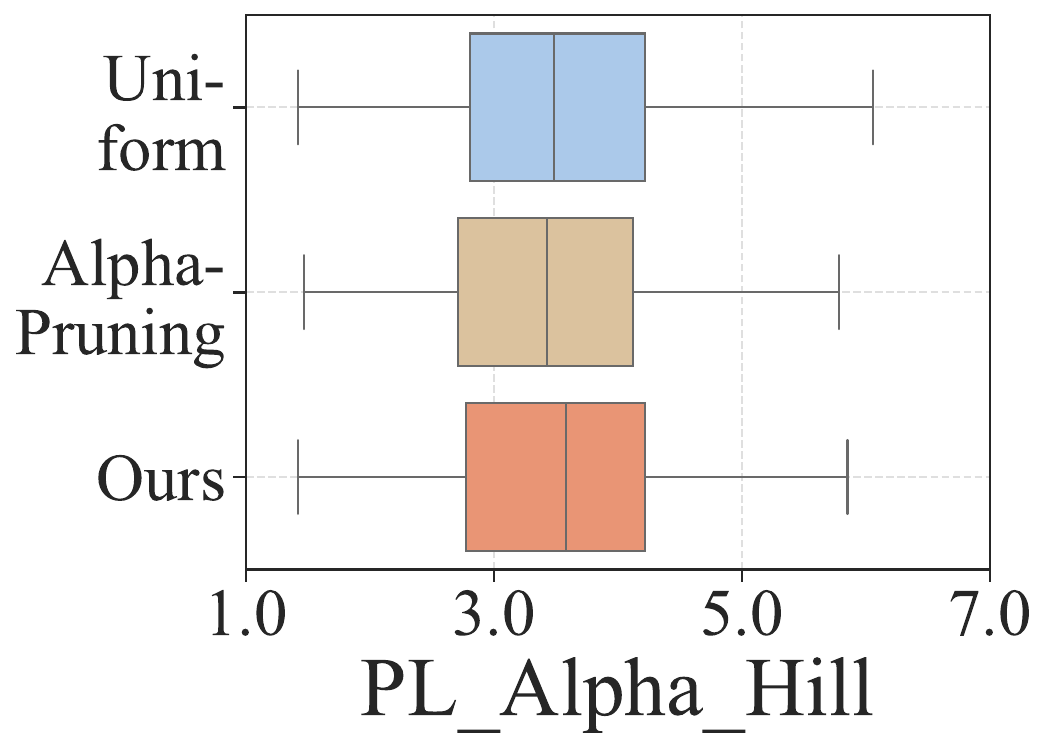}
        \caption{LLaMA-7B, SR=0.7}
    \end{subfigure}
    \begin{subfigure}{0.24\linewidth}
        \includegraphics[width=\linewidth]{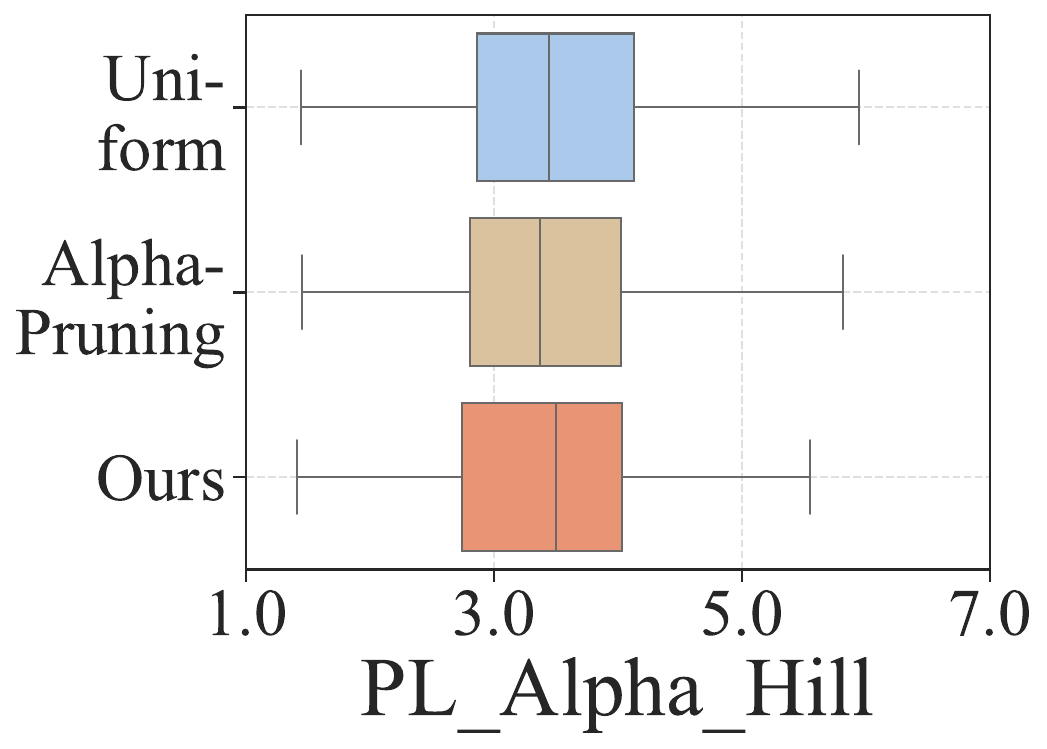}
        \caption{LLaMA-7B, SR=0.75}
    \end{subfigure}
    \begin{subfigure}{0.24\linewidth}
        \includegraphics[width=\linewidth]{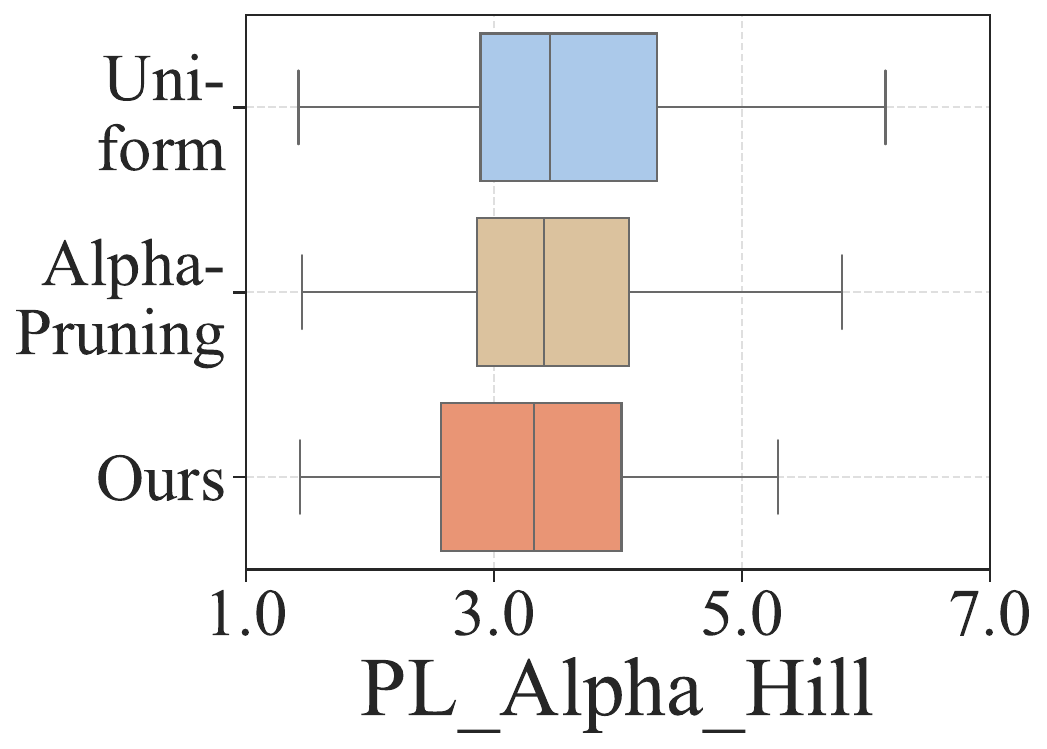}
        \caption{LLaMA-7B, SR=0.8}
    \end{subfigure}
    \begin{subfigure}{0.24\linewidth}
        \includegraphics[width=\linewidth]{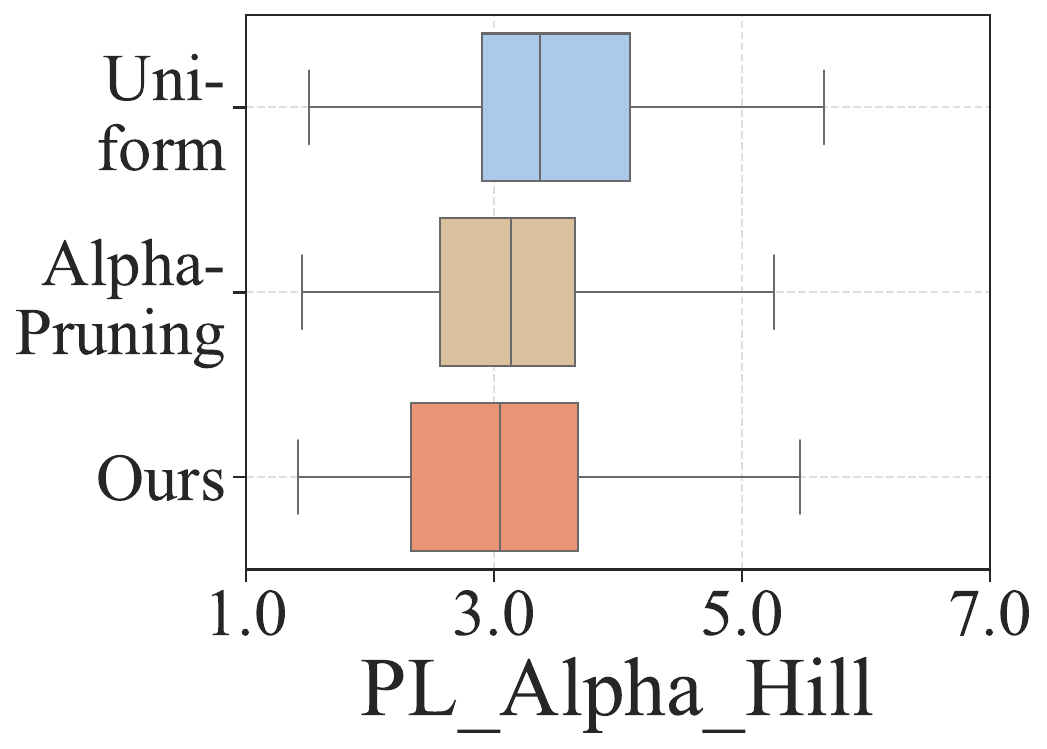}
        \caption{LLaMA-7B, SR=0.85}
    \end{subfigure}

    \begin{subfigure}{0.24\linewidth}
        \includegraphics[width=\linewidth]{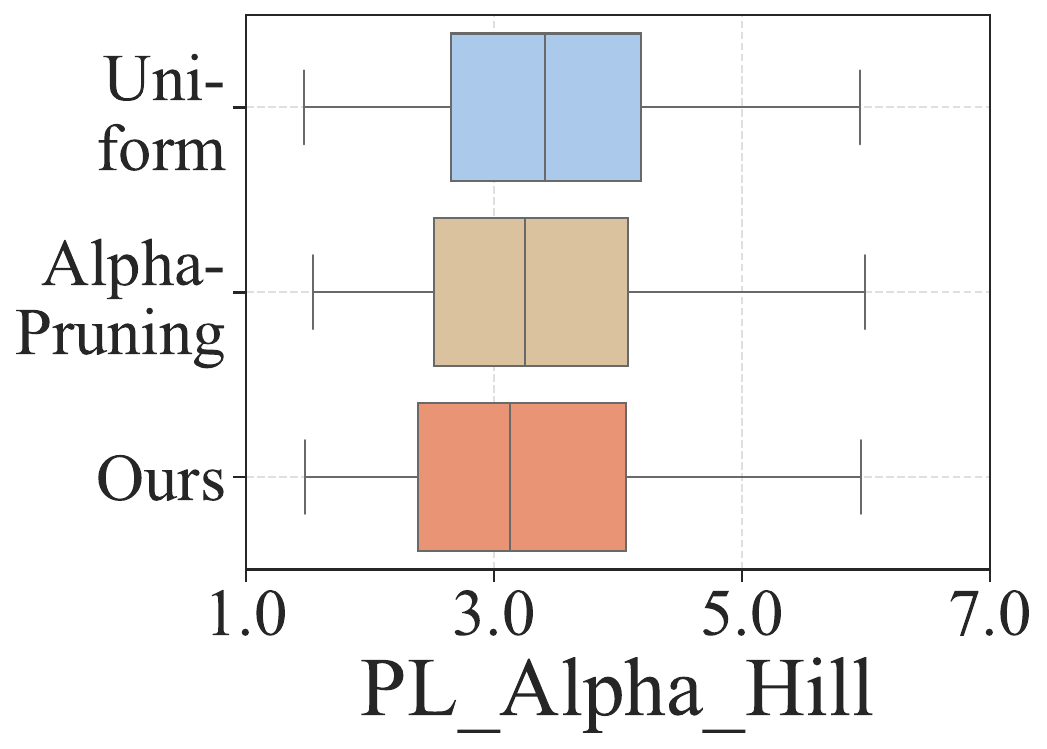}
        \caption{LLaMA-13B, SR=0.7}
    \end{subfigure}
    \begin{subfigure}{0.24\linewidth}
        \includegraphics[width=\linewidth]{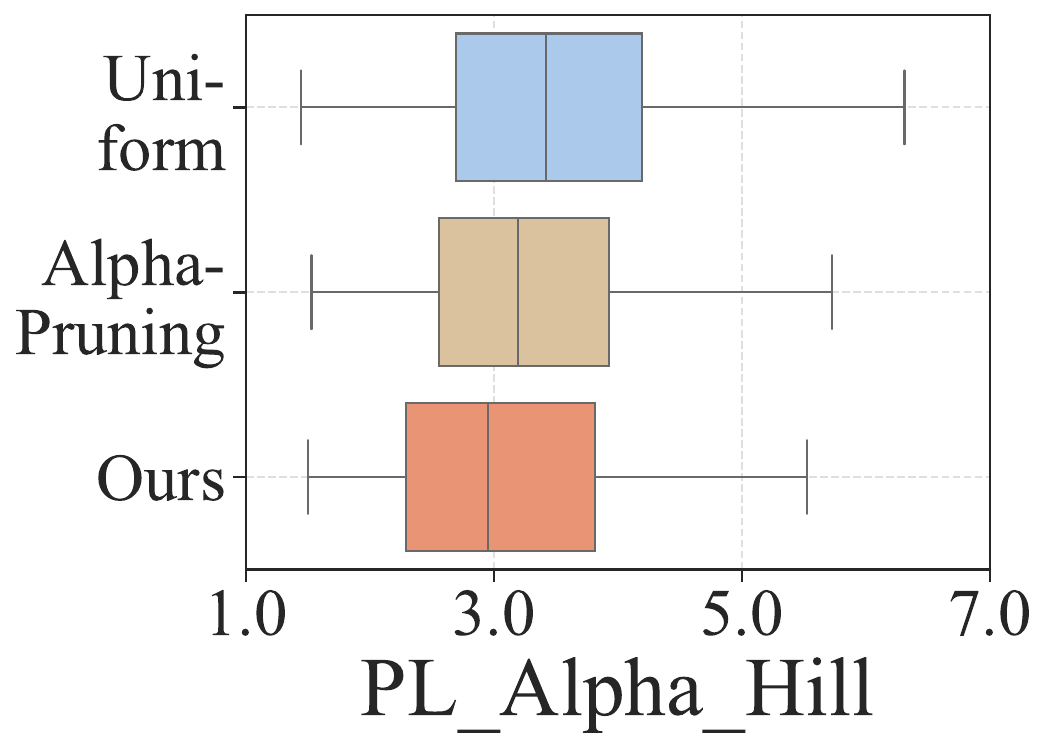}
        \caption{LLaMA-13B, SR=0.75}
    \end{subfigure}
    \begin{subfigure}{0.24\linewidth}
        \includegraphics[width=\linewidth]{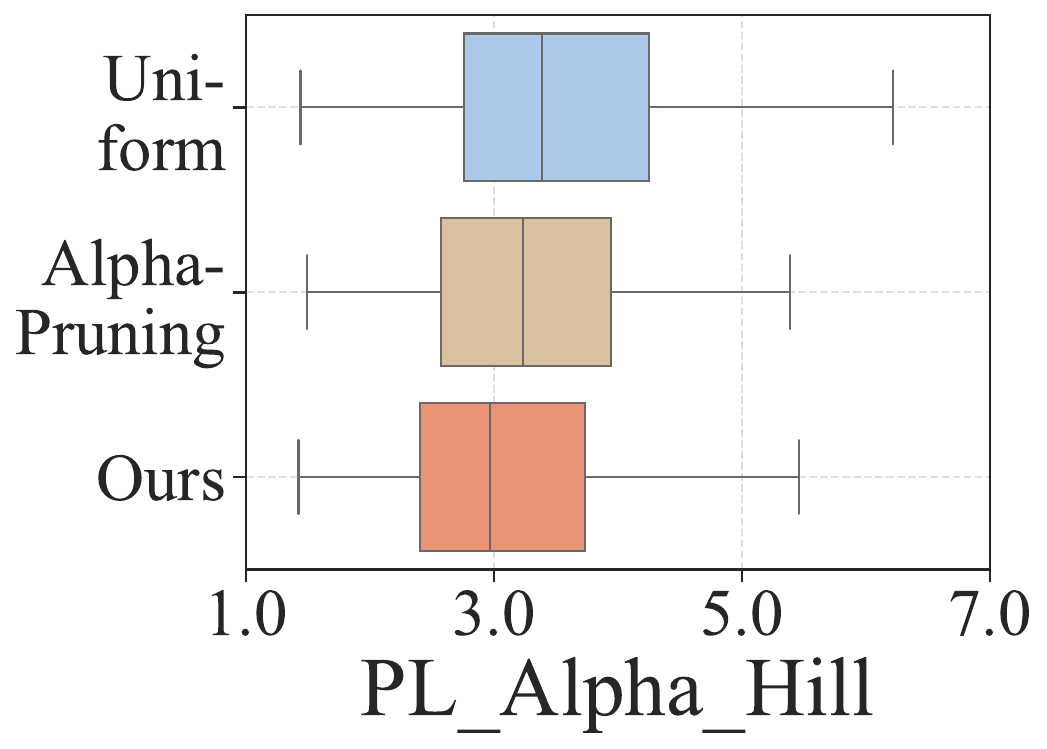}
        \caption{LLaMA-13B, SR=0.8}
    \end{subfigure}
    \begin{subfigure}{0.24\linewidth}
        \includegraphics[width=\linewidth]{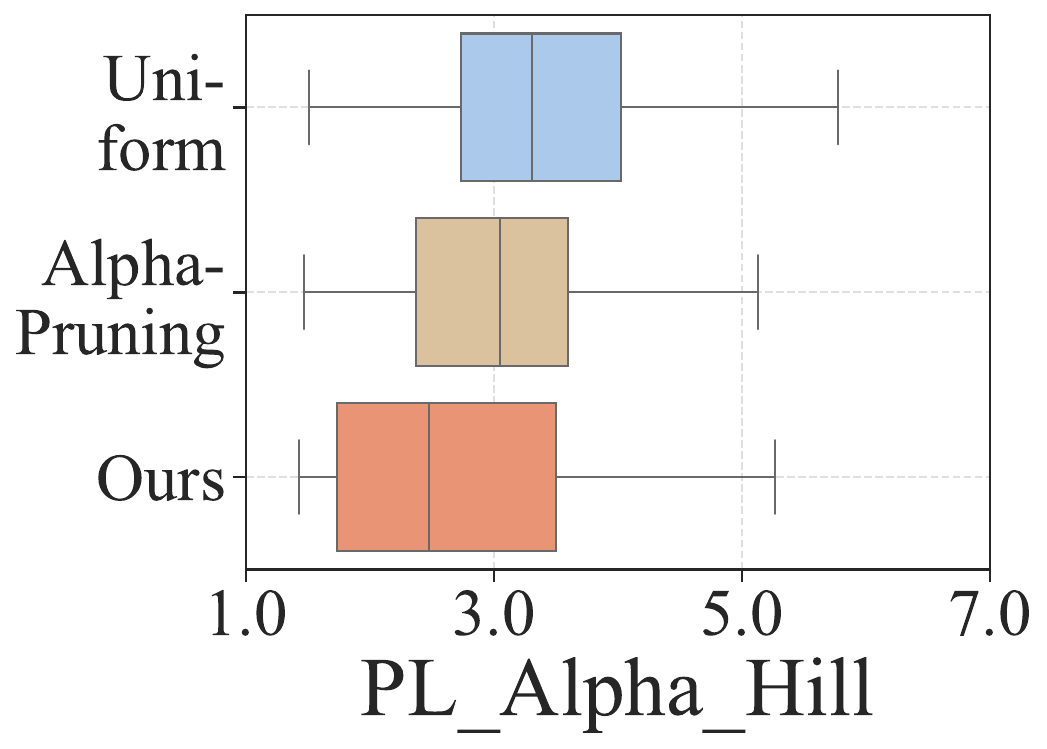}
        \caption{LLaMA-13B, SR=0.85}
    \end{subfigure}
    
    \caption{Comparing the distribution of \AlphaHill  of LLaMA-7B and LLaMA-13B in different layer-wise strategies and sparsity ratios (shown as "SR"). The pruning method is SparseGPT.} 
    \label{fig:llama_alpha_bar_all}
\end{figure*}
\fi

\subsection{Layer-wise Visualization over Training}

\begin{figure}[!tbh]
    \centering
    \begin{subfigure}{0.33\linewidth}
        \centering
        \includegraphics[width=\textwidth]{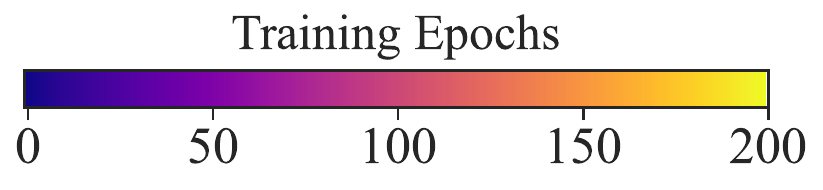}
    \end{subfigure}

    \begin{subfigure}{0.24\linewidth}
        \centering
        \includegraphics[width=\textwidth]{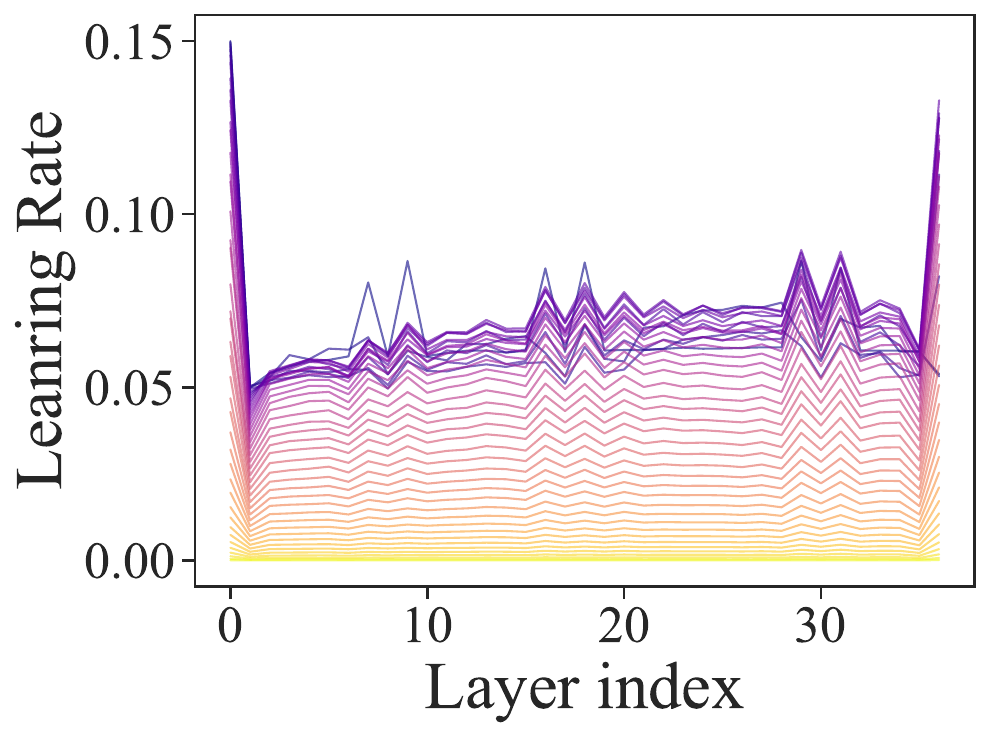}
    \end{subfigure}
    \hfill
    \begin{subfigure}{0.24\linewidth}
        \centering
        \includegraphics[width=\textwidth]{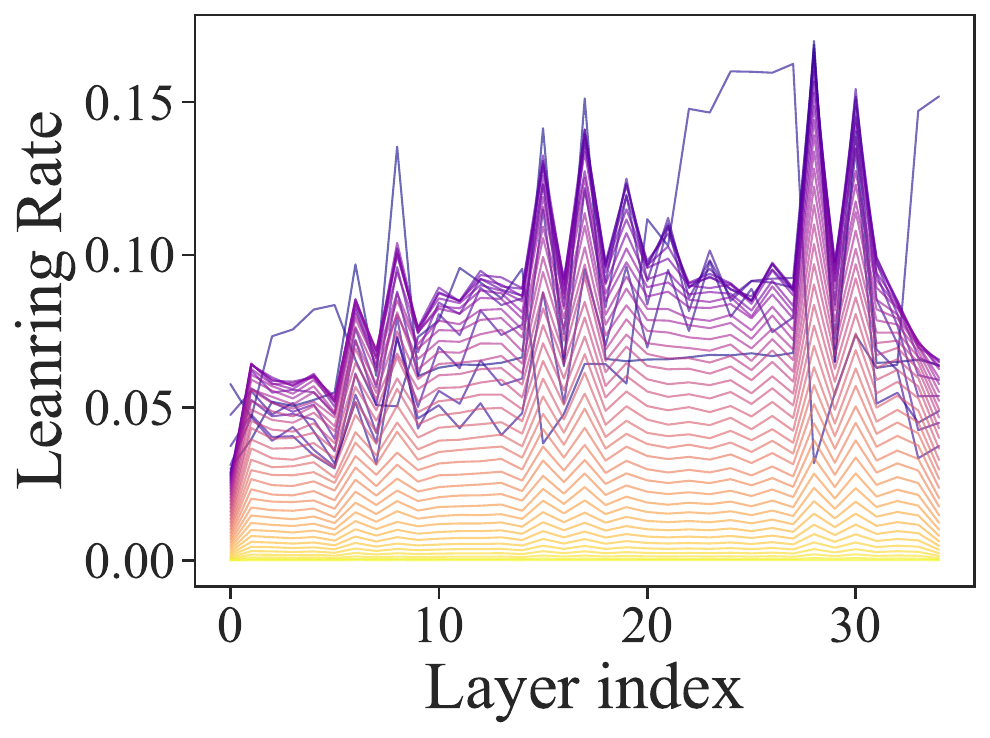}
    \end{subfigure}
    \hfill
    \begin{subfigure}{0.24\linewidth}
        \centering
        \includegraphics[width=\linewidth]{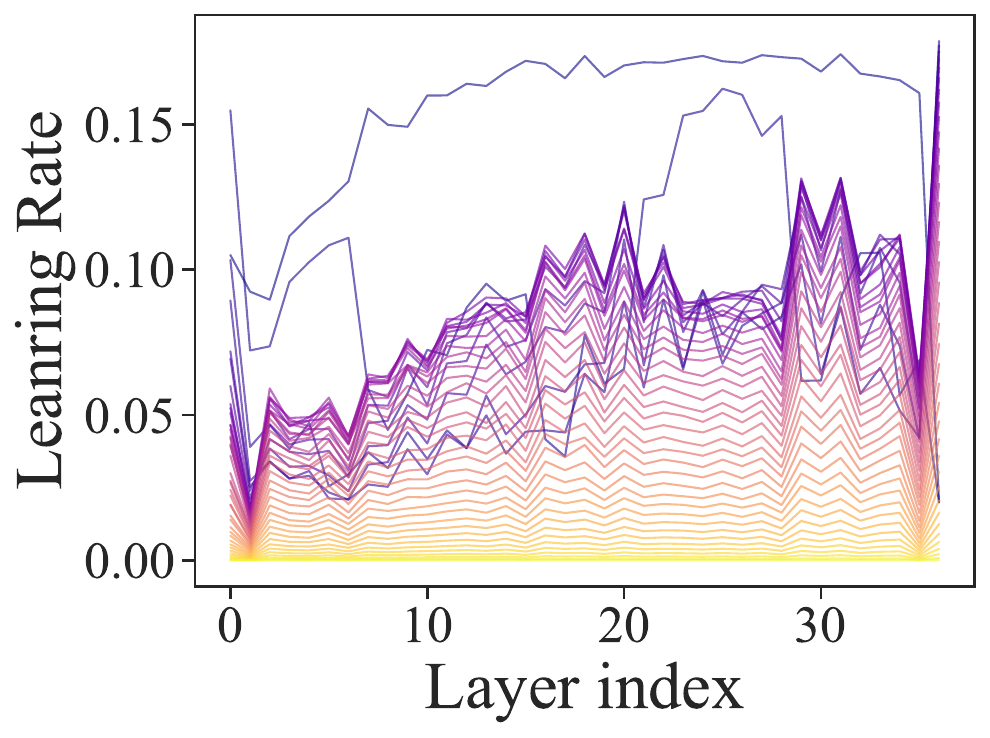}
    \end{subfigure}
    \hfill
    \begin{subfigure}{0.24\linewidth}
        \centering
        \includegraphics[width=\linewidth]{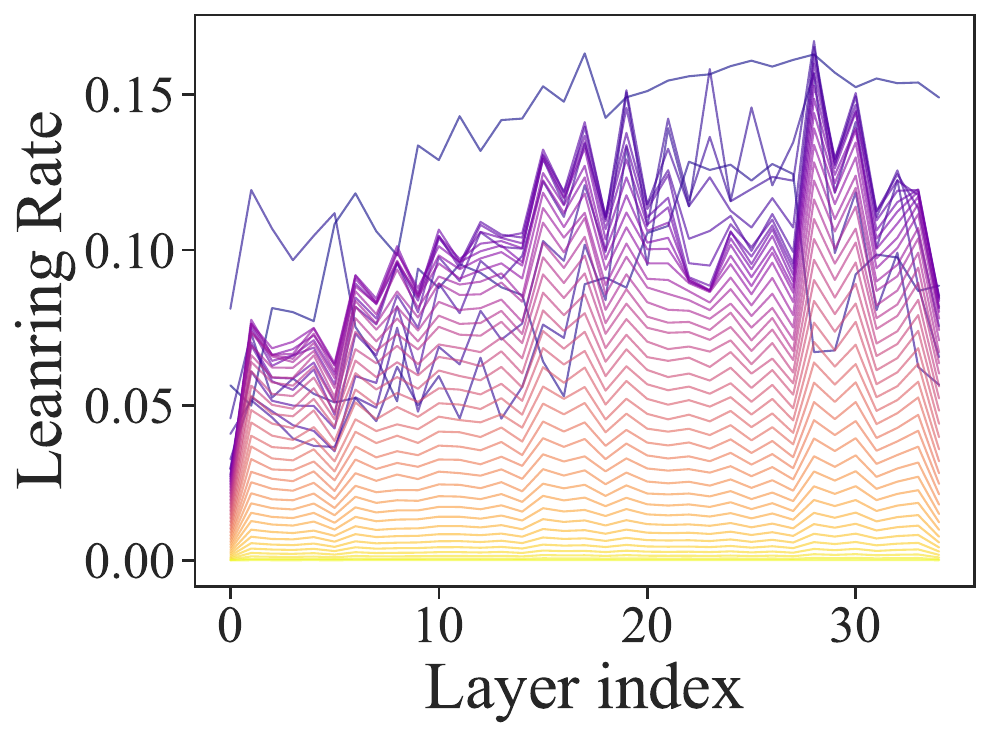}
    \end{subfigure}

    \begin{subfigure}{0.24\linewidth}
        \centering
        \includegraphics[width=\textwidth]{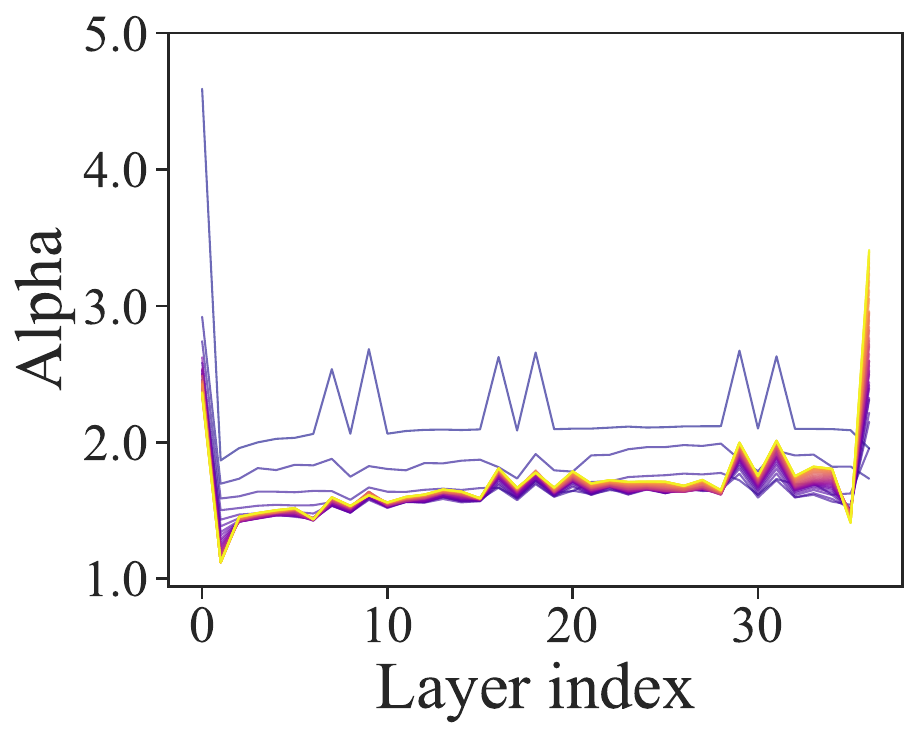}
        \caption{ResNet 34, TB(no LS)}
        \label{subfig:resnet34_tb}
    \end{subfigure}
    \hfill
    \begin{subfigure}{0.24\linewidth}
        \centering
        \includegraphics[width=\textwidth]{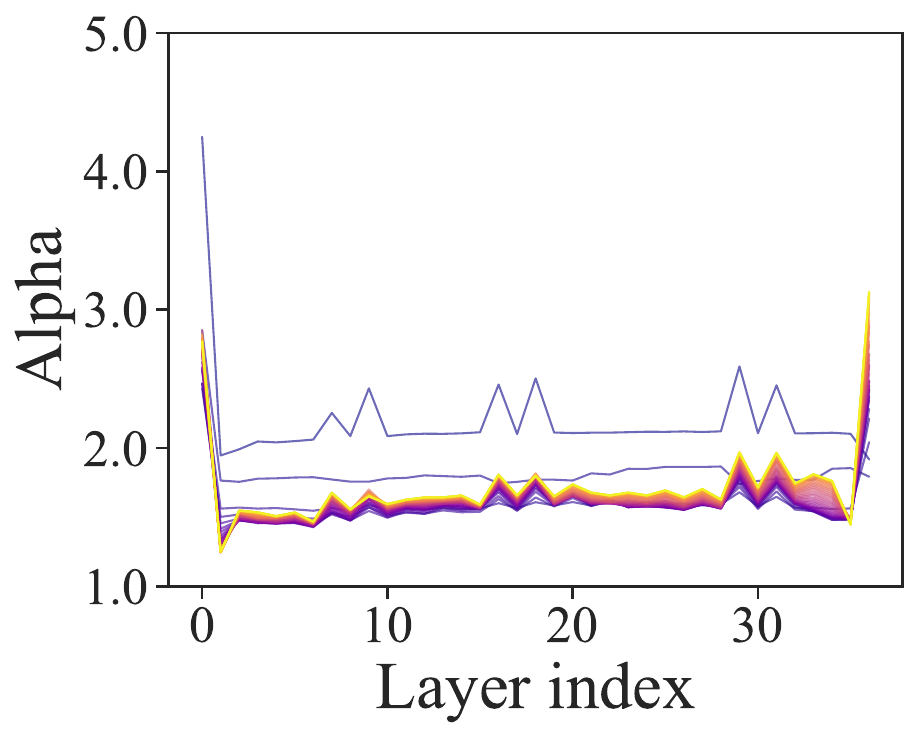}
        \caption{ResNet 34, LS+TB}
        \label{subfig:resnet34_lstb}
    \end{subfigure}
    \hfill
    \begin{subfigure}{0.24\linewidth}
        \centering
        \includegraphics[width=\linewidth]{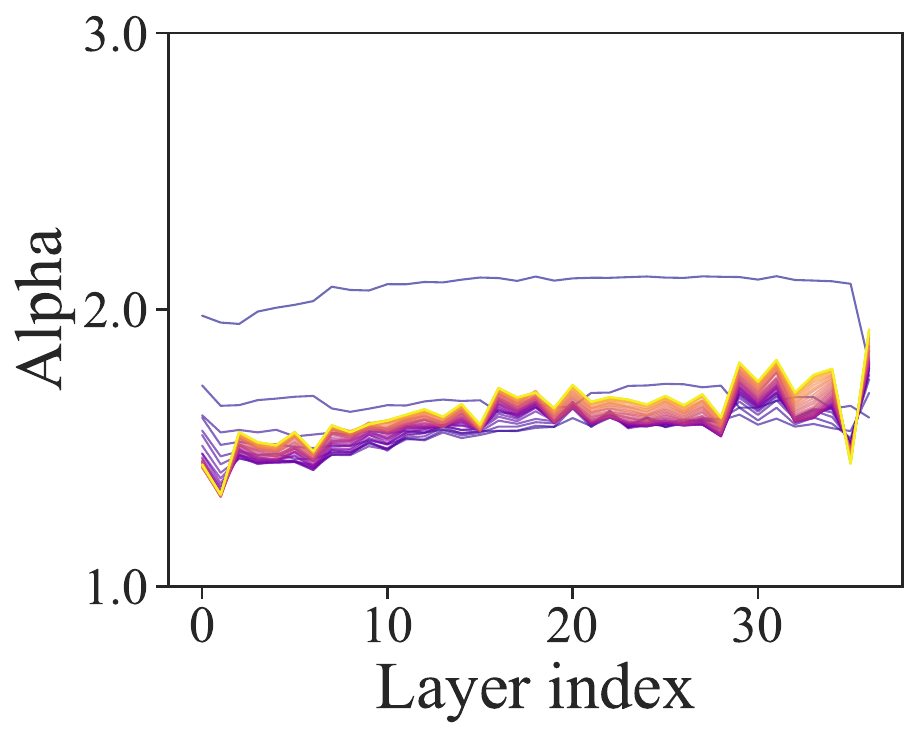}
        \caption{ResNet 34, Ours(no LS)}
        \label{subfig:resnet34_ours}
    \end{subfigure}
    \hfill
    \begin{subfigure}{0.24\linewidth}
        \centering
        \includegraphics[width=\linewidth]{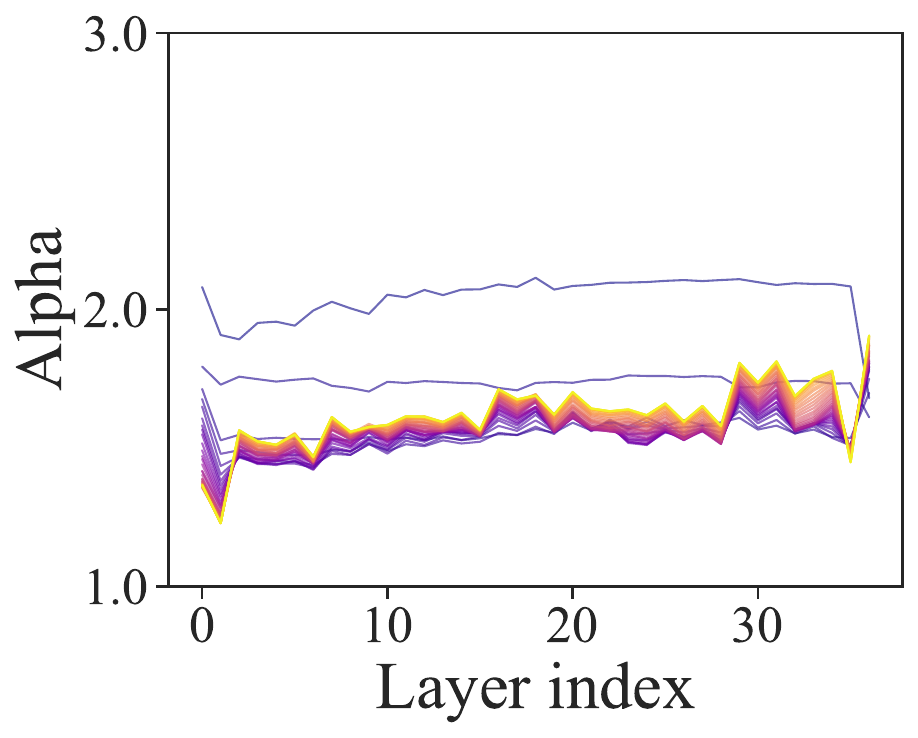}
        \caption{ResNet 34, LS+Ours}
        \label{subfig:resnet34_ours_ls}
    \end{subfigure}

    \begin{subfigure}{0.24\linewidth}
        \centering
        \includegraphics[width=\textwidth]{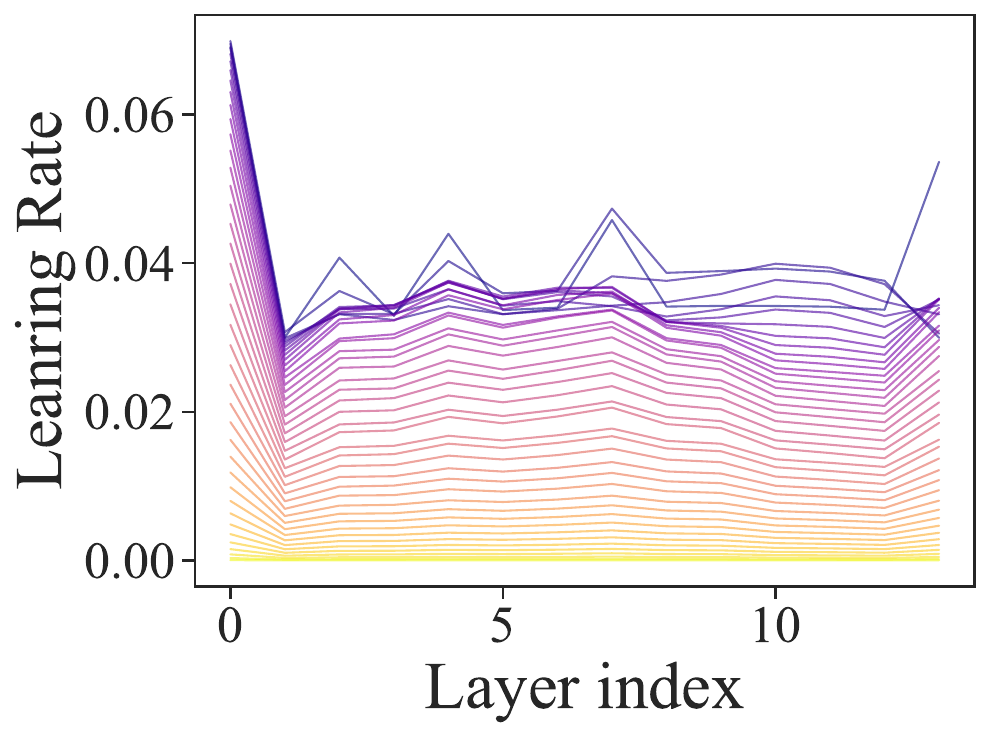}
    \end{subfigure}
    \hfill
    \begin{subfigure}{0.24\linewidth}
        \centering
        \includegraphics[width=\textwidth]{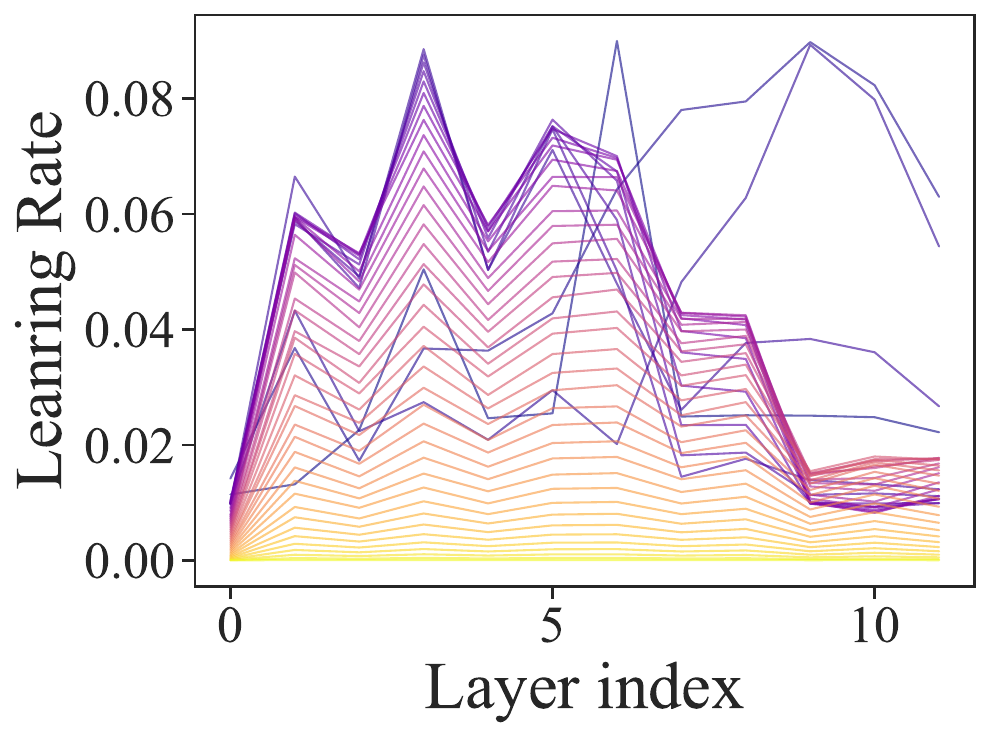}
    \end{subfigure}
    \hfill
    \begin{subfigure}{0.24\linewidth}
        \centering
        \includegraphics[width=\linewidth]{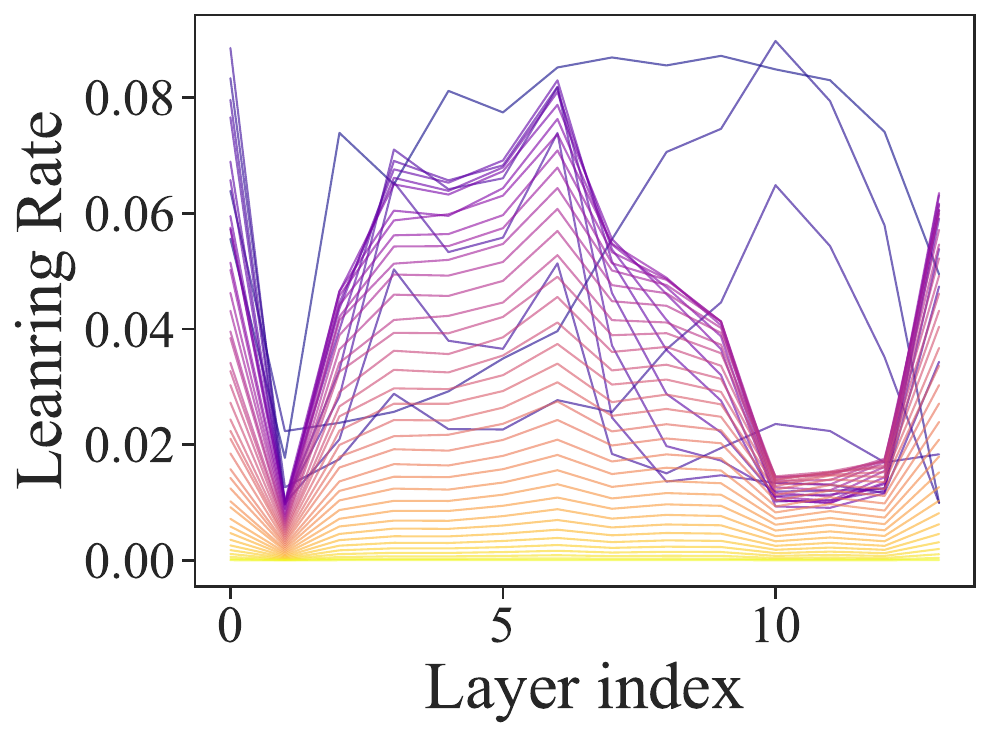}
    \end{subfigure}
    \hfill
    \begin{subfigure}{0.24\linewidth}
        \centering
        \includegraphics[width=\linewidth]{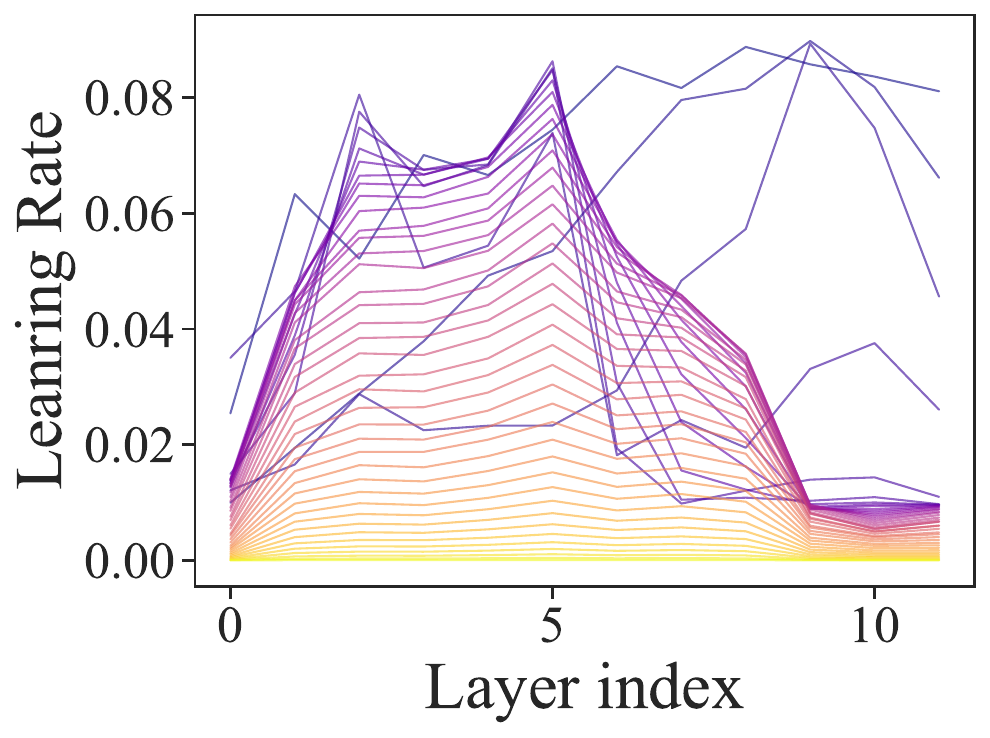}
    \end{subfigure}

    \begin{subfigure}{0.24\linewidth}
        \centering
        \includegraphics[width=\textwidth]{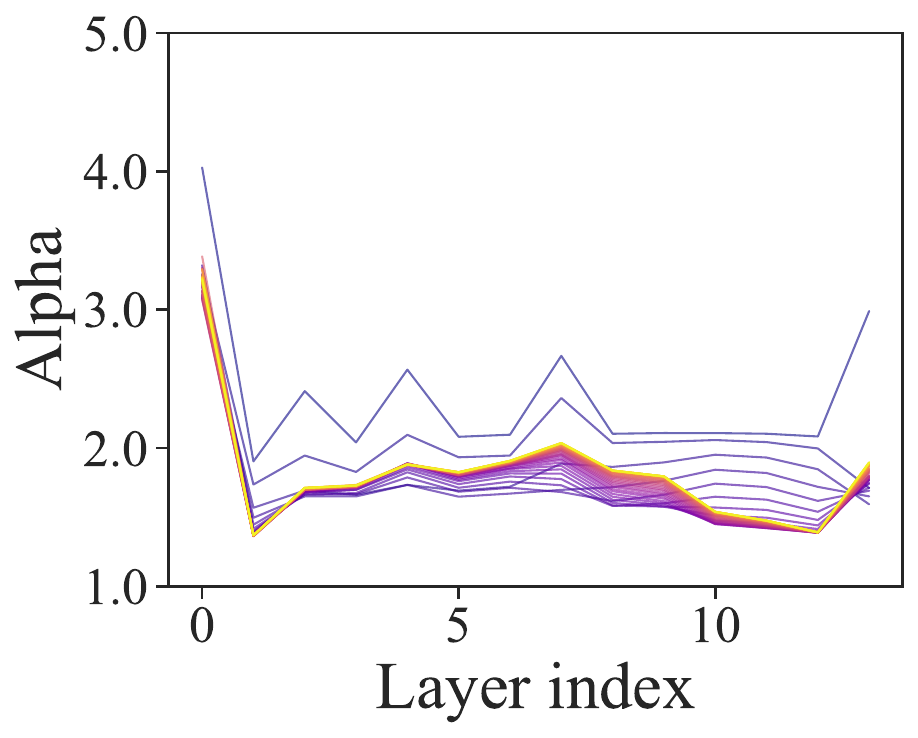}
        \caption{VGG 16, TB(no LS)}
        \label{subfig:vgg16_tb}
    \end{subfigure}
    \hfill
    \begin{subfigure}{0.24\linewidth}
        \centering
        \includegraphics[width=\textwidth]{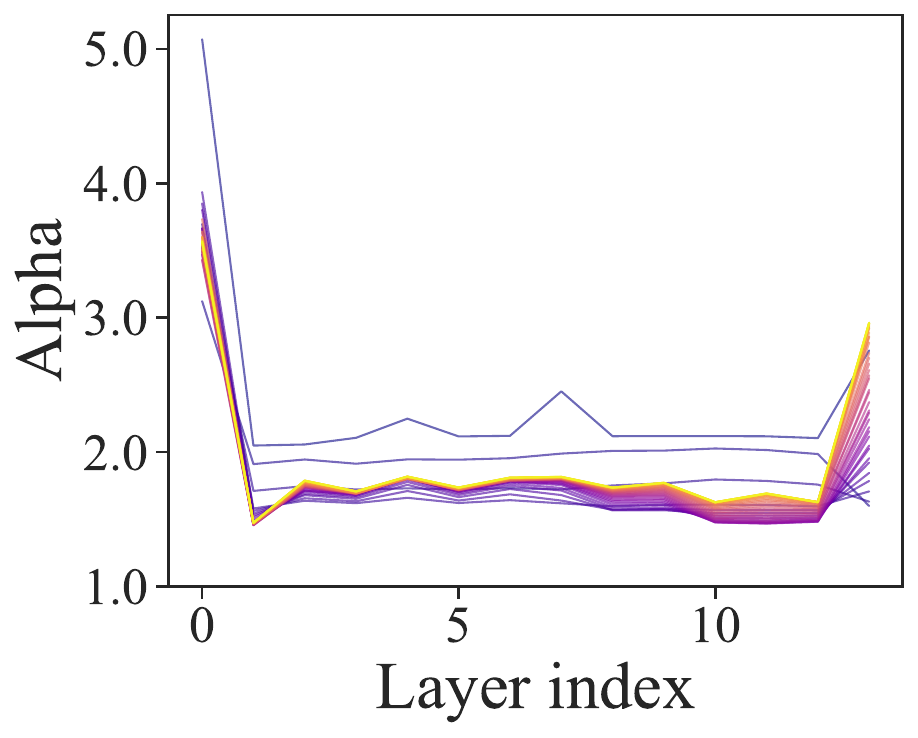}
        \caption{VGG 16, LS+TB}
        \label{subfig:vgg16_lstb}
    \end{subfigure}
    \hfill
    \begin{subfigure}{0.24\linewidth}
        \centering
        \includegraphics[width=\linewidth]{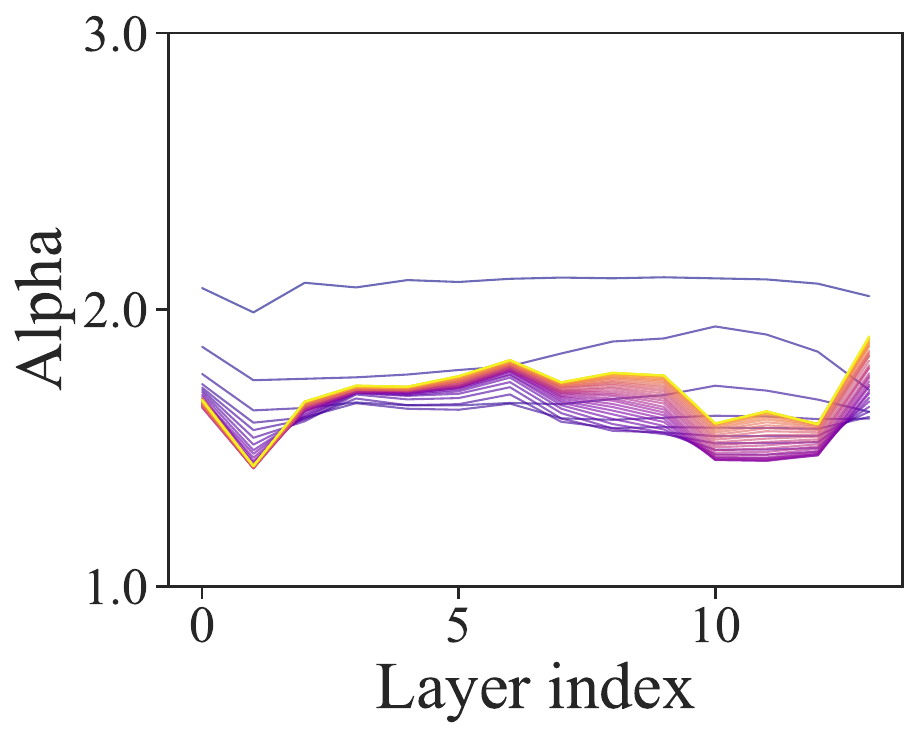}
        \caption{VGG 16, Ours(no LS)}
        \label{subfig:vgg16_ours}
    \end{subfigure}
    \hfill
    \begin{subfigure}{0.24\linewidth}
        \centering
        \includegraphics[width=\linewidth]{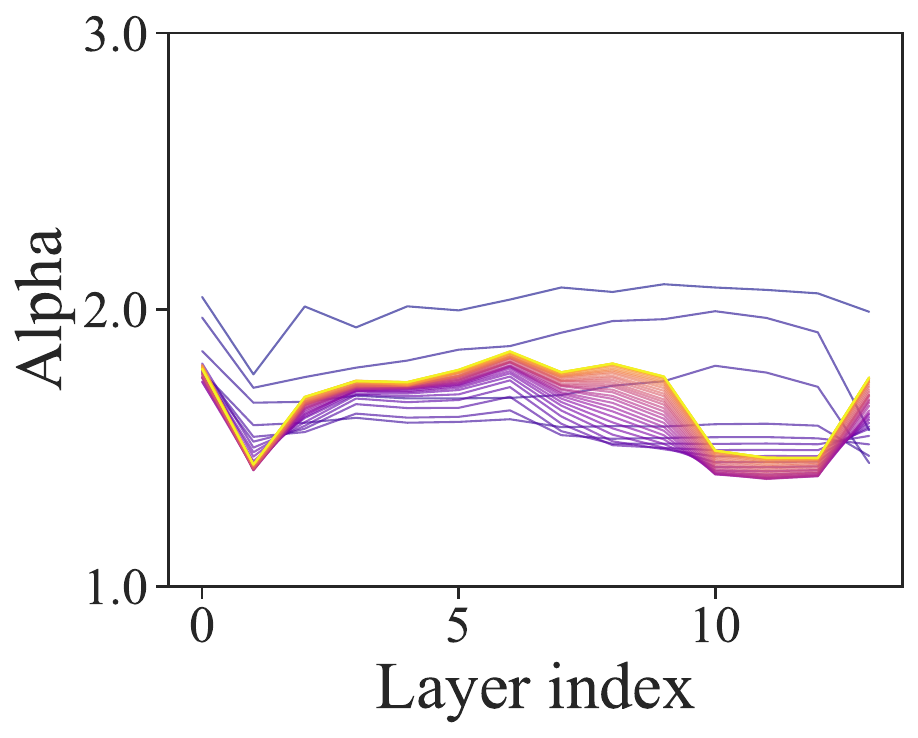}
        \caption{VGG 16, LS+Ours}
        \label{subfig:vgg16_ours_ls}
    \end{subfigure}
    
    \caption{Visualization of layer-wise Learning Rate and \AlphaHill (Alpha) over training. The top rows of each subfigure presents the evolution of layer-wise learning rates and the bottom rows presents \AlphaHill during training for different optimizing configurations of ResNet 34 and VGG 16 on CIFAR 100. The color gradient represents training epochs, transitioning from dark purple (early epochs) to yellow (later epochs).}

    \label{fig:alpha_lr_change_byepochs_acrosslayers}
\end{figure}

We provide visualization to compare how the \AlphaHill and learning rates are distributed over layers during the training with different layer-wise optimization settings. In Figure \ref{fig:alpha_lr_change_byepochs_acrosslayers}, we report the learning rate and \AlphaHill every five epochs throughout the 200-epoch training duration. We can find that in Figure \ref{subfig:resnet34_tb}, \ref{subfig:vgg16_tb} and Figure \ref{subfig:resnet34_lstb}, \ref{subfig:vgg16_lstb}, if the previous \TB does not apply LS to exclude layers with aspect ratio bias (e.g., the first and last layers of the model), the layer-wise learning rate allocation will be inaccurate. This misallocation leads to a few layers having excessively large learning rates while most layers have very small ones. This ultimately results in poor model performance or an imbalanced training process. In contrast, \ourmethod (see results in Figures \ref{subfig:resnet34_ours}, \ref{subfig:resnet34_ours_ls}, \ref{subfig:vgg16_ours} and \ref{subfig:vgg16_ours_ls}) effectively eliminates the measurement inaccuracies caused by aspect ratio bias. As a result, it ensures that learning rates are properly allocated regardless of whether LS is applied.

\subsection{Computation Cost Analysis}

In Table \ref{table:computation_cost_allexps}, we report the computational cost of eigenspectrum analysis across different models and methods. The most computation-intensive aspect of these layer analysis methods involves performing SVD on weight matrices, which can be optimized using parallel processing techniques. In the LLM pruning task, we use 8 L40 GPUs for weight analysis and record the \AlphaHill values for each layer. We select four different sparsity ratios, three pruning methods, and two evaluation approaches, yielding a total of 24 experimental configurations per model. However, since weight analysis is performed only once per model, the additional computational cost per experiment remains minimal. When the number of submatrices becomes particularly large, the additional computational overhead introduced by \ourmethod increases accordingly. In CV and SciML experiments, we use a single L40 GPU and do weight analysis every epoch during the training and fine-tuning. Compared to previous methods, the additional computational cost introduced by \ourmethod is not particularly significant.

We acknowledge that computational cost can indeed be higher than previous methods. But there are certain ways to mitigate the issue, e.g., by using a larger window size (like 4096 for LLaMA 7B/5120 for LLaMA 13B) and a limited number of sampling steps. For example, in Table ~\ref{tb:window_size_step_size_ablation_study}, when we use larger window size(such as 4096) and smaller sampling steps(such as 5), our method can still improve model performance and reduce the computational cost.

\newlength{\methodtbw}
\ificml
  \setlength{\methodtbw}{.7\linewidth}
\else
  \setlength{\methodtbw}{.80\linewidth}
\fi

\begin{table}[!htb]
    \centering
    \caption{Computation cost for each experiment.}
    \vspace{0.25cm}
    \resizebox{0.9\linewidth}{!}{

     \begin{minipage}[t]{\methodtbw}
        \centering
            \begin{tabular}{ccccc}
                \toprule
                \bf{Model}   & \bf{Experiment Settings}  & \bf{Weight Analysis Time (sec/experiment)} \\
                \midrule
                                & AlphaPruning                   &  2.65  \\
                 LLaMA-7B       & Ours (4096, 5)                 &  6.08  \\
                                & Ours (2000, $15 \times 15$)    &  134.5  \\

                 \midrule
                                & AlphaPruning                   &  6.23  \\
                 LLaMA-13B      & Ours (5120, 5)                 &  16.65  \\
                                & Ours (2000, $15 \times 15$)    &  163.26  \\
        
                \bottomrule 
            \end{tabular}
     \end{minipage}
     \begin{minipage}[t]{\methodtbw}
        \centering
            \begin{tabular}{ccccc}
                \toprule
                \bf{Model}   & \bf{Method}  & \bf{Weight Analysis Time (sec/epoch)} \\
                \midrule
                 \multirow{2}{*}{ResNet 18}  & TB     &  0.887  \\
                                             & Ours   &  1.054  \\
        
                 \midrule
                 \multirow{2}{*}{ResNet 34}  & TB     &  1.761  \\
                                             & Ours   &  1.946  \\
        
                 \midrule
                 \multirow{2}{*}{VGG 16}  & TB     &  1.048  \\
                                          & Ours   &  1.200  \\
        
                 \midrule
                 \multirow{2}{*}{VGG 19}  & TB     &  1.421  \\
                                          & Ours   &  1.578  \\
        
                 \midrule
                 \multirow{2}{*}{DPOT-Tiny}  & TB\_Sig  &  0.235  \\
                                             & Ours     &  0.278  \\
        
                 \midrule
                 \multirow{2}{*}{DPOT-Small}  & TB\_Sig  &  1.255  \\
                                              & Ours     &  1.340  \\

                \bottomrule
            \end{tabular}
        \end{minipage}  
    }
     \label{table:computation_cost_allexps} 
\end{table}

\subsection{LLM Pruning Stability Analysis}

The experiment results from ~\citep{yin2024outlier} show that layer-wise pruning LLMs at high sparsity is a challenging optimization problem. If sparsity ratios are not properly allocated, the performance of the model can become very unstable. Therefore, here we aim to demonstrate that the performance improvements observed in our experiments are due to more accurate layer-wise analysis, rather than accidental factors such as random seeds.

\begin{figure*}[!tbh]
    \centering
    \begin{subfigure}{0.6\linewidth}
        \includegraphics[width=\linewidth]{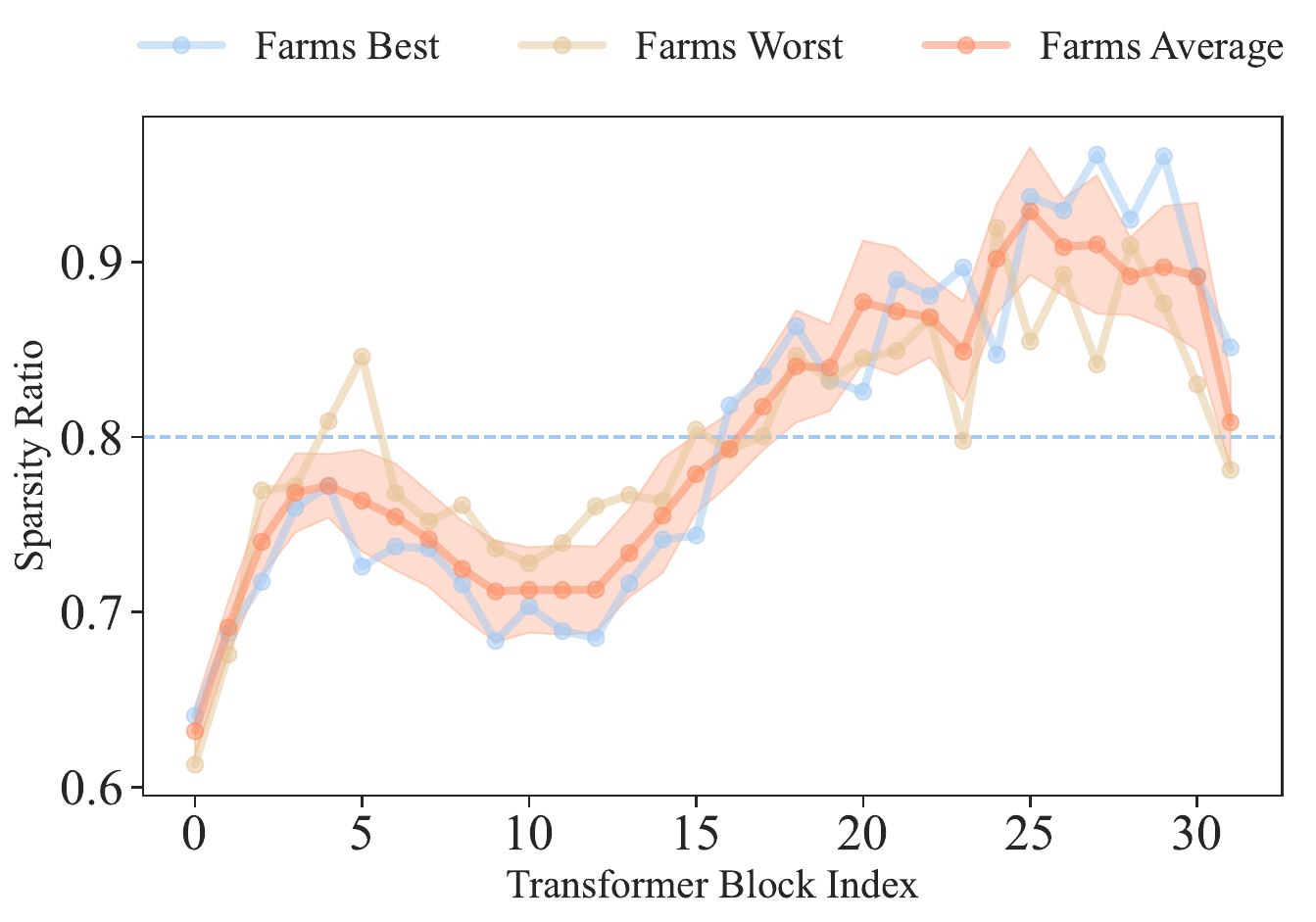}
    \end{subfigure}
    \caption{Block-wise Sparsity Ratio for LLaMA-7B assigned by FARMS. We set up FARMS with four different window sizes \{500, 1000, 2000, 4096\} and four sampling setps \{5, 10, 15, 20\}.  The pruning method is SparseGPT and the base sparsity ratio is 0.8. The FARMS Best indicates using this sparsity ratio distribution makes minimal perplexity \textbf{79.42{\scriptsize$\pm$3.86}} while the FARMS worst makes largest perplexity \textbf{99.23{\scriptsize$\pm$3.53}}. The distribution of block-wise sparsity ratios of FARMS Average represents the mean under the $4 \times 4$ settings, with the standard deviation intervals also illustrated. } 
    \label{fig: sparsity_ratio_vis}
\end{figure*}

We follow the experiment settings from Table ~\ref{tb:window_size_step_size_ablation_study} and visualize the sparsity ratio assigned to each transformer block in LLaMA-7B under different sampling settings in Figure ~\ref{fig: sparsity_ratio_vis}. This visualization indirectly reflects the degree of heavy-tailedness in the ESD of each transformer block’s model layer as analyzed by our method.  We observe that the differences in sparsity ratio assignments across these settings are not significant. However, the assignment from the best FARMS setting is still distinct from the worst setting, which explains the performance difference in Table ~\ref{tb:window_size_step_size_ablation_study}. 

\section{Training Quality Analysis}

\subsection{Previous work on training quality}

Previous work on HT-SR has established that the heavy-tailness of ESDs is strongly correlated with the test accuracy of models ~\citep{martin2021implicit,martin2021trends}. While this does not imply that "training quality" is identical to "test accuracy," the correlation between heavy-tailedness and test accuracy has been used to justify HT-SR metrics. Therefore, improving test accuracy or similar performance metrics (e.g., perplexity) remains our primary goal.

Although previous work on HT-SR does not explicitly define "training quality," several related quantities have been mentioned: (1) strong correlation between weight elements ~\citep{martin2021implicit,martin2021trends} and (2) feature spikes and their alignment with the target teacher model ~\citep{ba2022high,wang2023spectral}. The feature spike, analyzed in the context of a single-index teacher and a two-layer student model ~\citep{ba2022high,wang2023spectral}, is approximately a rank-one update to the model (in the limit of infinite matrix size with fixed aspect ratio) and also persists after matrix subsampling. This is because the specific form of the rank-one update makes it cover the whole matrix with probability one.

\subsection{A toy experiment to measure training quality}

We designed a toy experiment to test the correlation between "training quality" and the new HT-SR metric measured using FARMS. Following ~\citep{ba2022high,wang2023spectral}, we use a single-index teacher to generate signals to train a two-layer student. The first layer of the student model is a weight matrix, while the second layer is a weight vector. We only update the weight of first layer. To measure "training quality", during training, we measure the alignment between the weight matrix and the ground truth target vector of the teacher model similar to ~\citep{ba2022high,wang2023spectral}, and we define this alignment to be the "training quality" of the student model.

Throughout the training process, we select the student network checkpoint with the highest alignment and report both the alignment value and the \AlphaHill value. We then vary the sizes of the student model with different weight matrix aspect ratios on a fixed input dimension 500 to conduct multiple experiments. Each experiment provides one \AlphaHill value and one alignment value. The multiple experiments (conducted using varying sizes) produce one curve for \AlphaHill and one curve for the alignment value.

\begin{figure}[!tbh]
    \centering
    \begin{subfigure}{0.46\linewidth}
        \centering
        \includegraphics[width=\linewidth]{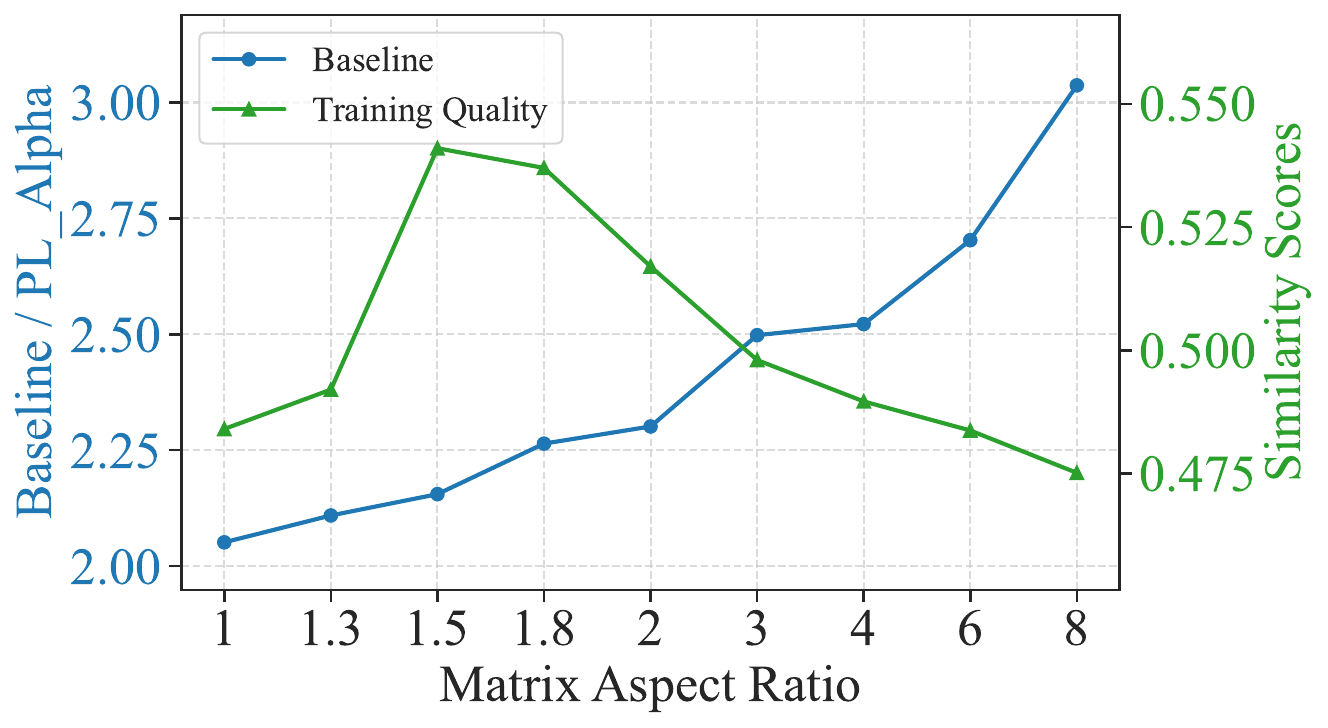}
        \caption{Baseline}
        \label{subfig:baseline_sim_score}
    \end{subfigure}
    \begin{subfigure}{0.46\linewidth}
        \centering
        \includegraphics[width=\linewidth]{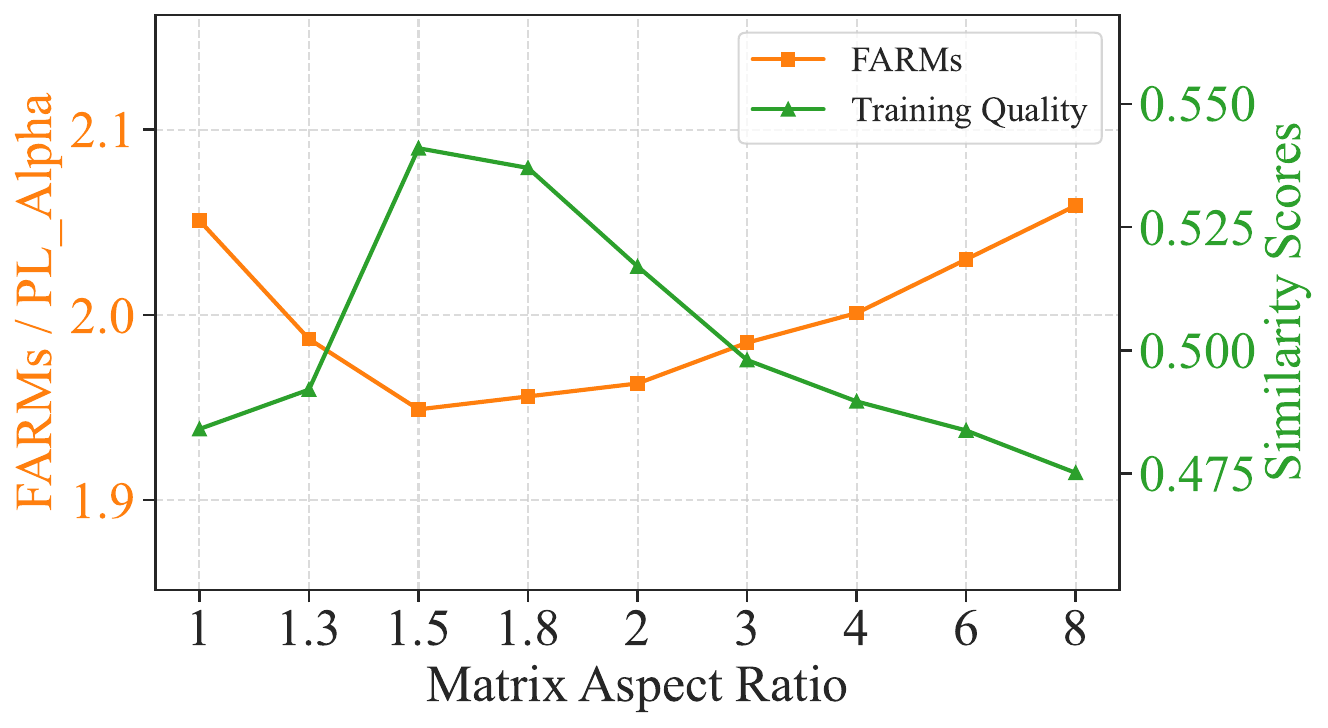}
        \caption{\ourmethod}
        \label{subfig:farms_sim_score}
    \end{subfigure}
    \caption{Compare the \AlphaHill from FARMS and Baseline in measuring the training quality of a single layer. The Correlation Coefficient between FARMS and training quality is -0.89 and for baseline is -0.51. We can find that FARMS can measure the training quality more precisely.} 
    \label{fig:sim_socre}
\end{figure}

We then plot the two curves using both existing methods for estimating \AlphaHill and our method FARMS. As shown in Figure ~\ref{fig:sim_socre}, FARMS reveals a clear negative correlation between the two curves: the better the training quality, the larger the alignment, and the smaller the \AlphaHill. However, for the existing method, due to the aspect ratio bias, the correlation is incorrect.

\section{Broader Discussion}

In this section, we discuss more related literature as well as some explanations regarding aspect ratio bias and issues related to our method.
 
\subsection{Relationship with Matrix Shape}

One should recognize that the numerical value of the same property measured in different network layers can naturally have different scales, and thus should not be directly compared without proper normalization.
In some real-world application scenarios, such as in LLM Pruning ~\citep{lu2024alphapruning,liu2025bawa}, this factor need to be considered.
When the network size changes, hyperparameters also need to be changed accordingly. For example, works such as Tensor Programs IV ~\citep{yang2021tensor} and Tensor Programs V ~\citep{yang2022tensor} have explored how layer sizes affect the optimal scaling of weight initializations and learning rates. Tensor Programs IV introduced the Maximal Update Parametrization (µP) to ensure feature learning by carefully choosing parameter scaling rules based on these sizes, addressing how existing parametrizations might otherwise collapse to kernel regimes.
Building on this, Tensor Programs V demonstrated that µP can enable zero-shot hyperparameter transfer, where hyperparameters tuned on smaller models remain optimal for significantly larger ones.
In this context, ``shape’’ awareness pertains to designing size-dependent hyperparameters that maintain stable and effective (non-kernel) training dynamics.

Nevertheless, our work is fine-grained in that we consider the aspect ratios of different layers (number of rows versus columns).
We demonstrate that varying aspect ratios can artificially stretch or compress the ESD. This distortion confounds the interpretation of heavy-tailed (HT) metrics. To mitigate this measurement bias, we propose \ourmethod to analyze the average ESD of submatrices sampled at a fixed aspect ratio. This approach provides a normalized HT metric, enabling more reliable comparisons of such spectral diagnostics across different layers within a given network.
Thus, while focusing on shape-aware parametrization for training stability and transfer, our contribution lies in a shape-aware analysis technique (\ourmethod) aimed at correcting measurement bias in spectral diagnostics. 

\subsection{Why does the \ourmethod preserve important spectral property about the original matrix?}

The goal of measuring heavy-tailness in HT-SR is to evaluate the strength of correlations introduced by training, as established in previous work ~\citep{martin2021implicit}. However, when we subsample a single submatrix and measure correlations only within that submatrix, some correlations between elements in the subsampled matrix and those outside it are inevitably lost. This motivates our approach of using multiple submatrices to capture a broader range of correlations.

\subsection{Does this approach introduce additional bias?}

Our approach could be viewed as introducing a form of "bias"; however, we interpret this more specifically as achieving partial coverage of the entire matrix. Conceptually, this is similar to the principle behind bootstrap sampling in random forests, where multiple samples, each with potentially limited coverage, are used collectively to mitigate the effects of this partial view and improve overall model robustness.

Further justification for this perspective comes from recent work ~\citep{wang2023spectral,kothapalli2024crafting} that aims to theoretically quantify heavy-tailedness. These studies interpret heavy-tailedness as the accumulation and evolution of feature spikes in the ESD that align with the teacher model's features. Critically, these feature spikes are characterized as being approximately rank-one updates to the original matrix Because a rank-one component inherently covers the whole matrix, sampling a submatrix will, with high probability, capture that rank-one component. Therefore, this subsampling process is unlikely to miss the feature spikes, which are identified by previous work as the cause of the heavy-tail structure. We believe this provides substantial evidence that FARMS can preserve important spectral information, specifically as measured by these feature spikes in the ESD.

\section{Hyperparameter Adjustment}
\label{appendix:Hyperparamter_settings}

In this section, we report the hyperparameters that we use in the experiments shown in the main paper (Section \ref{section:empirical_results}).

First, we report the common hyperparameters shared by Image Classification experiments (Section \ref{wwsampling: cv_tempbalance}): the optimizer is SGD, batch size 128, number of total training epochs 200, weight decay 5e-4, and momentum 0.9. For each experiment setting, we repeat our experiments with three random seeds \{43, 37, 13\}. We also report the mean and standard deviation of the test accuracy across these seeds. In Table \ref{table:resnet_vgg_hyper}, we report the details of experiments for each model and method. We use the same learning rate range from ~\citep{zhou2024temperature} and we expand the scaling ratio range into [(0.1, 1.9), (0.2, 1.8), (0.3, 1.7), (0.4, 1.6), (0.5, 1.5), (0.6, 1.4), (0.7, 1.3), (0.8, 1.2), (0.9, 1.1)] nine choices. 

Second, we provide the hyperparameters used in experiments of LLM pruning and SciML. We follow the common hyperparameter settings as described in \citet{lu2024alphapruning, liu2024model}. See more details for other hyperparameters like $\tau$ in LLM pruning and scaling ratios in SciML in Table \ref{table:llama_dpot_hyper}.

Finally, we report the detailed matrix subsampling settings used in every model in Table \ref{table:resnet_vgg_subsampling_hyper}. We cannot use a very small window size or sampling steps because doing so may not cover the entire matrix. Conversely, selecting a very large size would result in too much overlap between sampled matrices. For ResNet, VGG and DPOT series models, we use the minimum dimension of the weight matrices to construct the sampled submatrices based on the parameter $Q$. We also select the $\left\lfloor {m}/{n} \right\rfloor$ for the submatrices number, where $m$, $n$ is the dimension of weight matrix, $m \geq n$. But for the final layer of ResNet and VGG models, we select nine submatrices based on experiments results. For LLaMA series models, we apply sliding window sampling using multiple moderately sized submatrices, resulting in a smoother \AlphaHill estimation.  

\begin{table}[!htb]
    \centering
    \caption{Parameter settings of the experiment reported in Section \ref{wwsampling: cv_tempbalance}.} 
    \vspace{0.25cm}
    \resizebox{0.66\columnwidth}{!}{
    \begin{tabular}{ccccc}
        \toprule
        \bf{Model} & \bf{Method} & \bf{Initial} & \bf{Scaling Ratio}  & \bf{Test Accuracy} \\
         &  & \bf{Learning Rate}   & $(s_1, s_2)$  &  \\
        \midrule
        & CAL     &       0.05, \textbf{0.1}, 0.15   & -  & 78.23{\scriptsize$\pm$0.087}\\
        & TB(no LS)      &        0.1   & (0.6, 1.4) & 78.76{\scriptsize$\pm$0.111} \\
        ResNet 18 & LS+TB    & 0.1 & (0.2, 1.8) & 79.31{\scriptsize$\pm$0.180} \\
        & Ours(no LS)    &       0.1   & (0.1, 1.9) & 79.49{\scriptsize$\pm$0.080} \\
        & LS+Ours &       0.1   & (0.1, 1.9) & \textbf{79.53{\scriptsize$\pm$0.177}} \\
        
        \midrule
        & CAL     &       \textbf{0.05}, 0.1, 0.15  & - & 78.99{\scriptsize$\pm$0.137} \\
        & TB(no LS)      &       0.1   & (0.5, 1.5) & 79.64{\scriptsize$\pm$0.029} \\
        ResNet 34 & LS+TB    & 0.1 & (0.3, 1.7) & 80.00{\scriptsize$\pm$0.090} \\
        & Ours(no LS)    &       0.1   & (0.2, 1.8) & 80.17{\scriptsize$\pm$0.213} \\
        & LS+Ours &        0.1  & (0.3, 1.7) & \textbf{80.20{\scriptsize$\pm$0.221}} \\
        
        \midrule
        & CAL     &       0.025, \textbf{0.05}, 0.1   & - & 74.30{\scriptsize$\pm$0.078} \\
        & TB(no LS)      &        0.05  & (0.6, 1.4) & 74.43{\scriptsize$\pm$0.158} \\
        VGG 16    & LS+TB   &  0.05 & (0.3, 1.7) & 75.19{\scriptsize$\pm$0.131} \\
        & Ours(no LS)    &        0.05   & (0.2, 1.8) & \textbf{75.36{\scriptsize$\pm$0.118}} \\
        & LS+Ours &        0.05   & (0.2, 1.8) & 75.15{\scriptsize$\pm$0.247} \\
        
        \midrule
        & CAL     &       0.025, \textbf{0.05}, 0.1   & -  & 73.11{\scriptsize$\pm$0.113} \\
        & TB(no LS)      &       0.05   & (0.6, 1.4) & 73.22{\scriptsize$\pm$0.277} \\
        VGG 19    & LS+TB    & 0.05 & (0.2, 1.8) & 74.19{\scriptsize$\pm$0.159} \\
        & Ours(no LS)    &       0.05   & (0.2, 1.8) & \textbf{74.28{\scriptsize$\pm$0.392}} \\
        & LS+Ours &       0.05   & (0.4, 1.6) & 73.99{\scriptsize$\pm$0.300} \\
        \bottomrule
    \end{tabular}
    }
    \label{table:resnet_vgg_hyper} 
\end{table}

\begin{table}[!thb]
    \centering
    \caption{Hyperparameters for LLaMA and DPOT models. (Left) The range of $\tau$ used for LLM pruning. (Right) Learning rate and scaling ratio settings for DPOT series models at different subsampling ratios. } 
    \vspace{0.25cm}
    \resizebox{\linewidth}{!}{

      \begin{minipage}[t]{0.55\linewidth}
        \centering
        \begin{tabular}{c|ccc}
            \toprule
            \bf{Sparsity Ratio}  & \multicolumn{3}{c}{\bf{LLaMA-7B/13B}} \\
            \midrule
            0.7 
            & \multicolumn{3}{c}{0.1, 0.2, 0.3, 0.4, 0.5, 0.6}  \\
            \midrule
            0.75 
            & \multicolumn{3}{c}{0.1, 0.2, 0.3, 0.4, 0.5, 0.6} \\
            \midrule
            0.8 
            & \multicolumn{3}{c}{0.1, 0.2, 0.25, 0.3, 0.4, 0.5}  \\
            \midrule
            0.85 
            & \multicolumn{3}{c}{0.1, 0.15, 0.2, 0.25, 0.3}  \\
            \bottomrule
        \end{tabular}
    \end{minipage}
    \begin{minipage}[t]{0.8\linewidth}
    \centering
    \begin{tabular}{c|cccc}
        \toprule
        \bf{Model} & \multicolumn{2}{c}{\bf{DPOT-Tiny}} & \multicolumn{2}{c}{\bf{DPOT-Small}} \\
        \midrule
        \bf{Hyperparameters} & \bf{Learning Rate} & \bf{Scaling Ratio} & \bf{Learning Rate} & \bf{Scaling Ratio} \\
        \midrule
        5\%   & 2.5e-4 & (1.0, 1.0) & 1e-4   & (1.0, 1.0) \\
        \midrule
        10\%  & 2.5e-4 & (1.0, 1.0) & 1e-4   & (1.0, 1.0) \\
        \midrule
        25\%  & 2.5e-4 & (1.0, 1.0) & 2.5e-4 & (1.0, 1.0) \\
        \midrule
        50\%  & 5e-4   & (1.0, 1.0) & 2.5e-4 & (1.0, 1.0) \\
        \midrule
        100\% & 5e-4   & (1.0, 1.0) & 2.5e-4 & (1.0, 1.0) \\
        \bottomrule
    \end{tabular}
    \end{minipage}
    }
    \label{table:llama_dpot_hyper}
\end{table}

\begin{table}[!htb]
    \centering
     \caption{Subsampling Hyperparameters for different models.}
    \vspace{0.25cm}
    \resizebox{0.8\linewidth}{!}{
    \begin{tabular}{cccccc}
        \toprule
        \bf{Model}   & \bf{Aspect Ratio($Q$)}  & \bf{Window Size} & \bf{Submatrices Number}\\
        \midrule
         ResNet 18/34, VGG 16/19   & 1.0  & Minimum Dimension  & $\left\lfloor {m}/{n} \right\rfloor$, 9  \\

         \midrule
         DPOT-Tiny/Small   & 1.0  & Minimum Dimension  & $\left\lfloor {m}/{n} \right\rfloor$  \\
         
         \midrule
                     & 1.0  & 2000  & $15 \times 15$  \\
         LLaMA-7B    & 1.0  & 2000  & $10 \times 10$  \\
                     & 1.0  & 1000  & $10 \times 10$  \\

         \midrule
         LLaMA-13B    & 1.0  & 2000  & $15 \times 15$  \\
        \bottomrule
    \end{tabular}
    }
   \label{table:resnet_vgg_subsampling_hyper} 
\end{table}


\end{document}


\else


\usepackage{fullpage}

\usepackage[utf8]{inputenc}
\usepackage{graphicx}

\usepackage{enumitem}
\usepackage{amsmath}
\usepackage{amsfonts}
\usepackage{dcolumn} 
\usepackage{tabu}
\usepackage{array}
\usepackage{xspace}
\usepackage{makecell}
\usepackage{amsthm}
\usepackage{algorithmic}
\newtheorem*{remark}{Remark}

\usepackage{colortbl}
\usepackage{booktabs, multirow} 
\usepackage{soul}
\usepackage{changepage,threeparttable} 

\usepackage[dvipsnames]{xcolor}
\usepackage{color, colortbl}
\usepackage{multirow}
\usepackage{enumitem}
\usepackage{placeins}
\usepackage{subcaption}
\usepackage{microtype}
\usepackage{graphicx}
\usepackage{booktabs} 
\usepackage{multirow}
\usepackage{enumitem}
\usepackage{graphicx}
\usepackage{color}
\usepackage{xcolor}
\usepackage{colortbl}
\definecolor{dullGray}{HTML}{F3F3F3}
\definecolor{smallColorT}{HTML}{fff1e6}
\definecolor{largeColorT}{HTML}{edf2fb}

\newcommand{\ourmethod}{\texttt{FARMS}\xspace}
\newcommand{\TB}{\texttt{TempBalance}\xspace}
\newcommand{\CAL}{\texttt{CAL}\xspace}
\newcommand{\Alphapruning}{\texttt{AlphaPruning}\xspace}
\newcommand{\TBSigmoid}{\texttt{TB\_Sigmoid}\xspace}
\newcommand{\AlphaHill}{\texttt{PL\_Alpha\_Hill}\xspace}
\newcommand{\Alpha}{\texttt{PL\_Alpha}\xspace}
\newcommand{\AlphaLoRA}{\texttt{AlphaLoRA}\xspace}

\definecolor{LightGrey}{HTML}{F3F3F3}
\newcommand\glb{ \rowcolor{LightGrey}}
\usepackage[ruled,linesnumbered]{algorithm2e}
\usepackage{algorithmic}
\usepackage[round,semicolon,sort]{natbib}

\usepackage{tcolorbox}
\usepackage{microtype}
\usepackage{graphicx}
\usepackage{subcaption}
\usepackage{booktabs}
\usepackage{nicefrac}  
\newcommand{\theHalgorithm}{\arabic{algorithm}}
\newcommand{\bigspace}{\hspace{15em}}
\definecolor{mydarkblue}{rgb}{0,0.08,0.45}
\usepackage[colorlinks,citecolor=mydarkblue,urlcolor=mydarkblue,linkcolor=mydarkblue]{hyperref}
\usepackage{url}

\let\oldcite\cite
\renewcommand{\cite}[1]{\citep{#1}}

\begin{document}

\title{Eigenspectrum Analysis of Neural Networks without \\Aspect Ratio Bias}
\date{}

\author{%
  Yuanzhe Hu$^{1}$, 
  Kinshuk Goel$^{2,3}$, 
  Vlad Killiakov$^{4}$, 
  Yaoqing Yang$^{2}$ \\
  $^1$ Department of Computer Science and Engineering, University of California, San Diego \\
  $^2$ Department of Computer Science, Dartmouth College\\
  $^3$ Department of Computer Science \& Engineering, SRM Institute of Science \& Technology \\
  $^4$ Independent Researcher, University of California, Berkeley \\
}

\maketitle

\begin{abstract}

Diagnosing deep neural networks (DNNs) by analyzing the eigenspectrum of their weights has been an active area of research in recent years.
One of the main approaches involves measuring the \emph{heavytailness} of the empirical spectral densities (ESDs) of weight matrices.
This analysis has been shown to provide insights to help diagnose whether a model is well-trained or undertrained, and has been used to guide training methods involving layer-wise hyperparameter assignment.
In this paper, we address an often-overlooked challenge in estimating the heavytailness of these ESDs: the impact of the aspect ratio of weight matrices. 
We demonstrate that matrices of varying sizes (and aspect ratios) introduce a non-negligible bias in estimating the heavytailness of ESDs, leading to inaccurate model diagnosis and layer-wise hyperparameter assignment.
To overcome this challenge, we propose \ourmethod (\textbf{F}ixed-\textbf{A}spect-\textbf{R}atio \textbf{M}atrix \textbf{S}ubsampling), a method that normalizes the weight matrices by subsampling submatrices with a fixed aspect ratio.
Instead of measuring the heavytailness of the original ESD, we measure the average ESD of these subsampled submatrices.
We show that this method effectively mitigates the aspect ratio bias.
We validate our approach across various optimization techniques and application domains that involve eigenspectrum analysis of weights, including image classification in computer vision (CV) models, scientific machine learning (SciML) model training, and large language model (LLM) pruning. Our results show that despite its simplicity, \ourmethod uniformly improves the accuracy of eigenspectrum analysis while enabling more effective layer-wise hyperparameter assignment.
In one of the LLM pruning experiments, \ourmethod reduces the perplexity of the LLaMA-7B model by 17.3\% when compared with state-of-the-art methods.

\end{abstract}

\section{Conclusion}

We propose a subsampling strategy to address the measurement bias introduced by varying aspect ratios of weight matrices in HT-SR eigenspectrum analysis. The main idea of our method is to extract submatrices of the original matrix with a fixed aspect ratio, followed by averaging the ESDs of these submatrices to accurately assess the training quality of each weight matrix. Our extensive experiments demonstrate that our method can precisely estimate layer-wise training quality, improve the performance of layer-wise optimization algorithms, and contribute to more balanced training and efficient pruning. Furthermore, results on visual analytics confirm the effectiveness of our approach, showing that it successfully eliminates the measurement bias caused by aspect ratio variations.

\bibliographystyle{plainnat}

\bibliography{main}

\newpage
\appendix
\onecolumn

\end{document}

\fi